\documentclass[acmsmall,screen]{acmart}

\setcopyright{none}     
\renewcommand\footnotetextcopyrightpermission[1]{}  
\makeatletter
\renewcommand{\@authorsaddresses}{} 
\makeatother

\usepackage{hyperref} 
\usepackage[perpage]{footmisc} 

\AtBeginDocument{%
  \providecommand\BibTeX{{%
    \normalfont B\kern-0.5em{\scshape i\kern-0.25em b}\kern-0.8em\TeX}}}

\usepackage{graphicx}
\usepackage{wrapfig}
\usepackage{float} 

\usepackage{csquotes}
\usepackage{xcolor}
\usepackage{bm}
\usepackage{wrapfig}
\usepackage{tabularray}
\usepackage{amsmath}  
\usepackage{mathtools}  
\usepackage[most]{tcolorbox}
\usepackage{tikz}  
\usepackage{pgfplots}  
\pgfplotsset{compat=1.18}  

\usepackage[caption=false,font=normalsize,labelfont=sf,textfont=sf]{subfig}

\usepackage{booktabs}  
\usepackage{adjustbox}
\usepackage{tabularx}
\usepackage{subcaption}
\usepackage{multirow}  
\usepackage{enumitem}  
\usepackage{algorithmic}  
\usepackage{algorithm}  
\usepackage{ulem}
\usepackage{forest} 

\usepackage{todonotes}

\usepackage{url}
\usepackage{float}  
\usepackage{comment}  

\usepackage[switch]{lineno}

\makeatletter

\renewcommand{\l@section}[2]{%
  \par\addpenalty{-\@highpenalty}%
  \vskip 3pt 
  \@dottedtocline{1}{0em}{2.3em}{\textbf{#1}}{#2}%
}

\renewcommand{\l@subsection}[2]{%
  \par\addpenalty{-\@highpenalty}%
  \vskip 1pt 
  \@dottedtocline{2}{1.5em}{2.8em}{#1}{#2}%
}

\renewcommand{\l@subsubsection}[2]{%
  \par\addpenalty{-\@highpenalty}%
  \vskip 1pt 
  \@dottedtocline{3}{3em}{3.3em}{#1}{#2}%
}

\makeatother

\usepackage{ifthen}
\usepackage{pifont}

\definecolor{brandblue}{rgb}{0.54, 0.7, 1}

\newcommand{\mathbox}[2][]{%
  \ifthenelse{\equal{#1}{post-pro}}{
    \tcboxmath[colback=yellow!30!white, colframe=white, boxrule=0pt, rounded corners, fontupper=\bfseries\large, left=0pt, right=0pt, top=2.2pt, bottom=2.2pt, boxsep=0pt]{#2}  
  }{
    \ifthenelse{\equal{#1}{model}}{
      \tcboxmath[colback=green!10!white, colframe=white, boxrule=0pt, rounded corners, fontupper=\bfseries\large, left=1pt, right=1pt, top=2.2pt, bottom=2.2pt, boxsep=0pt]{#2}  
    }{
      \ifthenelse{\equal{#1}{eval}}{
        \tcboxmath[colback=brandblue!30!white, colframe=white, boxrule=0pt, rounded corners, fontupper=\bfseries\large, left=3pt, right=3pt,top=2.2pt, bottom=2.2pt,  boxsep=1pt]{#2}  
      }{
        \ifthenelse{\equal{#1}{icl}}{
          \tcboxmath[colback=orange!20!white, colframe=white, boxrule=0pt, rounded corners, fontupper=\bfseries\large, left=1pt, right=1pt,top=2pt, bottom=2pt, boxsep=0pt]{#2}  
        }{
          \textcolor{black}{\large #2} 
        }
      }
    }
  }
}

\begin{document}

\title{A Survey on LLM-as-a-Judge}



\author{Jiawei Gu\textsuperscript{1,2*}, Xuhui Jiang\textsuperscript{1,3*}, Zhichao Shi\textsuperscript{1,4,*}, Hexiang Tan\textsuperscript{4}, Xuehao Zhai\textsuperscript{5}, Chengjin Xu\textsuperscript{1,3}, Wei Li\textsuperscript{4}, Yinghan Shen\textsuperscript{4}, Shengjie Ma\textsuperscript{1,6}, Honghao Liu\textsuperscript{1}, \\Saizhuo Wang\textsuperscript{1,7},Kun Zhang\textsuperscript{4}, Zhouchi Lin\textsuperscript{1}, Bowen Zhang\textsuperscript{1}, Lionel Ni\textsuperscript{7,8}, Wen Gao\textsuperscript{9}, Yuanzhuo Wang\textsuperscript{4,†}, Jian Guo\textsuperscript{1,†}}

\begingroup
\renewcommand{\thefootnote}{\fnsymbol{footnote}}
\footnotetext[1]{These authors contributed equally to this research.}
\footnotetext[2]{Corresponding author.}
\endgroup

\affiliation{\\
  \textsuperscript{1}IDEA Research, International Digital Economy Academy\country{China}\\
  \textsuperscript{2}Sun Yat-sen University \country{China}\\
    \textsuperscript{3}DataArc Tech Ltd\country{China}\\
  \textsuperscript{4}Institute of Computing Technology, Chinese Academy of Sciences \country{China}\\
  \textsuperscript{5}Department of Civil and Environmental Engineering, Imperial College London\country{UK} \\
  \textsuperscript{6}Gaoling School of Artificial Intelligence, Renmin University of China \\
    \textsuperscript{7}The Hong Kong University of Science and Technology
  \country{China} \\
    \textsuperscript{8}The Hong Kong University of Science and Technology (Guangzhou) 
  \country{China} \\
  \textsuperscript{9}Department of Computer Science and Technology, Peking University
  \country{China}
}

\renewcommand{\shortauthors}{J. Gu, X. Jiang, Z. Shi, J. Guo, et al.}


\begin{abstract}\label{abstract}

\section*{Abstract}
Accurate and consistent evaluation is crucial for decision-making across numerous fields, yet it remains a challenging task due to inherent subjectivity, variability, and scale. 
Large Language Models (LLMs) have achieved remarkable success across diverse domains, leading to the emergence of "LLM-as-a-Judge," where LLMs are employed as evaluators for complex tasks. With their ability to process diverse data types and provide scalable and flexible assessments, LLMs present a compelling alternative to traditional expert-driven evaluations. 
However, ensuring the reliability of LLM-as-a-Judge systems remains a significant challenge that requires careful design and standardization. 
This paper provides a comprehensive survey on LLM-as-a-Judge, offering a \textbf{formal definition} and a \textbf{detailed classification}, while focusing on addressing the core question: \textbf{How to built reliable LLM-as-a-Judge systems?} We explore strategies to enhance reliability, including improving consistency, mitigating biases, and adapting to diverse assessment scenarios. Additionally, we propose methodologies for evaluating the reliability of LLM-as-a-Judge systems, supported by a novel benchmark designed for this purpose. 
To advance the development and real-world deployment of LLM-as-a-Judge systems, we also discussed practical applications, challenges, and future directions. 

This survey serves as a foundational reference for researchers and practitioners in this rapidly evolving field. Our contributions span multiple levels: we establish the conceptual boundaries of LLM-as-a-Judge, reorganize fragmented literature into a unified framework, and propose a novel reliability-oriented benchmark. Building on these, we also articulate a forward-looking research agenda, offering both theoretical foundations and practical guidance for constructing reliable and socially trustworthy LLM-as-a-Judge systems.
The associated resources can be accessed at \textbf{\url{https://awesome-llm-as-a-judge.github.io/}}.
\end{abstract}

\maketitle

\clearpage

\section{Introduction}\label{sec:introduction}

\begin{quote}
\textsf{Judgment is the faculty of thinking the particular as contained under the universal.  
It involves the capacity to subsume under rules, that is, to distinguish whether something falls under a given rule.}  
\begin{flushright}
\footnotesize ------ Kant, \textit{Critique of Judgment}~\citep{kant1790cj}, \textit{Introduction IV, 5:179}; \textit{Critique of Pure Reason}~\citep{kant1781cpr}, \textit{A132/B171}.
\end{flushright}
\end{quote}

Recently, Large Language Models (LLMs) have achieved remarkable success across numerous domains~\cite{inno_new_world}, ranging from technical fields~\cite{inno_aigeo_zhao2024artificial,inno_medicine_tang2024llms,inno_life_science_yuan2023advanced} to the humanities~\cite{inno_cognitive_qu2024promoting,inno_life_science_luo2024artificial,inno_cancer_zhong2023artificial,inno_proteins} and social sciences~\cite{nature_personality_wang2025,inno_science_xu2023artificial,inno_paper_he2023chatgpt,inno_drug_shi2024drug}.  
This growing interest stems from LLMs' ability to mimic human-like reasoning and thinking processes, enabling them to take on roles traditionally reserved for human experts while offering a cost-effective solution that can be effortlessly scaled to meet increasing evaluation demands.
For instance, the use of LLM-as-a-Judge in academic peer review\footnote{\url{https://blog.iclr.cc/2024/10/09/iclr2025-assisting-reviewers/}}  offers a potential means to address the sharp growth in submissions while sustaining expert-level judgments.

Before the era of LLMs, finding a balance between comprehensive and scalable evaluation posed a persistent challenge. On the one hand, widely used subjective methods like expert-driven assessments~\cite {shi2024judging,human_like_summarization} integrate holistic reasoning and fine-grained contextual understanding, making them the gold standard in comprehensiveness. However, these approaches are costly, difficult to scale, and susceptible to inconsistency.
On the other hand, objective assessment methods, such as automatic metrics, offer strong scalability and consistency. For example, tools such as BLEU~\cite{papineni2002bleu} or ROUGE~\cite{lin2004rouge} can rapidly evaluate machine-generated translations or summaries against reference texts without human intervention. However, these metrics, which heavily rely on surface-level lexical overlaps, often fail to capture deeper nuances, resulting in poor performance in tasks like story generation or instructional texts~\cite{schluter-2017-limits}.
As a solution to this persistent dilemma, ``LLM-as-a-Judge'' has emerged as a promising idea to combine the strengths of the above two evaluation methods. Recent studies have shown that this idea can merge the scalability of automatic methods with the detailed, context-sensitive reasoning found in expert judgments~\cite{nips_llm_as_a_judge,wang2023pandalm,zhu2023judgelm,auto-j,data-juicer}.
Moreover, LLMs may become sufficiently flexible to handle multimodal inputs~\cite{mllm-as-a-judge} under appropriate prompt learning or fine-tuning~\cite{khattak2023maple}.
These advantages suggest that the LLM-as-a-Judge approach could \textbf{\textit{serve as a novel and broadly applicable paradigm}} for addressing complex and open-ended evaluation problems.

LLM-as-a-Judge holds significant potential as a scalable and adaptable evaluation framework compared to the aforementioned two traditional methods~\cite{self_taught}. However, its widespread adoption is hindered by two key challenges. The first challenge lies in the absence of a systematic review, which highlights the lack of formal definitions, fragmented understanding, and inconsistent usage practices in the relevant studies. As a result, researchers and practitioners struggle to fully understand and apply effectively. The second challenge concerns reliability~\cite{yu2024xfinder}, as merely employing LLM-as-a-Judge does not ensure accurate evaluations aligned with established standards. 
These challenges emphasize the need for a deeper assessment of the outputs generated by LLM-as-a-Judge, as well as a crucial investigation into the question: \textbf{\textit{How to build reliable LLM-as-a-Judge systems?}}

To address these challenges, this paper provides a systematic review of research on LLM-as-a-Judge. It offers a comprehensive overview of the field and explores strategies for building reliable LLM-as-a-Judge systems. We begin by defining LLM-as-a-Judge through both formal and informal definitions, answering the foundational question: \textit{"What is LLM-as-a-Judge?"} Next, we categorize existing methods and approaches, exploring \textit{"How to use LLM-as-a-Judge?"}.
Following this, to tackle the critical question: \textit{"How to build reliable LLM-as-a-Judge systems?"}, we explore two core aspects: (1) strategies to enhance the reliability of LLM-as-a-Judge systems and (2) methodologies for evaluating the reliability of these systems.
For the first aspect, we review key strategies to optimize the performance of LLM-as-a-Judge. For the second aspect, we examine the metrics, datasets, and methodologies used to evaluate LLM-as-a-Judge systems, highlighting potential sources of bias and methods for their mitigation.
Building on this, we introduce a novel benchmark specifically designed for evaluating LLM-as-a-Judge systems. Finally, we discuss future research directions, emphasizing key areas for improving reliability, scalability, and applicability. The contributions of this study can be summarized as follows:
\begin{enumerate}
    \item \textbf{At the definitional level}, we establish both formal and informal definitions of LLM-as-a-Judge, thereby delineating the conceptual boundaries of this emerging paradigm. We also introduce a contextualized definition of reliability, which incorporates input variability, model characteristics, and contextual dependencies, providing a principled foundation for theorizing and building reliable systems.
    
    \item \textbf{At the framework level}, we conduct a systematic reorganization of fragmented literature into a unified conceptual structure. Specifically, we map prior work to four foundational questions: what it is, how to use it, how to improve it, and how to evaluate it—framing reliability as the unifying thread across these dimensions.
    
    \item \textbf{At the empirical level}, we perform comparative analyses of existing approaches and further propose a meta-evaluation benchmark specifically tailored for evaluating LLM-as-a-Judge systems. This benchmark facilitates systematic reliability assessment, uncovering key trade-offs such as robustness versus sensitivity, and offering actionable insights for constructing trustworthy evaluation frameworks.
    
    \item \textbf{At the perspective level}, we offer a comprehensive analysis that integrates the applications, challenges, and future directions of LLM-as-a-Judge, providing a roadmap that extends beyond the scope of existing surveys. By systematically reviewing its applications in core machine learning and high-stakes domains, we identify domain-specific reliability requirements and underexplored challenges such as meta-evaluation and long-term consistency. Building on these findings, we articulate a forward-looking agenda that emphasizes theoretically grounded methodologies, systematic benchmarks, and hybrid human–AI frameworks for constructing reliable and socially trustworthy systems.

\end{enumerate}

\begin{figure*}[b]
  \raggedright
  \includegraphics[width=\textwidth]{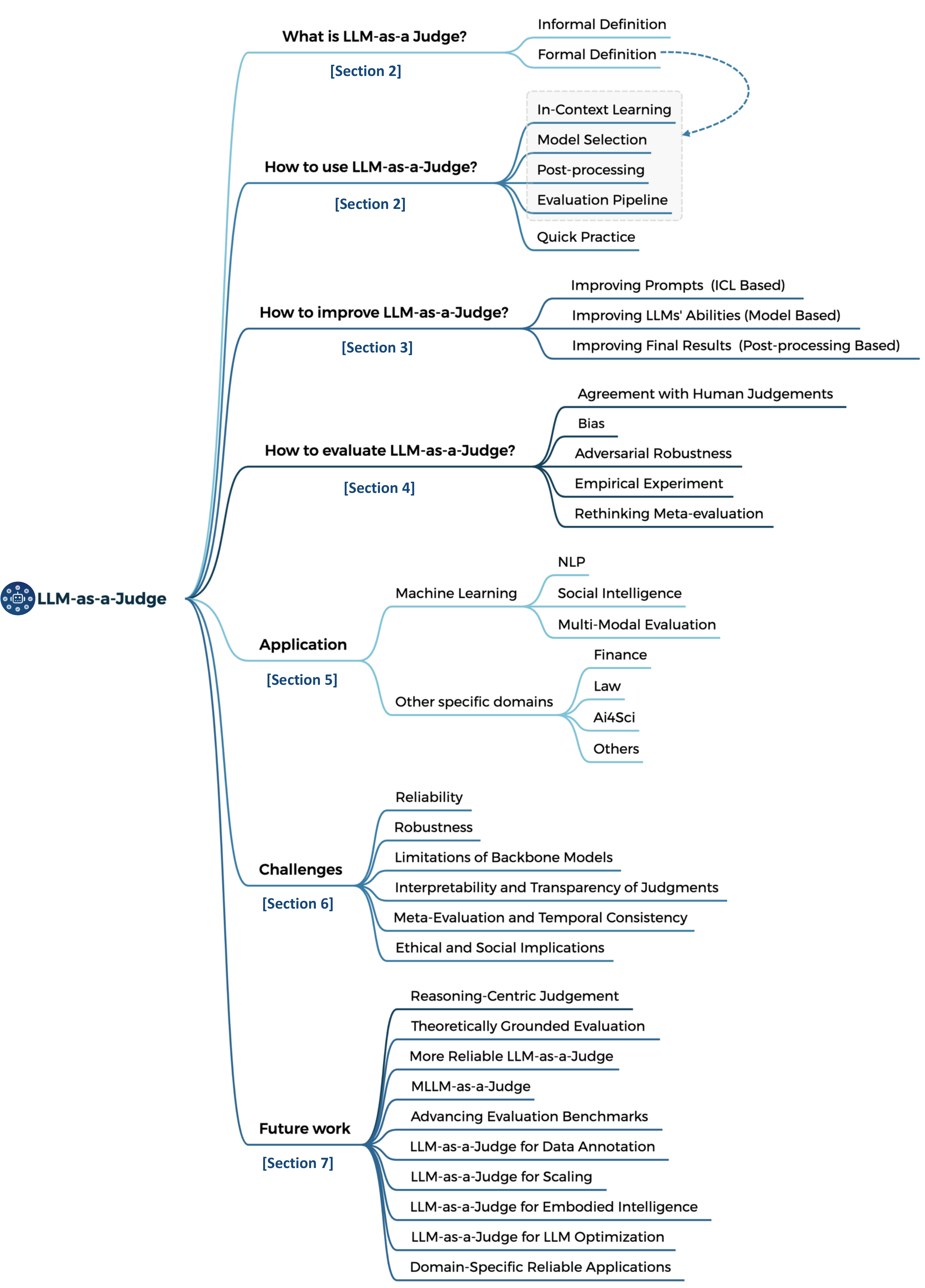}
  \caption{\centering The overall framework of this paper.} 
  \label{fig:fig1_llm-as-a-Judge_paper_structure}
\end{figure*}

The rest of this survey is organized as Figure~\ref{fig:fig1_llm-as-a-Judge_paper_structure}. Specifically, Section~\ref{sec:formulation} provides an overview of the LLM-as-a-Judge field, including its definitions and categorization of existing methods. For a quick guide on the implementation of an LLM as a judge for specific scenarios, you can find answers in Quick Practice~(\ref{quick_practice}). Strategies for enhancing and evaluating the reliability of LLM-as-a-Judge systems are discussed in Sections~\ref{sec:improvement} and ~\ref{sec:Evaluation} respectively. Notably, in Section~\ref{subsec:reasoning-centric}, we discuss the synergy between LLM-as-a-Judge and Reasoning-Centric enhancement, where dynamic feedback is used to optimize reasoning paths and significantly improve the model's ability to solve complex problems. Section~\ref{sec:applications} explores practical applications, while Sections~\ref{sec:challenges} and~\ref{sec:future_work} address challenges and outline future research directions. Finally, Section~\ref{sec:conclusion} presents our conclusions.

\addtocontents{toc}{\protect\setcounter{tocdepth}{-1}}

\clearpage  
\phantomsection  
\addcontentsline{toc}{section}{Table of Contents} 
\tableofcontents

\addtocontents{toc}{\protect\setcounter{tocdepth}{2}}




\vspace{-25pt}

\section{Background and Method}\label{sec:formulation}

The capacity of LLMs to emulate human reasoning and evaluate specific inputs against a set of predefined rules has paved the way for "LLM-as-a-Judge." Existing studies indicate that LLM's scalability, adaptability, and cost-effectiveness make them well-suited for a growing number of evaluative tasks that were traditionally done by humans. These abilities are key in utilizing LLMs flexibly across various evaluation scenarios and objectives. As a result, the adoption of LLM in evaluation has progressed rapidly in practice. Initially, the primary focus of LLMs was on language generation and comprehension. With advancements in training paradigms like Reinforcement Learning from Human Feedback (RLHF)~\cite{ouyang2022training}, LLMs became increasingly aligned with human values and reasoning processes. This alignment has allowed LLMs to transition from generative tasks to evaluation. At its core, LLM-as-a-Judge denotes the use of LLMs to evaluate objects, actions, or decisions based on predefined rules, criteria, or preferences. It encompasses a broad spectrum of roles, including:
\textbf{Graders}~\citep{trung2024reft,raft},  
\textbf{Evaluators/Assessors}~\citep{salad_bench,zhang2024evaluating},  
\textbf{Critics}~\citep{ke2024critiquellm,xiong2024llava,agentq},  
\textbf{Verifiers}~\citep{deductive_verification,nips_reflexion,speculative_rag},  
\textbf{Examiners}~\citep{lm-as-an-examiner},  
\textbf{Reward/Ranking Models}~\citep{qwen-math-reward-model,llava_rlhf,wizardmath,yuan2023rrhf}, etc.

Currently, the definition of how to effectively use LLM-as-a-Judge for evaluation tasks is largely informal or vague, lacking a clear and formal expression. Therefore, we will start with a formal definition of LLM-as-a-Judge as follows:

\vspace{-3pt}

\begin{equation*}
\mathbox[eval]{\mathcal{E}}\mathbox[post-pro]{\leftarrow} \mathbox[model]{\mathcal{P}_{\mathcal{LLM}}}\mathbox[icl]{\left( x \oplus \mathcal{C} \right)}
\end{equation*}

\begin{itemize}
    \item $\mathcal{E}$: The final evaluation obtained from the whole LLM-as-a-Judge process in the expected manner. It could be a score, a choice, a label or a sentence, etc.
    \item $\mathcal{P}_{\mathcal{LLM}}$: The probability function defined by the corresponding LLM, and the generation is an auto-regressive process.
    \item $x$: The input data in any available types~(text, image, video), which waiting to be evaluated.
    \item $\mathcal{C}$: The context for the input $x$, which is often prompt template or combined with history information in dialogue.
    \item $\oplus$: The combination operator combines the input $x$ with the context $\mathcal{C}$, and this operation can vary depending on the context, such as being placed at the beginning, middle, or end.
\end{itemize}

\begin{figure}[t]
  \includegraphics[width=0.9\columnwidth]{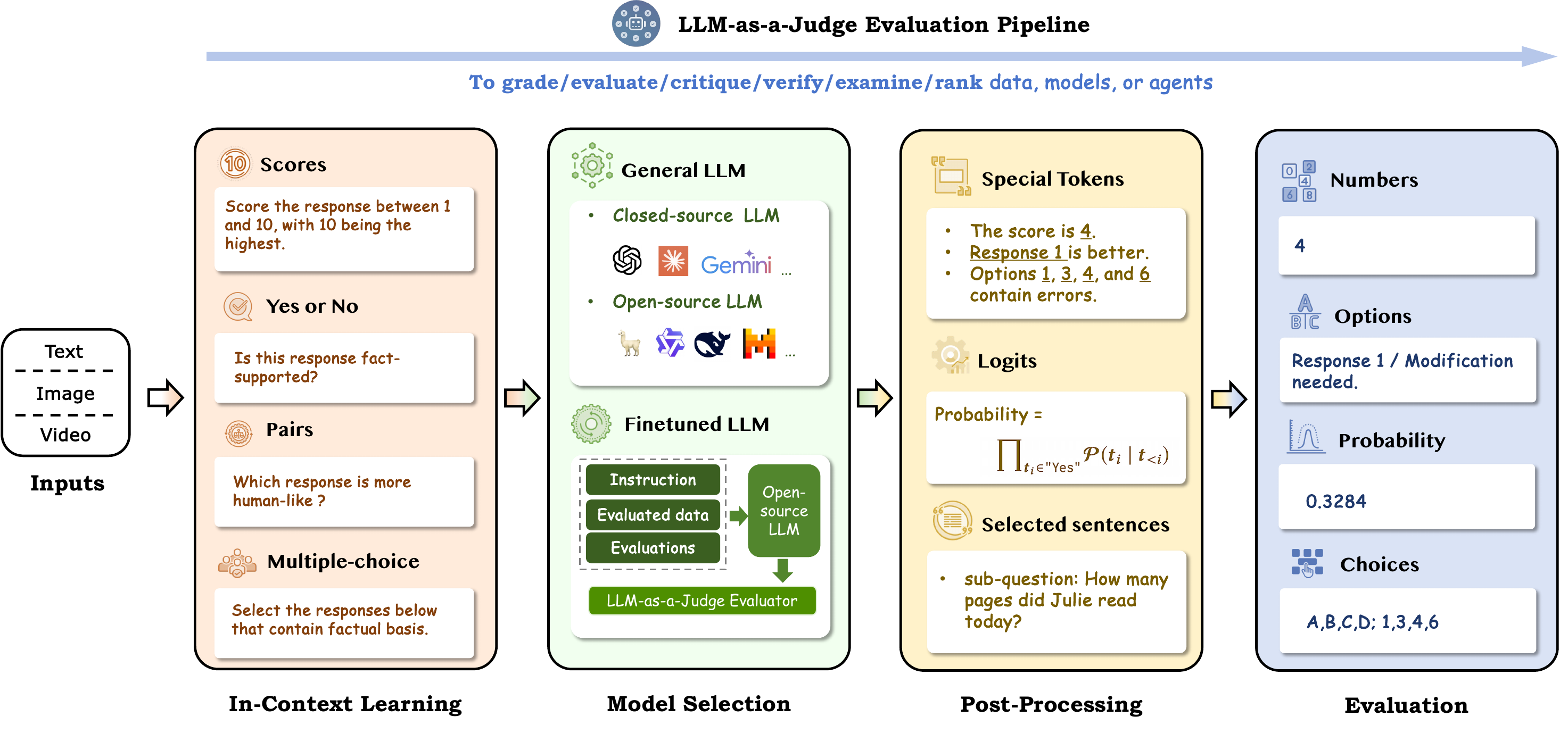}
  \caption{LLM-as-a-Judge evaluation pipelines.}
  \label{fig:fig2_evaluation_pipeline}
\end{figure}

The formulation of LLM-as-a-Judge reflects that LLM is a type of auto-regressive generative model, which generates subsequent content based on the context to obtain target evaluation. It illustrates how we utilize LLM for evaluation tasks, encompassing input design, model selection, and training, as well as output post-processing. 
The basic approaches of implementing LLM-as-a-Judge can be classified by the formulation: In-Context Learning, Model Selection, Post-processing Method, and Evaluation Pipeline in Figure~\ref{fig:fig2_evaluation_pipeline}. By following this pipeline, one can build a basic LLM-as-a-Judge for evaluation. A quick practice guide is available in section~\ref{quick_practice}.
However, the basic definition alone does not guarantee the reliability of evaluations. To explicitly highlight and address reliability, we further propose the following enhanced formal definition:
\vspace{-2pt}
\[
\Large
\mathcal{R} \leftarrow f_{\text{R}}\left(\mathcal{P}_{\mathcal{LLM}}, x, \mathcal{C}\right)
\]
\vspace{-10pt}

\begin{itemize}
\item $\mathcal{R}_{\text{}}$: The evaluation explicitly designed to ensure consistency, robustness, and alignment with human judgment. This reliability is verified through additional validation, calibration, and standardization steps beyond the basic pipeline.
\item $f_{\text{R}}$: A series of constraints and validation methods applied systematically to the basic LLM-as-a-Judge framework to enhance evaluation reliability. These include methods to mitigate biases, control variability, and confirm robustness against adversarial inputs.
\end{itemize}

\subsection[In-Context Learning]{$\mathbox[icl]{\text{In-Context Learning}}$}
\label{formulation:icl}

To apply LLM-as-a-Judge, evaluation tasks are typically specified using In-Context Learning methods, which provide instructions and examples to guide the model's reasoning and judgment. This process involves two key aspects: input design and prompt design. For input design, it is important to consider the type of variables to be evaluated (such as text, image, or video), the manner of input (e.g., individually, in pairs, or in batches), and its position (e.g., at the beginning, middle, or end). For the prompt design, four different methods can be adopted, as illustrated in Figure~\ref{fig:fig2_evaluation_pipeline}. These methods include generating scores, solving true/false questions, conducting pairwise comparisons, and making multiple-choice selections. Further details will be presented in the following sections.

\subsubsection{\textbf{Generating scores}}
\begin{figure*}[t]
  \centering
  \includegraphics[width=0.75
  \columnwidth]{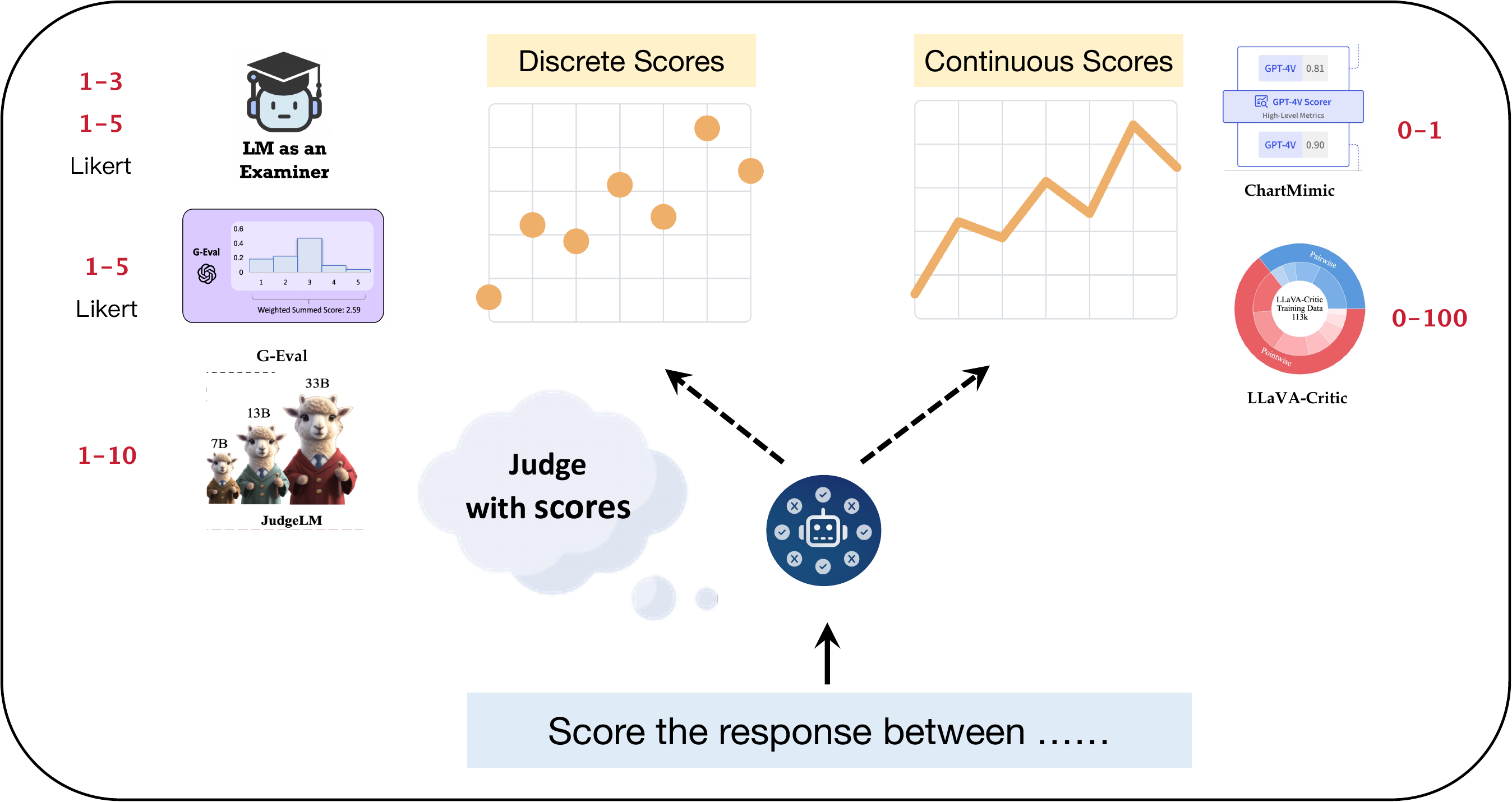}
  \caption{The illustrations of method \textit{generating scores} in ICL}
  \label{fig:illustration_of_generating_scores}
\end{figure*}

It is quite intuitive to represent an evaluation using a corresponding score, shown in Figure~\ref{fig:illustration_of_generating_scores}. What requires more careful consideration, however, is the nature and range of the score used for evaluation. The score can be discrete, with common ranges like 1-3, 1-5~\citep{multi-aspect_framework}, or 1-10~\citep{zhu2023judgelm,auto-j}. Alternatively, it can be continuous, ranging from 0 to 1 or 0 to 100~\citep{xiong2024llava}.
The simplest way to score is through the context, setting the range of scores and the main criteria for scoring. For example, "Please rate the helpfulness, relevance, accuracy, level of details of their responses. Each assistant receives an overall score on a scale of 1 to 10, where a higher score indicates better overall performance"~\citep{zhu2023judgelm}. A slightly more complex way is to provide more detailed scoring criteria. More complex scoring situations can be as \textit{Language-Model-as-an-Examiner}~\citep{lm-as-an-examiner}, which use Likert scale scoring functions as an absolute evaluative measure. The evaluator assigns scores to a given response along predefined dimensions, including accuracy, coherence, factuality, and comprehensiveness. Each of these dimensions is scored on a scale of 1 to 3, ranging from worst to best. The evaluator is also asked to provide an overall score ranging from 1 to 5, based on the scores assigned to the previous 4 dimensions. This score serves as an indicator of the overall quality of the answer.

\begin{tcolorbox}[colback=gray!10, colframe=gray!90!black, title={Evaluation Prompt Templates from \citet{human_like_summarization}}]
\textbf{Likert Scale Scoring:}\\
Evaluate the quality of summaries written for a news article. Rate each summary on four dimensions: \textcolor{blue}{\{Dimension\_1\}}, \textcolor{blue}{\{Dimension\_2\}}, \textcolor{blue}{\{Dimension\_3\}}, and \textcolor{blue}{\{Dimension\_4\}}. You should rate on a scale from 1 (worst) to 5 (best).\\
Article: \textcolor{blue}{\{Article\}} \\
Summary: \textcolor{blue}{\{Summary\}}

\vspace{6pt}
\textbf{Pairwise Comparison:}\\
Given a new article, which summary is better? Answer "Summary 0" or "Summary 1". You do not need to explain the reason.\\
Article: \textcolor{blue}{\{Article\}} \\
Summary 0: \textcolor{blue}{\{Summary\_0\}} \\
Summary 1: \textcolor{blue}{\{Summary\_1\}}
\end{tcolorbox}

\subsubsection{\textbf{Solving Yes/No questions}}

A Yes/No question requires a judgment on a given statement, focusing solely on its accuracy. This type of question is simple and direct, providing only two fixed responses—yes or no, true or false—without any additional comparisons or choices.


This type of evaluation is often utilized in intermediate processes, creating the conditions for a feedback loop. For example, it promotes a self-optimization cycle, as seen in \textit{Reflexion} \citep{nips_reflexion}, which generates verbal self-reflections to provide valuable feedback for future attempts. In scenarios with sparse reward signals, such as a binary success status (success/fail), the self-reflection model uses the current trajectory and persistent memory to generate nuanced and specific feedback.
Similarly, in self-improvement contexts \citep{ucla_creativity_macgyver}, Yes/No questions can be employed to evaluate custom phrases, such as \texttt{"Modification needed."} and \texttt{"No modification needed."}, facilitating entry into the next cycle.
Moreover, these evaluations are common for testing knowledge accuracy and assessing whether statements align with established facts \citep{think_on_graph}, like "Given a question and the associated retrieved knowledge graph triples (entity, relation, entity), you are asked to answer whether it's sufficient for you to answer the question with these triples and your knowledge (Yes or No)." A detailed and specific example can be seen in the Figure~\ref{fig:yes_no_pair_method}.

    

    

\begin{tcolorbox}[colback=gray!10, colframe=gray!90!black, title={Evaluation Prompt Templates for Yes/No and Multiple-Choice Tasks}]
\textbf{Yes/No Evaluation:}\\
Is the sentence supported by the article? Answer "Yes" or "No".\\
Article: \textcolor{blue}{\{Article\}} \\
Sentence: \textcolor{blue}{\{Sentence\}}

\vspace{6pt}
\textbf{Multiple-Choice Evaluation:}\\
You are given a summary and some semantic content units. For each semantic unit, choose those can be inferred from the summary, return their number.\\
Summary: \textcolor{blue}{\{Summary\}} \\
Semantic content units: \\
1. \textcolor{blue}{\{SCU\_1\}} \\
2. \textcolor{blue}{\{SCU\_2\}} \\
...... \\
n. \textcolor{blue}{\{SCU\_n\}}
\end{tcolorbox}

\subsubsection{\textbf{Conducting pairwise comparisons}}

\begin{figure*}[h]
  \includegraphics[width=0.9\columnwidth]{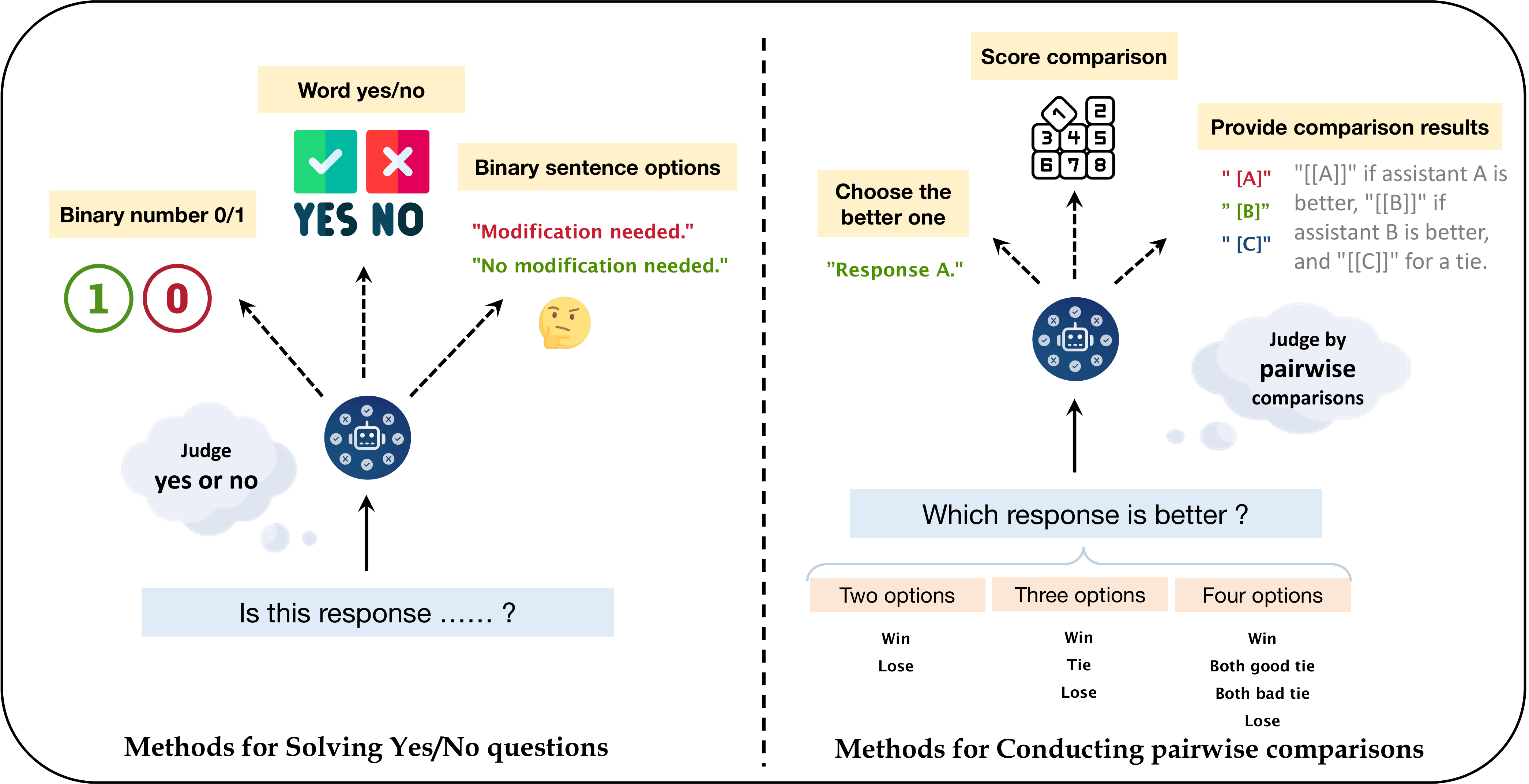}
  \caption{The illustrations of method \textit{Solving Yes/No questions} and \textit{Conducting pairwise comparisons} in ICL}
  \label{fig:yes_no_pair_method}
\end{figure*}

Pairwise comparison refers to comparing two options and selecting which one is superior or more aligned with a specific standard, showed in Figure~\ref{fig:yes_no_pair_method}. It involves making a decision between two options rather than judgement between 'yes' or 'no'. The comparison can be subjective or based on objective criteria. This evaluation is a relative evaluation. Pairwise comparison is often used for ranking multiple options or prioritizing them, where several comparisons are made between pairs to identify the better choice or establish a hierarchy.


Pairwise comparison is a well-established method that has significantly impacted a variety of fields \citep{naacl_text_ranker}. As noted by \cite{liu2024aligning}, LLM and human evaluations are more aligned in the context of pairwise comparisons compared to score-based assessments. Numerous studies have demonstrated that pairwise comparative assessments outperform other judging methods in terms of positional consistency \citep{nips_llm_as_a_judge, acl_comparative_assessment}. Furthermore, pairwise comparisons can be extended to more complex relation-based assessment frameworks, such as list-wise comparisons, using advanced ranking algorithms \citep{naacl_text_ranker, liu2024aligning}, data filtering~\cite{yuan2023rrhf}. In pairwise comparative assessments, LLM-as-a-Judge is prompted to select the response that better answers the question at hand. To accommodate the possibility of a tie, several option modes are introduced. The Two-Option mode requires judges to choose the better response from two given options. The Three-Option mode introduces an additional choice, allowing judges to indicate a tie if neither response is preferable, as shown in Figure~\ref{fig:yes_no_pair_method}. Evaluations typically involve determining the outcomes of win, tie, or loss for responses \cite{wang2023pandalm} through pairwise comparisons, with win rounds counted for each response. The Four-Option mode further expands the choices, allowing judges to classify responses as either a "both good tie" or a "both bad tie."

    
    
    
    





\subsubsection{\textbf{Making multiple-choice selections}}
Multiple-choice selections involve providing several options, not giving relative choices in pairwise comparison, nor making a yes/no judgment.
The evaluator must choose the most appropriate or correct one. This method allows for a broader range of responses compared to true/false questions and can assess deeper understanding or preferences. However, this kind of prompt design is more rare than the first three.

    


    











\begin{tcolorbox}[colback=white!5, colframe=black, title={Reliability Concerns of In-Context Learning}]
When leveraging in-context learning, certain issues can surface, potentially impacting the reliability of evaluations. These include the variability of LLM outputs due to minor prompt changes, which can lead to unstable results. Furthermore, score-based assessments often exhibit inconsistent inter-rater reliability, influenced by the inherent randomness of LLM generation and its sensitivity to phrasing. Similarly, evaluation formats like Yes/No or multiple-choice questions are prone to ambiguity in response interpretation. Lastly, LLM-as-a-Judge evaluations may inadvertently reflect biases, such as favoring responses based on their position or length. 
\end{tcolorbox}


\subsection[Model Selection]{$\mathbox[model]{\text{Model Selection}}$}
\label{formulation:model_selection}

\subsubsection{\textbf{General LLM}}
To automate evaluation by LLM-as-a-Judge, one effective approach is to employ advanced language models such as GPT-4~\citep{openai2023gpt4} instead of human evaluators~\cite{nips_llm_as_a_judge}. For instance, \citet{alpaca_eval} created a test set with 805 questions and assessed the performance by comparing it to text-davinci-003 using GPT-4. Additionally, \citet{nips_llm_as_a_judge} designed 80 multi-round test questions across eight common areas and used GPT-4 to automatically score the model's responses. The accuracy of the GPT-4-based evaluator has been demonstrated to be high compared to professional human evaluators, showing superior consistency and stability in evaluations. At the same time, if the general LLM used has limitations in instruction-following or reasoning abilities, the effectiveness of the LLM-as-a-Judge method may be significantly affected.

\subsubsection{\textbf{Fine-tuned LLM}}


However, relying on external API for evaluation may introduce consideration about privacy leakage, and the opacity of API models also challenges the evaluation reproducibility. Therefore, subsequent studies recommend refining language models tailored for evaluations by emphasizing the use of pairwise comparisons or grading.
For instance, PandaLM \cite{wang2023pandalm} constructs data based on Alpaca instructions and GPT-3.5 annotation, and then fine-tunes LLaMA-7B \cite{touvron2023llama} as an evaluator model. JudgeLM \cite{zhu2023judgelm} constructs data from diversified instruction sets and GPT-4 annotations, and fine-tunes Vicuna \cite{vicuna2023} as a scalable evaluator model. Auto-J \cite{auto-j} constructs evaluation data upon multiple scenarios to train a generative evaluator model, which can provide both evaluation and critical opinion. Prometheus \cite{prometheus} defines thousands of evaluation criteria and constructs a feedback dataset based on GPT-4, and fine-tunes a fine-grained evaluator model. 
The typical process for fine-tuning a judge model involves three main steps.
\noindent\textbf{Step 1: Data Collection.} The training data generally consists of three components: instructions, the objects to be evaluated, and evaluations. Instructions are typically sourced from instruction datasets, while evaluations can come from either GPT-4 or human annotations.
\noindent\textbf{Step 2-Prompt Design.} The structure of the prompt template can vary based on the evaluation scheme, which already detailed in \S~\ref{formulation:icl}.
\noindent\textbf{Step 3: Model Fine-Tuning.} Using the designed prompts and collected data, the fine-tuning process for the evaluator model typically adheres to the instruction fine-tuning paradigm \cite{ouyang2022training}. The model receives an instruction along with one or more responses to generate output that includes evaluation results and possibly explanations.

After fine-tuning, the evaluator model can be employed to evaluate the target object. While these fine-tuned models often demonstrate superior performance on self-designed test sets, they are identified several limitations in their evaluation capabilities,which detailed in Section~\ref{Evaluation:Bias}.  The current prompt and fine-tuning dataset designs often result in evaluation LLMs with poor generalization, making them difficult to compare with strong LLMs like GPT-4.

\begin{tcolorbox}[colback=white!5, colframe=black, title={Reliability Concerns of Model Selection}]
The choice of model significantly impacts the dependability of LLM-as-a-Judge systems. Concerns arise from the black-box nature and version dependency of general-purpose LLMs, which can hinder the reproducibility of evaluation outputs. Fine-tuned evaluators, while specialized, often exhibit overfitting and limited generalization beyond their training data. Moreover, these models can inherit subtle biases from their training datasets, necessitating careful meta-evaluation to ensure fairness. Finally, reliance on smaller open-source models, while cost-effective, may introduce inconsistencies and misalignment with human judgments.
\end{tcolorbox}

\subsection[Post-processing]{$\mathbox[post-pro]{\text{Post-processing}}$}
\label{formulation:post_processing}
Post-processing refines the probability distributions generated by LLM-as-a-Judge to ensure accurate evaluations. The evaluation format should align with our In-Context Learning design and may involve procedures to enhance the reliability of extracted evaluations, which should be applied consistently. We focus on three main post-processing methods: extracting specific tokens, normalizing the output logits, and selecting sentences with high returns. 
However, it is important to note that each method has significant limitations when evaluating objective questions. For example, in text response evaluation~\citep{yu2024xfinder}, failing to accurately extract the key answer token from the LLM's response can result in incorrect evaluation outcomes. These challenges in post-processing are tightly linked to the prompt design used in earlier ICL stages and the selected model's ability to follow instructions reliably.


\subsubsection{\textbf{Extracting specific tokens}}

As showed in In-context Learning~(Section~\ref{formulation:icl}), when the evaluation target take the form of a score, selecting specific options, or responding with Yes/No, applying rule-match to extract the corresponding token from the response generated during probability distribution iteration is common used. It is worth noting that Yes/No is a broad definition, including custom statements involving judgment.
Considering a Yes/No question for evaluation in custom phrases~\cite{ucla_creativity_macgyver}: \texttt{"Modification needed."} and \texttt{"No modification needed."} or a yes-no question \texttt{"Does the above answer need to be further modified?"}. 
When the input sample is put through the template, it might have outputs such as "Modification needed.", "Conclusion: Modification needed." or "Yes". This variance in response formats is difficult to parse consistently. The corresponding post-processing with the response is necessary. Using rules to extract specific tokens for our designed prompts and input content, as well as the backbone model used for the evaluator, all have higher requirements as we discussed in Section~\ref{formulation:model_selection}.
In contextual learning, if there is no clear indication of the output format for response, there may be various expressions of evaluation, which can be seen in Figure~\ref{fig:fig2_evaluation_pipeline}. For example, "Response 1 is better" and "The better one is response 1", which convey the same choice but differ in format leading to the difficulty of rule recognition. Simple solutions often involve providing clear instructions, such as "The last sentence should be started with 'The better response is'", or using a few-shot strategy.
Also, the general model with insufficient instruction following capability may not be able to generate the evaluation format and content of the target according to the instruction, resulting in the post-processing extracted according to the rules not as smooth as expected. 

Constrained decoding is a technique that enforces structured output from Large Language Models (LLMs) by restricting token generation according to predefined schemas, typically in formats like JSON. This approach uses a finite state machine (FSM) to compute valid next tokens at each decoding step, effectively masking the model's output probability distribution to ensure conformity with the desired schema. While this method guarantees syntactically valid outputs, it presents several challenges: it can distort the model's learned distribution and potentially degrade output quality, requires significant engineering implementation effort, and introduces computational overhead during inference.

Recent work has proposed various solutions to address these challenges. \cite{beurer-kellner_guiding_2024} introduces DOMINO, a decoding algorithm that preserves natural tokenization while enforcing constraints. Their system minimizes overhead through precomputation and speculative decoding, sometimes achieving faster performance than unconstrained decoding. \cite{dong_xgrammar_2024} develops XGrammar, which accelerates grammar-constrained generation by separating tokens into those that can be pre-checked and those requiring runtime verification. By co-designing the grammar engine with LLM inference, they achieve up to 100x speedup over existing approaches.\cite{zheng_sglang_2024} presents SGLang, combining a domain-specific language with an optimized runtime. Their system features efficient KV cache reuse and compressed finite state machines for faster decoding, demonstrating that thoughtful co-design of programming model and runtime can minimize constrained decoding overhead.

\subsubsection{\textbf{Normalizing the output logits}}
LLM-as-a-Judge in the intermediate steps with Yes/No setting often normalizes the output logits to obtain the evaluation in the form of a continuous decimal between 0 and 1. This is also very common in agent methods and prompt-based optimization methods~\cite{hao2023reasoning, zhuang2023toolchain, speculative_rag}. For example, the self-consistency and self-reflection scores~\citep{speculative_rag} within one forward pass of $\mathcal{M}_\text{Evaluator}$, are effectively obtained by constructing a prompt $[\left( x \oplus \mathcal{C} \right), \texttt{"Yes"}]$ and acquire the probability of each token conditioned on the previous tokens $P(t_i | t_{<i})$. The auto-regressive feature is leveraged, thus aggregate the probability of the relevant tokens to compute the self-consistent score $\rho_\text{Self-consistency}$ and self-reflection score $\rho_\text{Self-reflection}$. The final score is produced by $\rho_j = \rho_{\text{SC},j} \cdot \rho_{\text{SR},j}$.
\begin{align*}
    \underrightarrow{ \overbrace{\left( x \oplus \mathcal{C} \right)}^{\rho_\text{SC}} 
    \overbrace{\texttt{"Yes"}}^{\rho_\text{SR}}} \  \Rightarrow
    \begin{cases}
    \rho_\text{SC} = \prod_{t_i\in \alpha} P(t_i|t_{<i}) \cdot \prod_{t_i\in \beta} P(t_i|t_{<i})\\ \rho_\text{SR} = \prod_{t_i\in \texttt{"Yes"}} P(t_i|t_{<i})
    \end{cases}
\end{align*}
In addition, Self-evaluation~\citep{hao2023reasoning} is also common using this method for LLM-as-a-Judge. It can be helpful to let the LLM evaluate itself by asking, "Is this reasoning step correct?" and then reward it based on the probability of the next word being "Yes."

\subsubsection{\textbf{Selecting sentences}}

In addition to selecting specific tokens and normalizing the output logits, the content extracted by LLM-as-a-Judge may also be a sentence or paragraph. As showed in Figure~\ref{fig:fig2_evaluation_pipeline}, agent for reasoning task~\citep{hao2023reasoning}, builds a
reasoning tree by iteratively considering the most promising reasoning steps (actions, sub-questions) by LLM-as-a-Judge.


\begin{tcolorbox}[colback=white!5, colframe=black, title={Reliability Concerns of Post-processing Method}]
The post-processing steps applied to LLM outputs introduce their own set of reliability challenges. Rule-based token extraction methods are inherently brittle and susceptible to minor variations in responses, potentially leading to silent errors. While logit-based normalization offers probabilistic scoring, its effectiveness is highly dependent on precise prompt design and consistent model tokenization, where inconsistencies can introduce noise. Furthermore, methods focusing on sentence-level evaluation or response selection risk propagating biases if the scoring LLM is overly sensitive to stylistic rather than substantive cues. Ultimately, all post-processing strategies remain vulnerable to adversarial manipulations designed to inflate evaluation scores without genuine content improvement.
\end{tcolorbox}


\subsection[Evaluation Pipeline]{$\mathbox[eval]{\text{Evaluation Pipeline}}$}\label{subsec:evaluation_pipeline}
After completing the three processes, we obtain the final evaluation $\mathcal{E}$. From input to output, these steps collectively constitute the LLM-as-a-Judge evaluation pipeline, as illustrated in Figure~\ref{fig:fig2_evaluation_pipeline}. This pipeline is commonly applied in four scenarios: LLM-as-a-Judge for models, LLM-as-a-Judge for data, LLM-as-a-Judge for agents, and LLM-as-a-Judge for reasoning or thinking.

\begin{figure*}[h]
  \centering
  \includegraphics[width=0.8\textwidth]{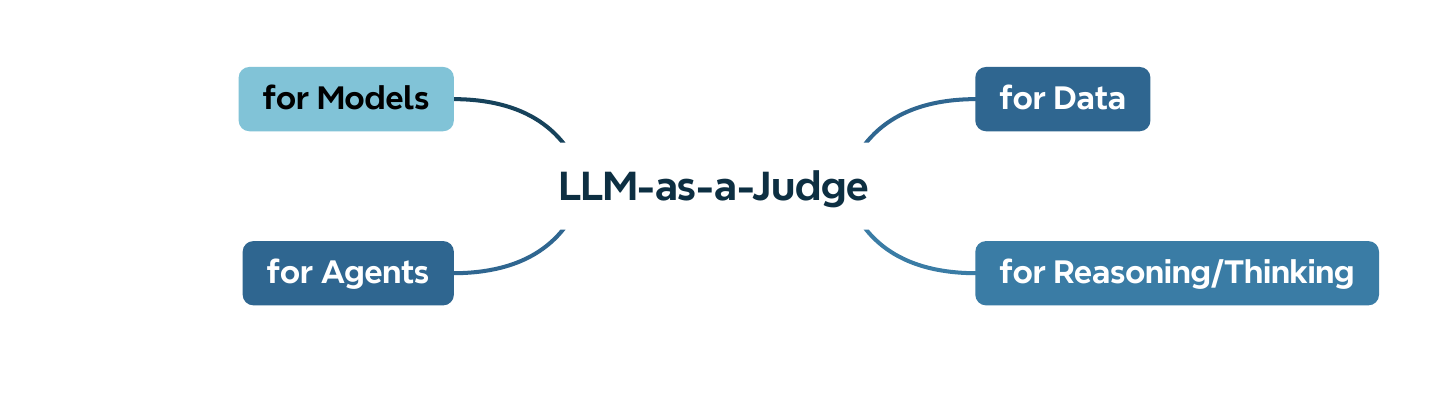}
  \caption{\centering Four typical scenarios using LLM-as-a-Judge evaluation pipeline.} 
  \label{fig:four_pipelines}
\end{figure*}

\begin{figure*}[h]
  \centering
  \includegraphics[width=0.85\textwidth]{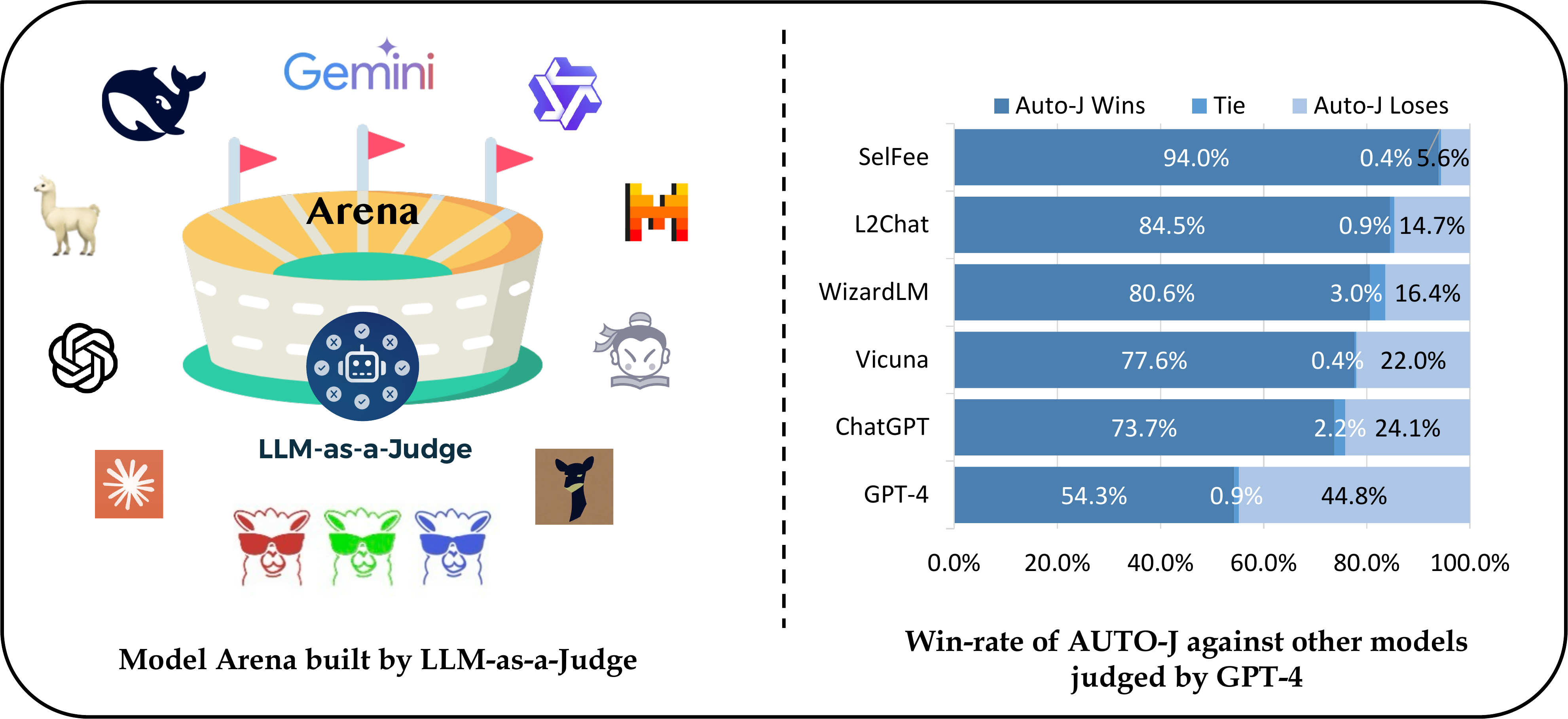}
  \caption{The illustrations of the scenario \textit{LLM-as-a-Judge for Models}. The example of "win-tie-lose" is from \citet{auto-j}}
  \label{fig:judge_for_model}
\end{figure*}

\subsubsection{\textbf{LLM-as-a-Judge for Models}} 

It is universally known that the best way to evaluate LLMs is human judgment, but collecting human annotations can be costly, time-consuming, and laborious \citep{ouyang2022training,nips_llm_as_a_judge}. Using strong LLMs (usually closed-source ones, e.g., GPT-4, Claude, ChatGPT) as an automated proxy for assessing LLMs has become a natural choice \citep{zhou2023lima}, as shown in Figure~\ref{fig:fig2_evaluation_pipeline}. With appropriate prompt design, the quality of evaluation and agreement to human judgment can be promising \citep{dubois2023alpacafarm,nips_llm_as_a_judge,zhang2023wider,wang-etal-2024-large-language-models-fair}. 
However, the cost concern still exists when calling the APIs of these proprietary models, especially when there is a frequent need for model validation on large-scale data. Moreover, closed-source LLM-as-a-Judge leads to low reproducibility due to potential changes in models behind the API.
Some recent works have started to make attempts for open-source alternatives. SelFee \citep{selfee2023} collects generations, feedback, and revised generations from ChatGPT and fine-tunes LLaMA models to build a critique model. Shepherd \citep{wang2023shepherd} trains a model that can output critiques for single-response with the data of feedback from online communities and human annotation. PandaLM \citep{wang2023pandalm} trains a model to conduct pairwise comparison for LLM Instruction Tuning Optimization, and \citet{nips_llm_as_a_judge} also fine-tune Vicuna \citep{vicuna2023} on a 20K pairwise comparison dataset to explore the potential of open-source models as a more cost-friendly proxy.

\subsubsection{\textbf{LLM-as-a-Judge for Data}} 
\label{Evaluator for data}
Data annotation generally refers to the labeling or generating of raw data with relevant information, which could be used for improving the efficacy of machine learning models. The process, however, is labor-intensive and costly. The emergence of LLMs presents an unprecedented opportunity to automate the complicated process of data annotation by LLM-as-a-Judge.
Most of the data need to be evaluated by LLM-as-a-Judge is generated by models, or large-scale crawled data.
Language models first conduct supervised fine-tuning to imitate how to align with human instructions \cite{selfinstruct,alpaca}.
After that, reinforcement learning techniques have been explored to align language models with human preferences \cite{ouyang2022training,rl4lms}.
The most successful way is applying a RLHF framework \cite{ouyang2022training} via training a reward model on human feedback and using PPO \cite{ppo} to obtain the policy model for language generation. 
However, in practices, the PPO training paradigm is complex in coding and hyper-parameter tuning while it needs four models that are hard for training. 
This motivates us to explore simpler and more straightforward methods to align language models with human preferences. This involves how to use LLM-as-a-Judge to evaluate whether different responses are aligned with human preferences. 
For example, 
~\citep{yuan2023rrhf,raft} use general LLM~(ChatGPT) to get better alignment with human preferences. The Aplaca prompts \cite{alpaca} is used as sampling queries to different models generate responses. And these data was evaluated by LLM-as-a-Judge to obtain human preference scores~(reward score) to train a new language model. Other works would like to use Supervised Fine-Tuning~(SFT) model itself as evaluator, like generating better-aligned datasets for SFT including hindsight-modified prompts \cite{zhang2023wisdom,liu2023languages} and principle-driven self-alignment \cite{sun2023principle}.

\begin{figure*}[t]
  \includegraphics[width=0.85\columnwidth]{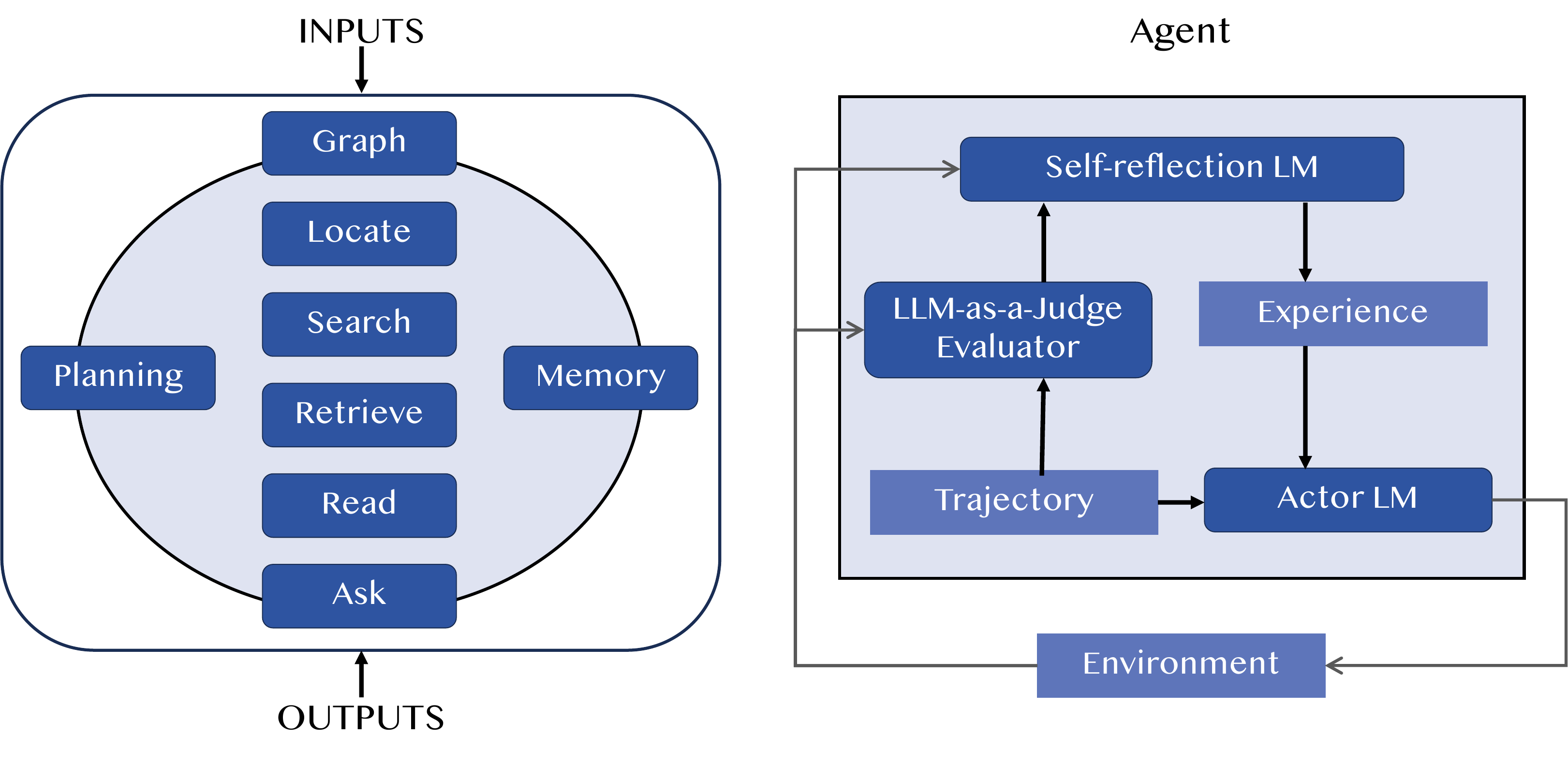}
  \caption{LLM-as-a-Judge appears in two common forms in the agent. The left diagram is Agent-as-a-Juge, designing a complete agent to serve as an evaluator. The right diagram shows using LLM-as-a-Judge in the process of an Agent.}
  \label{fig:fig3_agent_judge}
\end{figure*}

In addition, the lack of domain-specific model training data is a common phenomenon. In order to obtain annotated high-quality data, it is also very common to use LLM-as-a-Judge for the generation and evaluation of domain data. \textit{WizardMath}~\cite{wizardmath} would use its Instruction Reward Model (IRM) as Evaluator, aiming to judge the quality of the evolved instructions on three aspects: i) Definition, ii) Precision, and iti) Integrity. To produce the ranking list training data of IRM, for each instruction, ChatGPT and Wizard-E are used to generate 2-4 evolved instructions respectively. Then we leverage Wizard-E to rank the quality of those 4-8 instructions.

However, solely relying on LLM-as-a-Judge for data annotation poses challenges, particularly as the value of annotated data diminishes with the rapid improvement of model performance. To address this, approaches like Self-Taught Evaluator~\citep{self_taught} offer a promising alternative by eliminating the need for human annotations. This method leverages synthetic training data, starting with unlabeled instructions and generating contrasting outputs from models. These outputs are then used to train an LLM-as-a-Judge to produce reasoning traces and final judgments. With each iteration, the evaluator improves by learning from its refined predictions, creating a cycle of continuous self-enhancement. This iterative approach not only keeps annotations relevant but also ensures that evaluators evolve alongside advancing models.



Recent research on evaluating multimodal data focuses on addressing vision-language misalignments in Multimodal Large Language Models (MLLMs), which often cause hallucinations—outputs inconsistent with visual or contextual evidence~\citep{li2023evaluating, wang2023evaluation, cui2023holistic}. Techniques like RLHF and Factually Augmented RLHF have been employed to improve model alignment by incorporating structured ground-truth data and image captions, enhancing hallucination detection~\citep{llava_rlhf}. Benchmarks such as MLLM-as-a-Judge~\citep{mllm-as-a-judge} assess these models using tasks like scoring, pair comparison, and batch ranking, revealing limitations in alignment with human preferences. Persistent issues include biases (e.g., position, verbosity) and hallucinations, with even advanced models like GPT-4V displaying challenges. While pair comparison tasks align better with human judgment, scoring and batch ranking require significant improvements for reliable deployment. These findings emphasize the need for innovative frameworks and datasets to refine MLLM evaluation and alignment.

\subsubsection{\textbf{LLM-as-a-Judge for Agents}}

There are two ways to apply LLM-as-a-Judge for an agent. One is to evaluate the entire process of the intelligent agent~\citep{aaaj}, and the other is to evaluate it at a specific stage in the agent framework process~\citep{hao2023reasoning,nips_reflexion}. Both approaches are briefly illustrated in Figure~\ref{fig:fig3_agent_judge}.
Using LLM as the brain of agent, an agentic system~\citep{aaaj} could evaluate like a human, it would reduce the need for human involvement and eliminate the trade-off between thoroughness and effort. In addition, the agent~\citep{nips_reflexion} can interact with the environment through language and receive feedback on actions through LLM to make decisions for the next action.

\subsubsection{\textbf{LLM-as-a-Judge for Reasoning/Thinking}}

Reasoning~\cite{reasoning_survey}, defined as the cognitive process of applying logic, arguments, and evidence to draw conclusions, is central to intellectual tasks such as decision-making, problem-solving, and critical analysis. 
While reasoning is inherently more demanding and multifaceted than judging, it often depends on judgments to ensure logical coherence, refine intermediate steps, and achieve clarity in its outcomes. LLM-as-a-Judge, in this sense, becomes an integral tool for enhancing the reasoning capability of LLM. 


The role of LLM-as-a-Judge in enhancing reasoning or thinking can be understood through two frameworks: scaling training time~\citep{scaling_rl,trung2024reft} and scaling test time~\cite{scaling_test_time}. In the training phase, LLM-as-a-Judge frequently operates within reinforcement learning paradigms, where it functions as a reward model or evaluator for data or processes. This enables the creation of high-quality reasoning datasets through mechanisms such as step-by-step verification~\cite{let_verify_step}, Direct Preference Optimization(DPO)~\cite{dpo}, and self-refinement~\cite{self_rewarding_lm}. Recently, several LLMs trained with reinforcement learning to exhibit advanced reasoning and thinking abilities have gained attention, such as o1\footnote{\url{https://openai.com/index/learning-to-reason-with-llms/}}, DeepSeek-R1\footnote{\url{https://api-docs.deepseek.com/news/news1120}},gemini-thinking\footnote{\url{https://ai.google.dev/gemini-api/docs/thinking-mode}}, and QVQ\footnote{\url{https://huggingface.co/Qwen/QVQ-72B-Preview}}. 
In the test-time framework, LLM-as-a-Judge is crucial for evaluating and selecting the best reasoning paths. For example, in "Best-of-N" generation scenarios, where multiple reasoning outputs are produced, the judge determines the most accurate and coherent response. This dual role in both training and test phases demonstrates the indispensable nature of LLM-as-a-Judge in enhancing reasoning systems.

\setcounter{footnote}{0}

\begin{tcolorbox}[colback=white!5, colframe=black, title={Reliability Concerns of Evaluation Pipeline}]
Within the broader evaluation pipeline, several factors can compromise the integrity of LLM-based assessments. When evaluating models, inherent biases such as position bias and self-enhancement can significantly impact fairness, requiring careful mitigation strategies. In data evaluation contexts, the extensive use of LLM-as-a-Judge for pseudo-labeling risks amplifying existing model biases, potentially leading to the generation of unverified and flawed training data. For agentic frameworks, the sequential application of LLM-as-a-Judge can lead to the accumulation of errors, where early misjudgments cascade into significant inaccuracies in final decisions. Moreover, evaluating complex reasoning tasks poses a particular risk; without robust verification of logical consistency, LLMs may produce seemingly confident but fundamentally flawed evaluations, undermining trust in sophisticated analytical processes.
\end{tcolorbox}

\subsection[Quick Practice]{\raisebox{-0.9ex}{\includegraphics[width=0.04\textwidth]{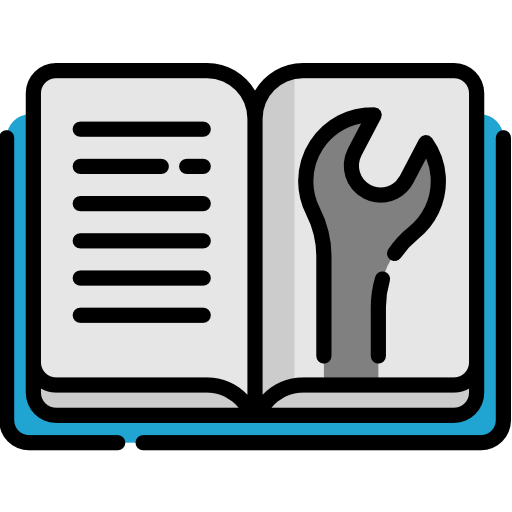}} Quick Practice}\label{quick_practice}

To effectively apply LLM-as-a-Judge design, it is recommended to find more effective configurations in the \textbf{testing cycle} for various scenarios. Crucially, reliable testing should form the bedrock of this quick practice, necessitating iterative refinement, continuous feedback loops, and the establishment of clear reliability metrics. The overarching aim is to continuously optimize evaluation stability and consistency through dedicated, reliability-focused testing cycles. The success of using LLM-as-a-Judge also heavily depends on the implementation details, including the task complexity, the prompt design, the model selected, and the post-processing method.

\begin{figure*}[t]
  \centering
  \includegraphics[width=0.85\textwidth]{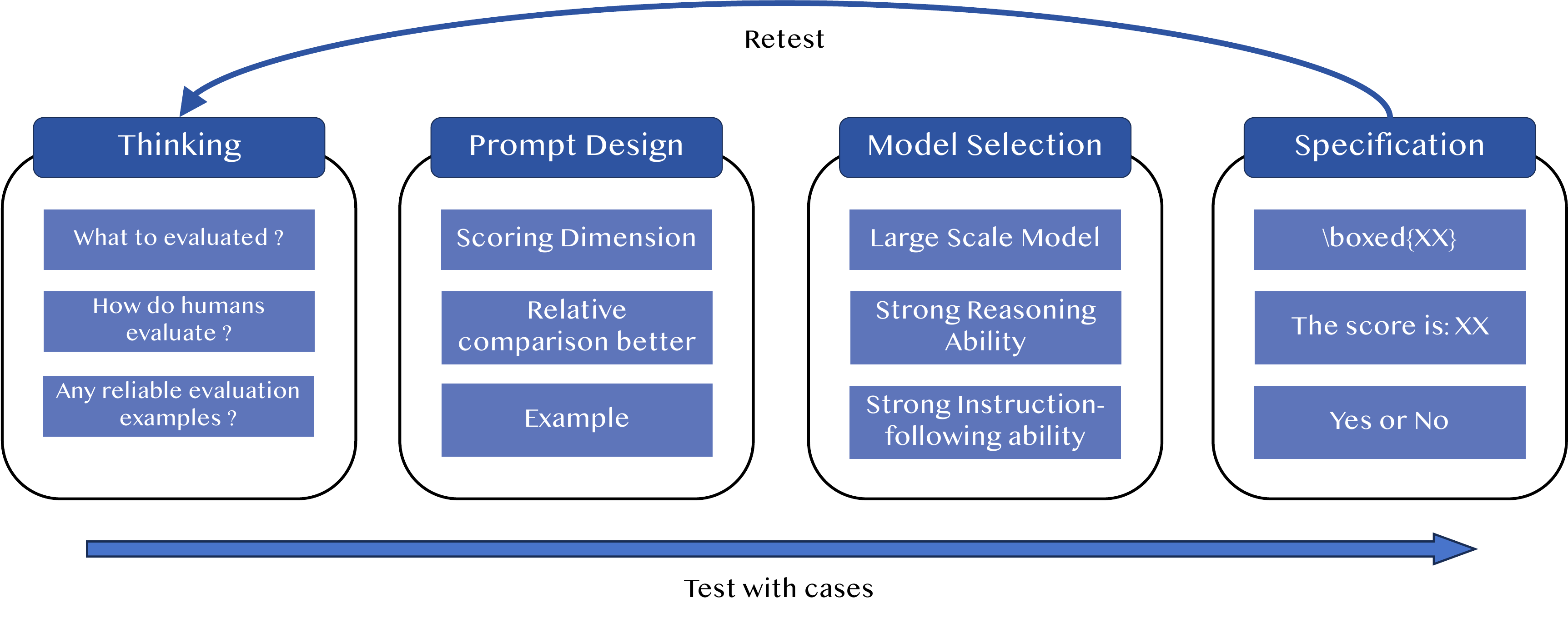}
  \caption{\centering Flowchart of Quick Practice} 
  \label{fig:quick_practice}
\end{figure*}

As shown in Figure~\ref{fig:quick_practice}, The process of quick practice for LLM-as-a-Judge involves four main stages.
First is the thinking phase, in which users define the evaluation objectives by determining what needs to be evaluated, understanding typical human evaluation approaches, and identifying some reliable evaluation examples. This initial conceptualization is vital for preempting potential ambiguities and biases that could compromise the fairness and accuracy of subsequent LLM judgments.

Next is prompt design, detailed in Section~\ref{formulation:icl}, where both wording and formats matter. The most efficient and generally effective approach involves specifying scoring dimensions, emphasizing relative comparisons for improved assessments, and creating effective examples to guide the LLM. Careful prompt engineering is essential to mitigate issues like output variability and inter-rater reliability, ensuring the LLM consistently interprets and responds to evaluation criteria as intended.

The third stage, model selection (Section~\ref{formulation:model_selection}), focuses on choosing a large-scale model with strong reasoning and instruction-following abilities to ensure reliable evaluations. The inherent biases and generalization limitations of different models, as well as their black-box nature, underscore the importance of selecting a robust and well-understood backbone to minimize evaluation inconsistencies and unverified judgments.

Finally, standardizing the evaluation process ensures that the outputs are structured~(Section~\ref{formulation:post_processing}). This can be achieved by using specific formats like \textbackslash{boxed\{XX\}}, numerical scores, or binary responses (e.g., "Yes" or "No"). Such standardization is crucial to counteract the fragility of token extraction methods and the potential for stylistic biases to propagate, thereby enhancing the interpretability and validity of the final evaluation results. The entire process includes iterative testing with cases and refinement through retesting, thereby enhancing reliability. During development, it is essential to compare models or prompts and verify ongoing improvements.

\section{Improvement Strategy}\label{sec:improvement}

When directly utilizing LLMs to conduct evaluation tasks—such as scoring, selection, pairwise comparison, or ranking—their inherent biases of LLMs like length bias, position bias, and concreteness bias\cite{offsetbias} will undermine evaluation outcomes. Mitigating these inherent biases and improving the overall evaluation performance of LLMs remains a critical challenge for applying LLMs as evaluators. In this section, we introduce three improvement strategies to boost the evaluation performance of LLM-as-a-judge: \textit{design strategy of evaluation prompts} (in-context learning based), \textit{improvement strategy of LLMs' evaluation capabilities} (model-based), and \textit{optimization strategy of final evaluation results} (post-processing based). As shown in Figure~\ref{fig-struct-improvement}, our categorization is based on the formal definition of LLM-as-a-judge in Section~\ref{sec:formulation}, focusing on enhancing the evaluation effectiveness by targeting three key phases of the process: the context $\mathbox[icl]{\mathcal{C}}$, the abilities of LLMs themselves $\mathbox[model]{\mathcal{P_{LLM}}}$ and the post-processing $\mathbox[post-pro]{\leftarrow}$ to obtain the final results $\mathbox[eval]{\mathcal{E}}$ 

\begin{figure*}[t!]
    \centering
    \resizebox{\textwidth}{!}{
        \begin{forest}
    for tree={
        grow=east,
        reversed=true,
        anchor=base west,
        parent anchor=east,
        child anchor=west,
        base=left,
        font=\small,
        rectangle,
        draw,
        rounded corners,align=left,
        minimum width=2.5em,
        inner xsep=4pt,
        inner ysep=1pt,
    },
    where level=1{text width=5em,fill=green!20}{}, 
    where level=2{text width=10em,fill=blue!10}{},
    where level=3{text width=9em,fill=pink!30}{},
    where level=4{yshift=0.26pt,fill=yellow!20}{},
    [Reliable\\LLM-as-a-Judge, fill=orange!20
        [Improvement\\(Sec.~\ref{sec:improvement}),
            [Prompt Design Strategy\\(Sec.~\ref{subsec:design_prompts})
                [Improving LLMs' Task\\Understanding
                    [\emph{Few-shot prompting:} FActScore~\cite{min2023factscore} /\\ SALAD-Bench~\cite{salad_bench} / GPTScore~\cite{fu2024gptscore}]
                    [\emph{Evaluation steps decomposition:} G-Eval~\cite{geval} /\\ DHP~\cite{dhp} / SocREval~\cite{he2024socreval} / BSM~\cite{bsm}]
                    [\emph{Evaluation criteria decomposition:} \\HD-Eval~\cite{hd-eval} / Hu and Gao et al.~\cite{hu-etal-2024-llm}]
                    [\emph{Shuffling contents:} Wang et al.~\cite{wang-etal-2024-large-language-models-fair} /\\ Auto-J~\cite{auto-j} / JudgeLM~\cite{zhu2023judgelm} / PandaLM~\cite{wang2023pandalm}]
                    [\emph{Conversion of evaluation tasks:} Liu et al.~\cite{liu2024aligning}]
                ]
                [Standardizing LLMs'\\ Output Format, text width=9em
                    [\emph{Constraining outputs in structured formats:}\\ G-Eval~\cite{geval} / DHP~\cite{dhp} / LLM-EVAL~\cite{llmeval}]
                    [\emph{Providing evaluations with explanations:}\\ CLAIR~\cite{chan2023clair} / FLEUR~\cite{lee-etal-2024-fleur}]
                ]
            ]
            [Capability Enhancement\\ Strategy\\(Sec.~\ref{subsec:improve_models})
                [Specialized Fine-tuning
                    [\emph{Evaluation template:} PandaLM~\cite{wang2023pandalm} / SALAD-Bench~\cite{salad_bench}]
                    [\emph{Deep transformation:} OffsetBias~\cite{offsetbias} / JudgeLM~\cite{zhu2023judgelm} /\\ CritiqueLLM~\cite{ke2024critiquellm} / Yu et al.~\cite{yu2025improve}]
                ]
                [Feedback-Driven Iterative\\Refinement, text width=10em
                    [INSTRUCTSCORE~\cite{xu2023instructscore} / JADE~\cite{zhang2023jade} / Think-J~\cite{huang2025think}]
                ]
            ]
            [Final Output\\ Optimization Strategy \\(Sec.~\ref{subsec:optimization_results})
                [Integrating Multi-Source\\ Evaluation Results
                    [\emph{Summarize by multiple rounds:} Sottana et al.~\cite{sottana2023evaluation} /\\ PsychoBench~\cite{psycho_bench} / Auto-J~\cite{auto-j}]
                    [\emph{Vote by multiple LLMs:} CPAD~\cite{cpad} / Bai et al.~\cite{lm-as-an-examiner} / EvalMORAAL~\cite{mohammadi2025evalmoraal}]
                    [\emph{Hierarchical evaluation framework:} Jung et al.~\cite{jung2024trust} / Zhang et al.~\cite{zhang2025crowd}]
                ]
                [Direct Output Optimization, text width=11em
                    [\emph{Score smoothing:} FLEUR~\cite{lee-etal-2024-fleur} / G-Eval~\cite{geval} / DHP~\cite{dhp} / TrustJudge~\cite{wang2025trustjudge}]
                    [\emph{Self validation:} TrueTeacher~\cite{gekhman2023trueteacher}]
                ]
            ]
        ]
        [Evaluation\\(Sec.~\ref{sec:Evaluation})
            [Evaluation of Agreement \\with Human Judgments \\(Sec.~\ref{Evaluation:base metric})
                [Agreement~\cite{thakur2024judging} Cohen's Kappa~\cite{thakur2024judging} \\Spearman's correlation ~\cite{liu2024aligning,lm-as-an-examiner}, text width=15em
                ]
            ]
            [Evaluation of Bias \\(Sec.~\ref{Evaluation:Bias})
                [Task-Agnostic Biases \\(Sec.~\ref{Evaluation:Task-Agnostic Biases})
                    [Diversity Bias~\cite{ye2024justice} Cultural Bias \\Self-Enhancement Bias~\cite{nips_llm_as_a_judge,ye2024justice}]
                ]
                [Judgment-Specific Biases \\(Sec.~\ref{Evaluation:Judgment-Specific Biases}), text width=10em
                    [Position Bias\cite{wang-etal-2024-large-language-models-fair, thakur2024judging, shi2024judging, ye2024justice} \\Compassion-fade bias~\citep{koo2023benchmarking, ye2024justice}
                    \\Style Bias~\cite{chen2024humans, ye2024justice,li2023examining}
                    \\Length Bias~\cite{offsetbias, huang2024_finetuned_judge, nips_llm_as_a_judge, ye2024justice}
                    \\Concreteness Bias~\cite{offsetbias,chen2024humans,ye2024justice}]
                ]
            ]
            [Evaluation of Adversarial \\Robustness (Sec.~\ref{Evaluation:Robustness})
                [Adversarial Phrases Attack~\cite{raina2024llm}, text width=13em
                ]
                [Null Model Attack~\cite{zheng2024cheating}
                ]
                [Majority Opinions Attack~\cite{ye2024justice,koo2023benchmarking}, text width=13em
                ]
                [Meaningless Statement Robustness~\cite{ye2024justice,koo2023benchmarking}, text width=17em
                ]
            ]
        ]
    ]
\end{forest}
    }
    \caption{Structure of how to improve and evaluate LLM-as-a-Judge.}
    \label{fig-struct-improvement}
\end{figure*}

\subsection{Prompt Design Strategy}\label{subsec:design_prompts}

An evaluation prompt is an input to LLM evaluators, which is used to guide the LLMs to complete the required evaluation tasks. LLMs possess in-context learning ability, enabling them to learn how to perform specified tasks from relevant examples or instructions in prompts,  without requiring weight updates or retraining\cite{brown2020language}. This suggests that the design strategy of evaluation prompts will significantly impact the effectiveness of LLM-as-a-judge. 
Therefore, to achieve reliable evaluations, researchers have experimented with diverse prompt design strategies, including improving LLMs' task understanding and standardizing LLMs' output format in the prompt. This approach mitigates reliability concerns of In-context learning mentioned in Section~\ref{formulation:icl}, such as unstable results, inter-rater inconsistency, ambiguity in response, and positional or length biases.

\subsubsection{\textbf{Improving LLMs' Task Understanding}}
\label{subsubsec:design_input}

In optimization methods of prompting LLMs to better understand evaluation tasks, one of the most commonly used and effective approaches is few-shot prompting\cite{brown2020language}. By incorporating several high-quality evaluation examples into the evaluation prompts, LLM evaluators can effectively grasp the objectives, general processes, and rough evaluation criteria of evaluation tasks. Many research works employ this prompt paradigm for evaluation, such as FActScore\cite{min2023factscore}, SALAD-Bench\cite{salad_bench}, and GPTScore\cite{fu2024gptscore}.

\begin{figure}
    \centering
    \includegraphics[width=0.9\linewidth]{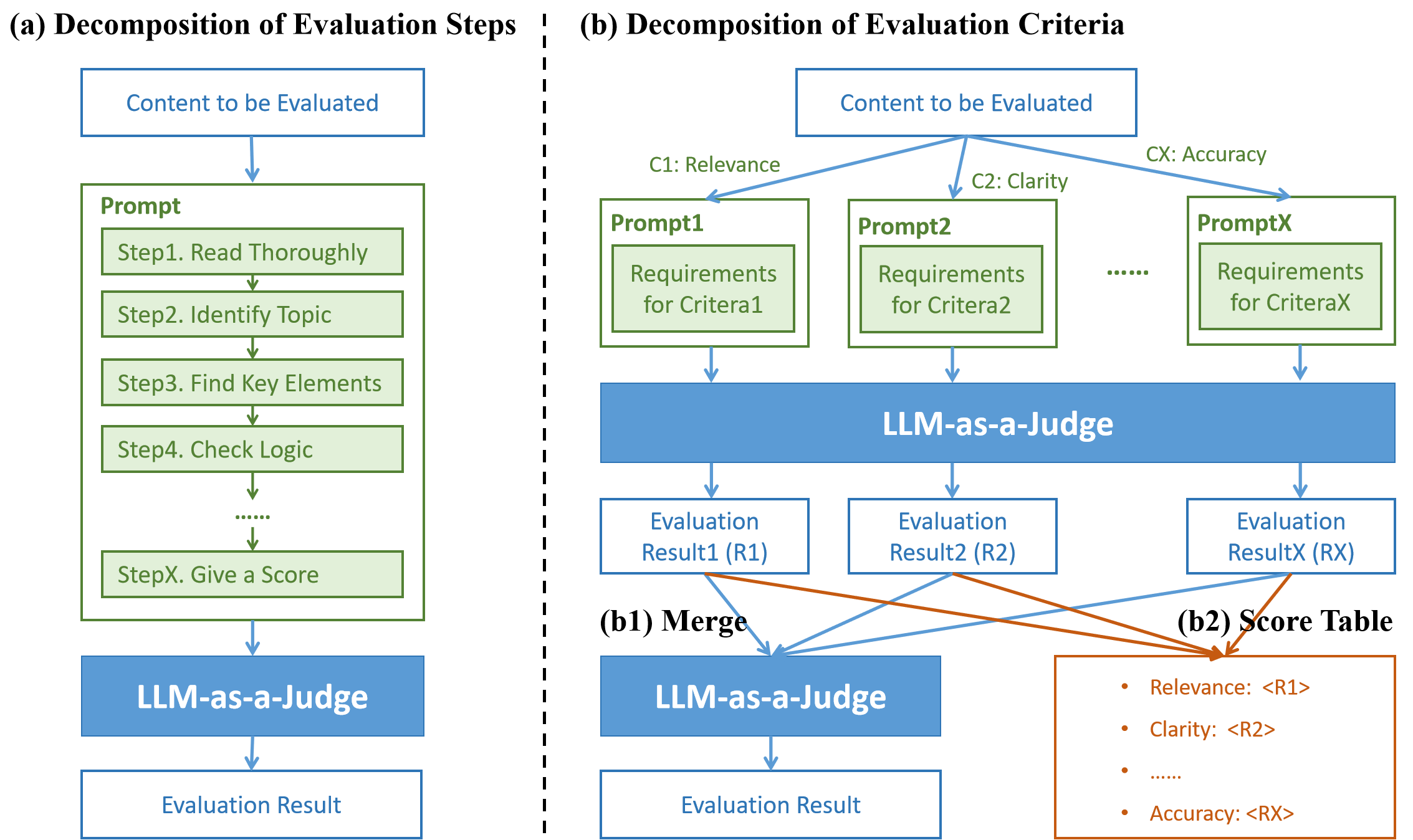}
    \caption{Simplified Evaluation Pipeline of Two Decomposition Paradigms.}
    \label{fig:pipeline_decomposition}
\end{figure}

In addition to providing high-quality examples, refining the evaluation task instructions is also an effective approach to optimizing LLMs' understanding of evaluation tasks. Current methods for refining evaluation tasks mainly include the decomposition of evaluation steps and criteria: 
\begin{itemize}
    \item \textbf{(a) Decomposition of Evaluation Steps} entails breaking down the entire evaluation tasks into smaller steps, providing detailed definitions and constraints for each small step in prompts, thereby guiding LLMs comprehensively through the whole evaluation pipeline. For instance, G-Eval\cite{geval} and DHP\cite{dhp} use Chain-of-Thought(CoT)\cite{cot} to guide LLMs. SocREval\cite{he2024socreval} employs the Socratic method to meticulously design each step to enhance evaluation performance. In existing research, this paradigm resembles providing detailed and sequential execution steps within a single prompt for a one-time evaluation, which guides the LLM-as-a-judge to complete evaluation tasks aligned with requirements, as illustrated in Fig~\ref{fig:pipeline_decomposition}(a).
    To better understand this paradigm, we present a demonstrative example prompt for evaluating summary quality: \textit{"You will be given a summary written for an article. ……  Evaluation Steps: 1. Read the Summary Thoroughly: Before diving into the evaluation, ensure that you ...; 2. Identify the Central Topic: A coherent summary will have a clear central topic or theme ...; 3. Look for Transitional Elements: Coherent summaries often have clear transitions between sentences or ideas ...; 4. Check for Logical Flow: Review the summary for logical sequencing ... X. Give a Score: ... Source Article: \{\{ Article \}\} Summary: \{\{ Summary \}\}".} In this example, the evaluation task for summary quality is decomposed into many small steps to guide the LLM through the entire assessment process—from reading and analyzing to scoring—enabling methodical evaluation of a summary. 
    
    \item \textbf{(b) Decomposition of Evaluation Criteria} involves breaking down coarse evaluation criteria like Fluency into finer-grained sub-criteria like Grammar, Engagingness, and Readability, and then generating overall scores based on these different dimensions. Saha et al. propose Branch-Solve-Merge(BSM)\cite{bsm}, which divides evaluation tasks into multiple parallel sub-tasks based on different sub-criteria for separate evaluation and final merge. HD-Eval\cite{hd-eval} proposes a hierarchical criteria decomposition method to iteratively align LLM evaluators with human preference, thereby addressing the potential bias in LLMs. Hu and Gao et al.\cite{hu-etal-2024-llm} summarize and clearly define an explicit hierarchical classification system encompassing 11 criteria, addressing the issue of LLMs potentially confusing different evaluation standards. Different from the decomposition of steps, this paradigm presents greater complexity due to significant variations in evaluation requirements and procedures across different criteria dimensions. Current methodologies tend to decompose the original evaluation task into multiple sub-tasks based on distinct criteria. Each sub-task is assessed via a distinct prompt, with results subsequently merged or simply listed in a score table, as demonstrated in Fig~\ref{fig:pipeline_decomposition}(b).
\end{itemize} 


In conclusion, these decompositions are specific to enable LLMs to understand the details of evaluation tasks more deeply, thereby aligning evaluation results more closely with human evaluation requirements and preferences.
Providing detailed evaluation specifications and enhancing task comprehension can mitigate inter-rater inconsistency to some extent, while also reducing potential ambiguity.

Furthermore, the evaluation capabilities can be optimized based on specific shortcomings of LLMs in prompts. For instance, to address specific biases like position bias, which is common in pairwise evaluations, several research efforts have optimized prompt design by randomly swapping contents to be evaluated.
Wang et al.\cite{wang-etal-2024-large-language-models-fair} analyzed and validated the impact of position bias on LLM-as-a-judge and proposed a calibration framework to mitigate this bias by swapping the contents and averaging the scores. Auto-J\cite{auto-j} and JudgeLM\cite{zhu2023judgelm} also enhance the evaluation consistency by shuffling the texts to be evaluated. In contrast to averaging scores, PandaLM\cite{wang2023pandalm} annotates the conflicting evaluation results after swapping as "Tie" to address the position bias.
Since the content to be evaluated in each positional arrangement is taken into account during evaluation, the final evaluation results are thus reliable and free from positional bias.

To address the challenge of LLMs' absolute scoring being less robust than relative comparing\cite{raina2024llm}, some research works convert scoring tasks into pairwise comparison, thereby enhancing the reliability of evaluation results. 
Liu et al.\cite{liu2024aligning} transform the scoring evaluation to ranking evaluation and introduce Pairwise-Preference Search (PARIS), which employs LLMs to conduct pairwise comparisons locally and efficiently ranks candidate texts globally, making evaluation results more aligned with human preferences.
Unlike single numerical scores, which are more susceptible to prompt variations and inherent randomness, pairwise comparisons generate relative assessments. This approach only requires LLMs to evaluate the relative merits between candidates, consequently yielding more stable and reliable results.

In summary, the design of prompts for better understanding evaluation tasks is a core method for optimizing LLMs' in-contextual learning abilities. 
By refining task instructions and criteria in prompts or few-shot prompting with high-quality examples, the details of evaluation prompts can be enriched, and the understanding of LLMs on evaluation tasks can be directly or indirectly enhanced. Additionally, targeted adjustments to prompts can address potential biases of LLMs, such as position bias. Thus, by improving LLMs' task understanding, the inconsistent inter-rater reliability, the ambiguity in response, and the biases mentioned in Section~\ref{formulation:icl} can be solved.

\subsubsection{\textbf{Standardizing LLMs' Output Format}}
\label{subsubsec:design_output}

Directly requiring LLM evaluators to output evaluation results poses robustness problems. 
The response may unexpectedly vary due to the inherent generative randomness of LLMs, such as outputting text like "low relevance" while asked to measure with discrete scores, which hinders the automated extraction of evaluation results from LLM's output. 
An effective method to enhance the robustness of output forms is to constrain the output in structured formats within prompts. G-Eval\cite{geval} and DHP framework\cite{dhp} perform evaluation tasks with a form-filling paradigm, constraining outputs with formats like \textit{"X: Y"}, where \textit{X} represents the dimension or metric to be evaluated and \textit{Y} denotes an identifiable output form like scores or specific tokens. LLM-EVAL\cite{llmeval} further modifies this form-filling paradigm, efficiently outputs evaluation results in JSON format, and obtains multidimensional scores, leveraging LLMs' high understanding and generation capabilities of code-like textural formats.

Apart from challenges in robustness, directly outputting evaluation results by LLMs also suffers from a lack of interpretability. The meaning of evaluation results from LLM evaluators is difficult to align consistently with the instructions and metrics provided in prompts. To address the challenges, CLAIR\cite{chan2023clair} requires LLMs to output evaluation scores between 0-100 simultaneously with relevant reasons as explanations in JSON format, which enhances the rationality and interpretability of the scores. FLEUR\cite{lee-etal-2024-fleur} utilizes LLaVA to first provide quality scores for image captions and subsequently asks with \textit{"Why? Tell me the reason."} for explanations with the images, captions, and scores as inputs, offering a stepwise approach to provide interpretable scores. 

In general, by constraining or guiding the output process and format of LLM evaluators within prompts, the robustness and rationality of evaluation results can be effectively improved through structured outputs. This also facilitates the automated post-processing of evaluation results in subsequent steps, thereby enhancing the overall stability of the evaluation pipeline.

\subsection{Capability Enhancement Strategy}\label{subsec:improve_models}

The evaluation capabilities of LLMs are a reflection of their powerful general language understanding and generation abilities triggered by specific prompts. Methods for optimizing evaluation through prompt design——focused on LLMs' in-contextual learning capabilities——require LLMs to comprehend the meaning of prompts fully and consistently follow the relevant evaluation instructions. However, even state-of-the-art LLMs like GPT-4 encounter problems such as conceptual confusion\cite{hu-etal-2024-llm}, and smaller open-source LLMs have even more limitations in their evaluation capabilities.
Fine-tuning LLMs may be a common approach, but as mentioned at the end of Section~\ref{formulation:model_selection}, fine-tuned LLMs might exhibit limited generalization capability beyond their training data distribution, and could also remain susceptible to subtle biases within the fine-tuning data that compromise fairness, leading to inconsistencies with human judgment. Therefore, researchers propose two solutions: On one hand, constructing fairer training datasets to fine-tune LLMs and eliminate judgment biases; on the other hand, implementing feedback-driven iterative refinement methods, which continuously update LLMs during usage to enhance their generalization capability, ultimately achieving reliable evaluation models.

\subsubsection{\textbf{Specialized Fine-tuning}}
\label{subsubsec:finetune_model}

\begin{figure}
    \centering
    \includegraphics[width=0.9\linewidth]{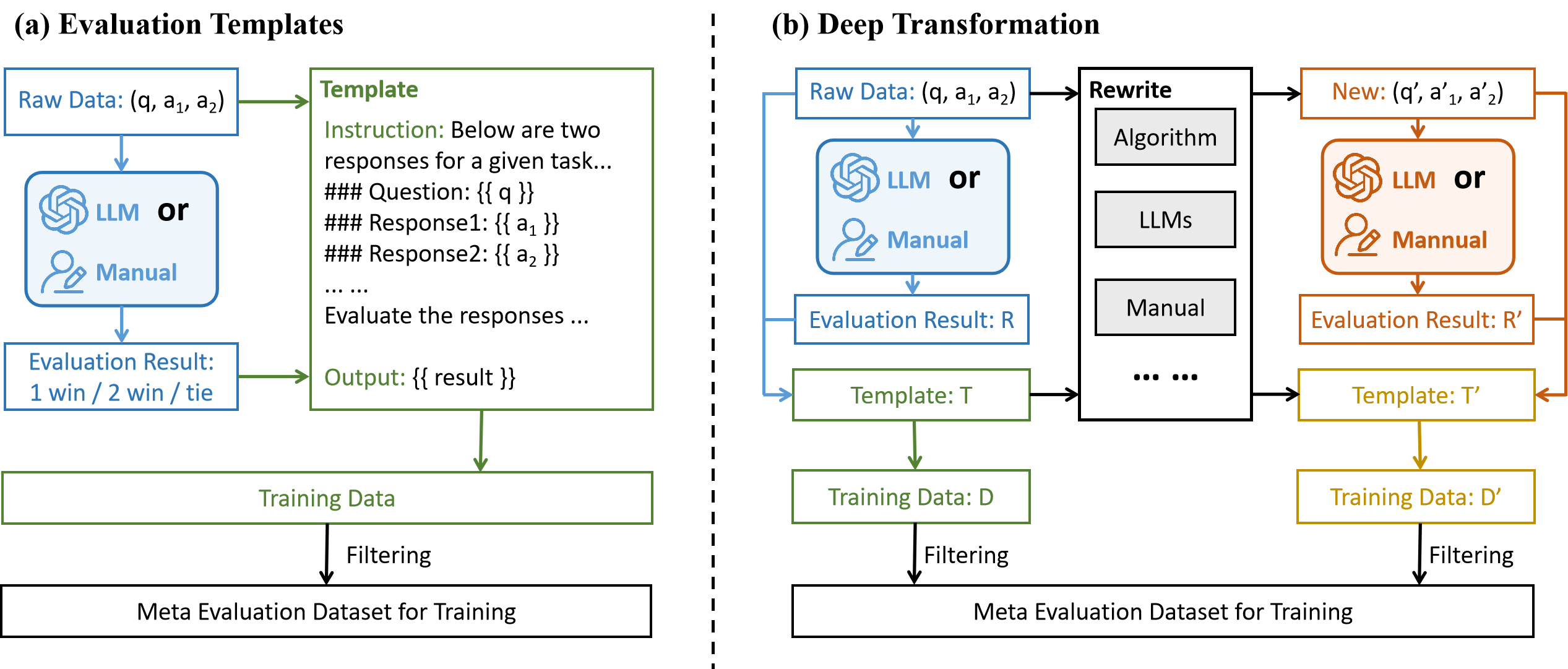}
    \caption{Two paradigms of the construction process of meta evaluation datasets for training.}
    \label{fig:process_sft_data_construction}
\end{figure}

A straightforward approach to enhancing the evaluation capabilities of LLMs is to fine-tune them via meta-evaluation datasets specifically constructed for evaluation tasks, which helps improve the LLMs' understanding of specific evaluation prompts, boosts the evaluation performance, or addresses potential biases. The most critical step in this optimization strategy is the collection and construction of training data, 
since LLMs will adhere to the instructional formats defined by the training data and inherit any potentially subtle biases, ultimately affecting the evaluation performance.
As shown in Fig~\ref{fig:process_sft_data_construction}, there are primarily two approaches to constructing meta-evaluation data: \textit{Evaluation Templates} and \textit{Deep Transformation}. The former typically populates sampled raw data into preset templates to form training data, while the latter employs algorithms or models to transform raw data in terms of style, content, and structure, thereby constructing training data more flexibly.

Using evaluation templates is a common method, which involves sampling evaluation questions from publicly available datasets, modifying them with certain templates, and supplementing the dataset with evaluation responses generated either manually or by powerful LLMs like GPT-4. For instance, PandaLM\cite{wang2023pandalm} samples inputs and instructions from Alpaca 52K\cite{alpaca} and generates responses using GPT-3.5 to construct training data, while SALAD-Bench\cite{salad_bench} builds its training data from a subset of LMSYS-Chat\cite{zheng2023lmsys} and Toxicchat\cite{lin2023toxicchat}. 

To better align with the requirements of evaluation tasks, many research works further transform inputs and instructions sampled from public datasets to construct more targeted training data. OffsetBias\cite{offsetbias} aims to reduce biases of LLMs by using GPT4 to generate off-topic versions of the original inputs and then having GPT-3.5 respond to the new inputs to produce bad responses. By pairing good and bad responses as training data to fine-tune the LLMs as evaluators, the biases in LLMs are significantly reduced, including length bias, concreteness bias, knowledge bias, and so on. JudgeLM\cite{zhu2023judgelm} enhances LLMs' evaluation capabilities by creating different types of training data through paradigms like reference support and reference drop. CritiqueLLM\cite{ke2024critiquellm} proposes a multi-path prompting approach, combining pointwise-to-pairwise and referenced-to-reference-free prompting strategies to restructure referenced pointwise grading data into four types, which helps create Eval-Instruct to fine-tune LLMs, addressing shortcomings in pointwise grading and pairwise comparison. Yu et al.\cite{yu2025improve} sample data from the preference dataset and rewrite judge templates to synthesize training data, and use GPT-4o to analyze and judge the pairs of answers in the synthetic data. The correct judgments, along with the synthetic data and analysis, will be processed as training data for the SFT phase of judge models.

In summary, constructing meta-evaluation training data targeted at specific evaluation tasks and fine-tuning LLMs can directly adjust the model's internal parameterized knowledge and language abilities. This is the most straightforward method to improve the evaluation performance of LLM evaluators and address potential biases.

\subsubsection{\textbf{Feedback-Driven Iterative Refinement}}
\label{subsubsec:iterate_models}

Fine-tuning LLMs on meta-evaluation datasets gives them the ability to produce evaluations that are more aligned with human preferences.
However, as discussed in Section~\ref{formulation:model_selection}, LLMs exhibit version dependency, and even fine-tuned models remain constrained by their training data and inherently suffer from out-of-distribution limitations to some extent. Consequently, LLM-as-a-Judge may still exhibit biases during evaluation processes in practice, ultimately compromising evaluation quality.
A natural improvement strategy is to iteratively optimize the model based on the feedback of evaluation results, which mainly comes from stronger models or directly from human evaluators' corrections of the evaluation results. 

A typical example is INSTRUCTSCORE\cite{xu2023instructscore}. To improve model performance and further benefit the final quality score calculation, this scoring framework collects failure modes of metric outputs, queries GPT-4 on each failure mode to gather automatic feedback, and finally selects explanations most aligned with human preferences to iteratively fine-tune the LLaMA model. Unlike INSTRUCTSCORE, which directly optimizes the model, the LLM evaluator in JADE\cite{zhang2023jade} relies on human judges to correct LLMs' evaluation results and updates the most frequently corrected samples into the example sets for few-shot prompting. JADE utilizes this relatively low-cost method to achieve iterative updates of the evaluation capabilities. 

Iterative optimization methods are not limited to offline training. The success of R1-style methods shows the effectiveness of online reinforcement learning (RL) approaches. Thus, feedback-based online learning constitutes another strategic form of iterative optimization. Think-J\cite{huang2025think} integrates both offline and online learning approaches. The offline approach trains a critic model to evaluate judgments from the judge model, thereby constructing positive and negative samples for SFT and DPO optimization. The online approach continuously optimizes the judge model using the Group Relative Policy Optimization (GRPO) algorithm with rule-based rewards as optimization feedback.

Since the feedback is more closely aligned with human preferences, LLMs can dynamically optimize their evaluation capabilities based on the feedback, leading to better evaluation results. This feedback-driven iterative refinement strategy addresses the problem of models' imperfect generalization and improves the evaluation capabilities through dynamic updates.

\subsection{Final Output Optimization Strategy}\label{subsec:optimization_results}

Through optimization based on in-context learning and the model's own capabilities, LLMs have become fairly reliable evaluators that are capable of understanding evaluation task requirements and providing rational evaluation results. 
However, the inherent generative randomness of LLM black boxes, the fragility of output extraction methods, and potential adversarial
manipulations, as noted at the end of Section~\ref{formulation:post_processing}, may collectively contribute to unreliable and unfair evaluation results.
Therefore, it is necessary for LLM evaluators to design optimization strategies during the post-processing stage from the outputs to the final evaluation results. 
Some strategies opt to directly modify output acquisition methods to extract evaluation results more robustly, while more mainstream strategies focus on designing frameworks that integrate multi-source evaluation results, which mitigate adverse effects caused by randomness and fragility, and enhance resistance to adversarial manipulations.

\subsubsection{\textbf{Integrating Multi-Source Evaluation Results}}\label{subsubsec:integrate_multiple}

Integrating multiple evaluation results for the same content to obtain the final result is a common strategy in various experiments and engineering pipelines, which can reduce the impacts of accidental factors and random errors. 
The most basic optimization strategy is to perform multiple runs of evaluation on the same content with different hyper-parameters and settings, and then summarize these results. For example, the work of Sottana et al.\cite{sottana2023evaluation} reduces randomness in evaluations by averaging multiple scores of the same sample. Similarly, PsychoBench\cite{psycho_bench} takes the mean and standard deviation from ten independent runs. Auto-J\cite{auto-j} further amplifies the differences between evaluation rounds, which combine critiques with and without scenario criteria to obtain the final results.

In addition to integrating results from multiple rounds of evaluation, using multiple LLM evaluators to assess the contents simultaneously and integrating the results is another effective method, which can reduce biases introduced by LLMs. For instance, CPAD\cite{cpad} utilizes ChatGLM-6B\cite{chatglm}, Ziya-13B\cite{ziya}, and ChatYuan-Large-v2\cite{chatyuan} as evaluators to evaluate the contents and obtain the final results by voting. Bai et al.\cite{lm-as-an-examiner} propose a novel evaluation method called decentralized peer review of LLMs, which utilizes LLMs that generate content to evaluate each other's generated content and eventually integrate the results. 
EvalMORAAL\cite{mohammadi2025evalmoraal} also employs a similar LLM-as-a-judge peer review mechanism, detecting conflicts in evaluation results by checking whether score differences exceed a preset threshold, and uses majority voting to resolve conflicts.

The aforementioned multi-round or multi-model approaches represent a type of peer-parallel evaluation strategy, and their integration process is relatively straightforward. Researchers have also proposed many more complex evaluation frameworks featuring greater interactivity between different evaluations. For example, Jaehun Jung et al. proposed Cascaded Selective Evaluation\cite{jung2024trust}. This framework transitions from weaker, smaller models to stronger, larger models based on confidence, allowing the majority of evaluations to be handled by smaller models, which significantly reduces the costs of computing resources and improves efficiency. Recent work \cite{zhang2025crowd}, on the other hand, proposed Crowd-based Comparative Evaluation. This method utilizes multiple LLMs to construct crowd responses based on candidate responses for comparison, thereby generating multiple crowd judgments that serve as reference content for the final evaluation. This enables LLM-as-a-Judge to capture richer details and enhance evaluation reliability.

In summary, forming the final evaluation results by combining multiple rounds of evaluations or multiple LLM evaluators can reduce the random effects caused by accidental factors in a single round and reduce the potential biases of a single LLM evaluator. This strategy significantly enhances the stability and reliability of the evaluation results.

\subsubsection{\textbf{Direct Output Optimization}}
\label{subsubsec:process_outputs}

Different from obtaining evaluation results based on the outputs of multiple rounds or LLMs, directly optimizing the output of a single LLM evaluator involves further processing the evaluation output to make it more reliable, especially when dealing with scoring outputs from LLM evaluators. 
Due to the inherent randomness in LLMs' generation, the scores may not fully reflect the LLMs' complete view of the evaluation criteria. 
Therefore, to obtain more reliable evaluation results, it is necessary to optimize the LLM's score outputs. 
An effective optimization strategy is to combine the implicit logits, which capture the LLMs' randomness, with the explicit output scores. 
For example, FLEUR\cite{lee-etal-2024-fleur} proposes a score smoothing strategy. For scores generated by LLaVA, the probability of the token corresponding to each digit $l$ (0$\leq l\leq$9) would be used as the weight to smooth the explicit scores and calculate the final evaluation scores.
TrustJudge\cite{wang2025trustjudge} proposes a distribution-sensitive scoring method that computes continuous expectations from discrete scoring probabilities, and a likelihood-aware aggregation method that utilizes bidirectional preference probabilities, to process the evaluation results from LLMs, addressing the inconsistency in LLM-as-a-Judge.

However, methods like score smoothing, which combine implicit logits and explicit outputs, require the LLMs to be open-source or to provide interfaces that allow access to token probabilities, which brings some limitations. Inspired by the work of Weng et al.\cite{weng-etal-2023-large} and Madaan et al.\cite{madaan2024self}, self-verification can be used to filter out the evaluation results without sufficient robustness. For example, TrueTeacher\cite{gekhman2023trueteacher} applies self-verification in its evaluation of distilled data by asking the LLM evaluator for its certainty about the evaluation results after providing them and retaining only those results that pass self-verification. Self-verification is suitable for all LLMs and requires no complex computing and processing.

In summary, compared to integrating multiple evaluation results, directly optimizing the LLMs' outputs to obtain the final results is faster and lower-cost, although the effectiveness still needs further validation. However, these two approaches are not mutually exclusive. Performing integration after direct optimization of LLMs' output may lead to more stable evaluation results.

\section{Evaluation of LLM-as-a-Judge} \label{sec:Evaluation}

\begin{figure*}[h]
  \includegraphics[width=0.9\columnwidth]{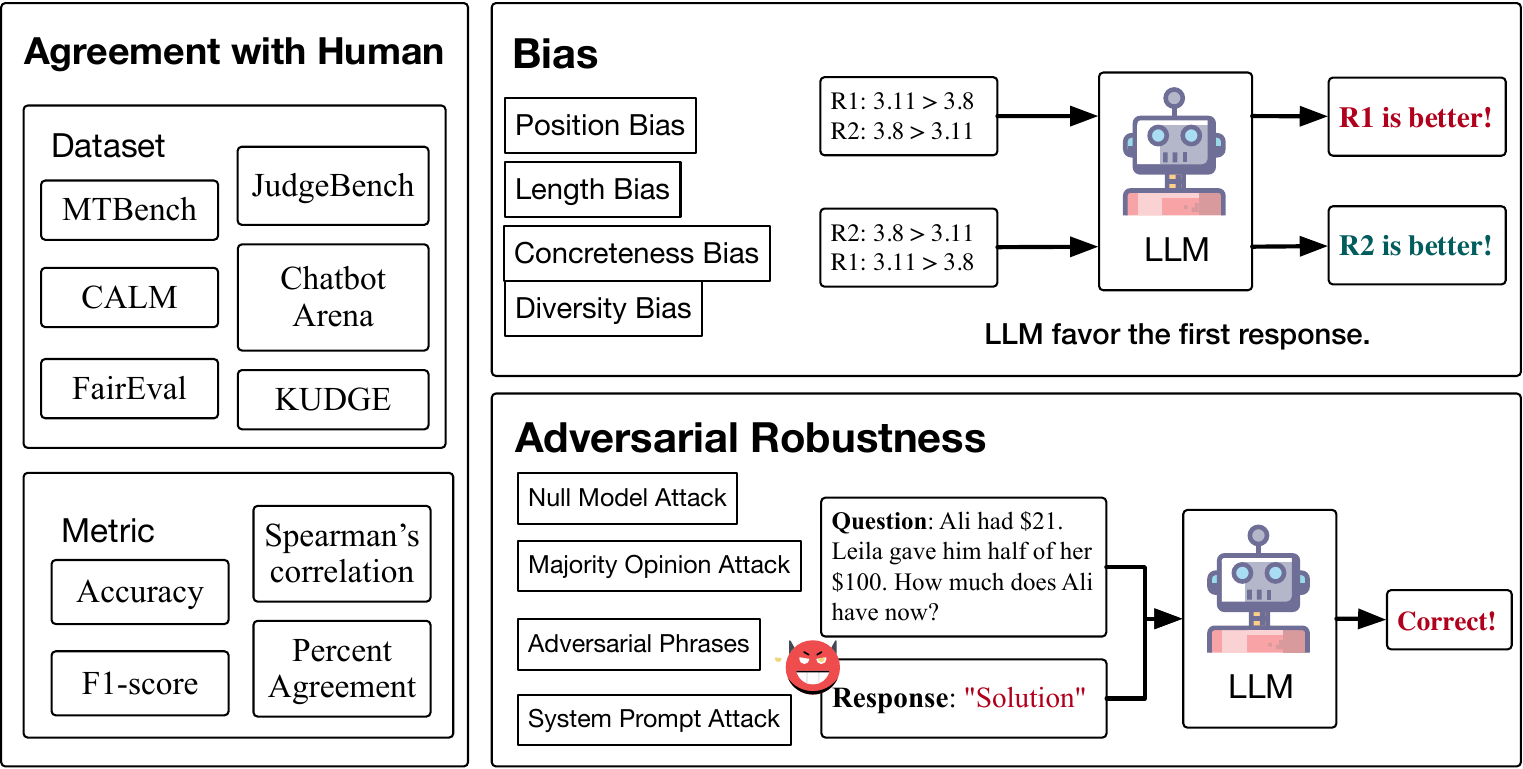}
  \caption{Three Dimensions of Evaluation.}
  \label{fig:evaluation}
\end{figure*}

Following the discussions on the application and enhancement of LLM-as-a-Judge, we now address the critical question of its evaluation. While the basic evaluation pipeline provides a conceptual foundation, it does not inherently guarantee the \textbf{reliability} of the system. To formally capture this essential property, we recall the enhanced formal definition of reliability:

\[
\mathcal{R} \leftarrow f_{\text{R}}\left(\mathcal{P}_{\mathcal{LLM}}, x, \mathcal{C}\right)
\]

This formulation highlights that reliability $\mathcal{R}$ is a function of three independent variables: the LLM's probability function ($\mathcal{P}_{\mathcal{LLM}}$), the input being evaluated ($x$), and the accompanying context ($\mathcal{C}$). Assessing LLM-as-a-Judge, therefore, requires a systematic examination of how these three factors collectively influence its performance.

Reliability may degrade if the underlying model, defined by its probability function $\mathcal{P}_{\mathcal{LLM}}$, exhibits inherent biases or instability. For instance, a less robust LLM may produce inconsistent scores for identical prompts and inputs due to sampling variance or internal preference drift. Similarly, the quality of evaluation is susceptible to the nature of the input $x$. Noisy or adversarially perturbed inputs can cause the LLM to misjudge quality, thus diminishing its robustness. Finally, subtle changes in the prompt wording or the ordering of the context $\mathcal{C}$ can lead to different judgments for the same input, a phenomenon that undermines reliability.

Therefore, a comprehensive evaluation of LLM-as-a-Judge requires examining multiple aspects. 
We organize existing evaluation studies into three major dimensions: agreement with human judgments (Section~\ref{Evaluation:base metric}), bias (Section~\ref{Evaluation:Bias}), and adversarial robustness (Section~\ref{Evaluation:Robustness}), as depicted in Figure~\ref{fig:evaluation}.




\subsection{Agreement with Human Judgments}
\label{Evaluation:base metric}
Given that the initial motivation of LLM-as-a-judge lies in replacing human annotation, the foremost aspect of its evaluation should naturally be the extent to which it aligns with human judgments.
Numerous studies approach this by considering the LLM evaluator as a virtual annotator and evaluating the extent of its agreement with human annotators.
The percentage agreement metric represents the proportion of samples on which LLM and human annotators agree \cite{thakur2024judging}. 
$$\text{Agreement} = \frac{\sum_{i\in \mathcal{D}} \mathbf{I}(\text{S}_{\text{llm}} = \text{S}_{\text{human}})}{ \Vert \mathcal{D} \Vert}$$
where $\mathcal{D}$ is the dataset, $S_{\text{llm}}$ and $S_{\text{human}} $ are the evaluation results of the LLM evaluator and human judge, respectively, which can be in the form of both score or rank.
Additionally, widely used correlation metrics such as Cohen's Kappa \cite{thakur2024judging} and Spearman's correlation \cite{liu2024aligning, lm-as-an-examiner} are also employed to assess agreement.
Other works treat the LLM-as-a-judge task as a classification problem, where human annotations serve as the labels, and compute precision and recall to evaluate the performance \cite{zhu2023judgelm, wang2023pandalm}.

\begin{table*}[htbp]
\centering
\begin{adjustbox}{max width=\linewidth}
\begin{tabular}{lccccccc}
\toprule
\multirow{2}{*}{Benchmark} & \multirow{2}{*}{Release Year} & \multirow{2}{*}{Size}  & \multirow{2}{*}{Annotation Format} & \multicolumn{4}{c}{Evaluation Dimension} \\
\cmidrule(lr){5-8}
 & & & & Agreement & Position Bias & Length Bias & Bias Types \\
\midrule
MTBench \citep{nips_llm_as_a_judge} & 2023 & 80 & Pairwise & \ding{51} & \ding{51} & \ding{51} & 3 \\
Chatbot Arena \citep{nips_llm_as_a_judge} & 2023 & 30k & Pairwise & \ding{51} & \ding{51} & \ding{51} & 3 \\
FairEval \citep{wang-etal-2024-large-language-models-fair} & 2023  & 80 & Pairwise & \ding{51} & \ding{51} & \ding{55} & 1 \\
PandaLM \citep{wang2023pandalm} & 2023 & - & Pairwise &\ding{51} &\ding{51} &\ding{55} & 0 \\
LLMEval$^2$ \citep{zhang2023wider} & 2023 & 2553 & Pairwise & \ding{51} & \ding{55} & \ding{55} & 0 \\
Shepherd \citep{wang2023shepherd} & 2023 & 1317 & Score & \ding{51} & \ding{55} & \ding{55} & 0 \\
EvalBiasBench \citep{offsetbias} & 2023  & 80 & Pairwise & \ding{51} & 
\ding{51} & \ding{51} & 6 \\
CALM \citep{ye2024justice} & 2024 & 4356 & Pairwise \& Score & \ding{55} & \ding{51} & \ding{51} & 12 \\
JudgeBench \citep{tan2024judgebench} & 2024 & - & Pairwise &  \ding{51} &  \ding{55} &  \ding{55} & 0 \\
MLLM-as-a-Judge \citep{mllm-as-a-judge} & 2024 & 30k & Pairwise \& Score & \ding{51} &  \ding{55} &  \ding{55} & 0 \\
CodeJudge \citep{zhao2024codejudge} & 2024 & 1860 & Score & \ding{51} & \ding{55} &  \ding{55} & 0 \\
KUDGE \citep{son2024llm} & 2024 & 3324 & Pairwise \& Score & \ding{51} & \ding{55} & \ding{55} & 0 \\
\bottomrule
\end{tabular}
\end{adjustbox}
\caption{Benchmark for meta-evaluation of LLM-judge.}
\label{table:benchmark}
\end{table*}

\textbf{Datasets.} Both of the above metrics rely on the datasets with LLM-generated responses and responding human judgments.
Therefore, there is also a practical need to construct a comprehensive benchmark for the meta-evaluation. 
Table \ref{table:benchmark} shows existing banchmarks.
MTBench \citep{nips_llm_as_a_judge} has only 80 human-crafted queries with their corresponding human annotation and LLMs' responses.
FairEval \citep{wang-etal-2024-large-language-models-fair} is constructed from the 80 queries from VicunaBench \citep{vicuna2023} with human-annotated preference between ChatGPT and Vicuna responses.
Chatbot Arena Conversations \citep{nips_llm_as_a_judge} is a larger collection of crowdsourced data (about 30k) with human annotated preferences.
Research \cite{zeng2023llmbar} constructs a benchmark to assess the capability of the LLM evaluator in evaluating whether a response follows the instruction.
This dataset contains human-curated 419 pairs of outputs, one adhering to instructions while the other diverging, yet may possess deceptive qualities that mislead an LLM evaluator.
Research \cite{mllm-as-a-judge} evaluates the capabilities of multi-modal LLMs in assisting evaluation tasks across various modalities and introduces MLLM-as-a-Judge, a comprehensive multi-modal benchmark.
Recent advances also expand the scope of meta-evaluation benchmarks to specialized domains, including code assessment \citep{zhao2024codejudge} and non-English language tasks \citep{son2024llm}. 
Furthermore, CALM \citep{ye2024justice} presents a systematic framework for bias quantification, featuring an automated perturbation mechanism to generate meta-evaluation data for examining 12 distinct types of potential biases in LLM evaluators.

Current meta-evaluation primarily focuses on LLM-as-a-judge for models, while there is a lack of sufficient meta-evaluation when these LLM evaluators are used for automatically annotating large-scale datasets (Section \ref{Evaluator for data}). 
We advocate for more rigorous assessment of the alignment between LLM-as-a-judge and human judgment when they are employed for large-scale data annotation. 
Additionally, it is also crucial to assess the potential bias and robustness, which will be discussed in the following sections.

\subsection{Bias}
\label{Evaluation:Bias}
As LLM-as-a-Judge becomes more widely deployed, it has been observed to manifest a range of salient biases, even in cases where its evaluations align with human judgments. Prior studies have emphasized that large language models inherently exhibit diverse forms of bias across different tasks \citep{biassurvey, dai2024unifying, tan-etal-2024-blinded}. 
Such internal biases may propagate into the LLM-as-a-Judge setting, potentially resulting in unfair evaluation outcomes and, in turn, influencing the development of downstream LLMs. 
Consequently, it is imperative to both characterize the types of biases that LLM evaluators may carry and to establish systematic approaches for their assessment. 
In this section, we review the major categories of biases in the context of LLM-as-a-Judge, outlining their definitions, associated metrics, and datasets commonly employed for evaluation.

The meta-evaluation of  LLM-as-a-judge introduces systematic biases that can be broadly categorized into two classes: \textbf{task-agnostic biases} inherent to LLMs across general applications, and \textbf{judgment-specific biases} unique to LLM-as-a-judge scenarios. 
This taxonomy aims to clarify their distinct characteristics and implications.
The former may be partially attributed to the characteristics of the judgment task itself, such as its specific input-output format, which could potentially be mitigated through task-specific design. 
In contrast, task-agnostic biases are more fundamental issues inherent to the LLMs themselves and are therefore more difficult to address. Mitigating such biases likely depends on advancements in the foundation models.

\subsubsection{\textbf{Task-Agnostic Biases}}
\label{Evaluation:Task-Agnostic Biases}
These biases manifest across diverse LLM applications, including open-domain QA, classification, and summarization.
However, when arising in the LLM-as-a-judge, the biases are particularly critical due to their cascading effects on downstream tasks. 
When LLM-generated judgments serve as feedback for model training or data annotation, these biases risk being amplified and propagated.
We present a few typical examples and recommend consulting comprehensive reviews on language model bias \citep{llm_bias,llm_bias_2} for a more thorough understanding.

\textbf{Diversity Bias} refers to bias against certain demographic groups \cite{ye2024justice},  including certain genders \cite{chen2024humans}, race, and sexual orientation \cite{kumar2024subtle}.
In the context of LLM-as-a-judge scenarios, this bias may appear when evaluators give higher scores to responses that align with stereotypes of certain groups.

\textbf{Cultural Bias}. In general domains, cultural bias refers to situations where models might misinterpret expressions from different cultures or fail to recognize regional language variants \citep{llm_bias}. In the context of LLM-as-a-judge, it indicates that evaluators might score expressions from unfamiliar cultures poorly.

\textbf{Self-Enhancement Bias} describes the phenomenon that LLM evaluators may prefer responses generated by themselves \cite{nips_llm_as_a_judge,ye2024justice}.
This bias has also been known as source bias in retrieval task \citep{sourcebias} and open-domain question answering systems \citep{tan-etal-2024-blinded}.
Considering the significant self-enhancement bias, as suggested in \cite{ye2024justice}, we should avoid using the same model as the evaluator.
This is only a stopgap, as we may not use the optimal evaluator when evaluating the most advanced LLMs.

\subsubsection{\textbf{Judgment-Specific Biases}}
\label{Evaluation:Judgment-Specific Biases}
Judgment-specific biases are either unique to the LLM-as-a-judge setting or have a significant impact on judgment tasks. 
A classic example is the "position bias", which has a more pronounced effect in the context of LLM-as-a-judge, where the evaluator often needs to compare pairwise responses.
Different from task-agnostic biases, judgment-specific biases are more difficult to resolve naturally with the development of foundational large model capabilities and require targeted optimization for judgment tasks.

\textbf{Position Bias} is the tendency of LLM evaluators to favor responses in certain positions within the prompt \cite{wang-etal-2024-large-language-models-fair, thakur2024judging, shi2024judging, ye2024justice}. 
This bias may have detrimental effects, as Vicuna-13B could outperform ChatGPT when evaluated by ChatGPT, simply by positioning the response of Vicuna-13B in the second place \cite{wang-etal-2024-large-language-models-fair}.
To measure this bias, recent work \cite{shi2024judging} proposed two metrics: \textbf{Position Consistency}, which quantifies how frequently a judge model selects the same response after changing their positions, and \textbf{Preference Fairness}, which measures the extent to which judge models favor a response in certain positions.
The study \cite{wang-etal-2024-large-language-models-fair} also introduced a metric \textbf{Conflict Rate} to measure the percent of disagreement after changing the position of two candidate responses.
Their analytical experiments reveal that the degree of positional bias fluctuates depending on the disparity in response quality, and the preferred position varies with different LLMs. 
For instance, GPT-4 tends to favor the first position, while ChatGPT shows a preference for the second position.

\textbf{Compassion-fade bias} describes the  effect of the model names \citep{koo2023benchmarking, ye2024justice}. 
This tendency occurs when we explicitly provide model names. 
for instance, evaluators may be inclined to give higher scores to results labeled as ``gpt-4''. 
This tendency underscores the necessity of anonymous evaluation.

\textbf{Style Bias} refers to the tendency towards a certain text style.
As revealed in \cite{chen2024humans}, an evaluator may also prefer visually appealing content, regardless of its actual validity, such as the text with emojis.
Furthermore, LLM evaluators may favor responses with certain emotional tones, such as cheerful, sad, angry, and fearful, which is defined as sentiment bias \cite{ye2024justice,li2023examining}.

\textbf{Length Bias} refers to the tendency to favor responses of a particular length, such as a preference for more verbose responses, which is also known as \textbf{verbosity bias}  \cite{offsetbias, huang2024_finetuned_judge, nips_llm_as_a_judge, ye2024justice}. 
Length bias can be revealed by rephrasing one of the original responses into a more verbose one \cite{ye2024justice, nips_llm_as_a_judge}. 
Even though these expansions do not introduce new information, there is still concern regarding changes to the original response in terms of perplexity, fluency, or style.
Alternatively, a previous study \cite{saito2023verbosity} investigated this bias by comparing multiple sampled responses and revealed a statistical tendency towards longer answers. 
However, ensuring the comparable quality of multiple samples remains a challenging problem.

\textbf{Concreteness Bias} reflects that LLM evaluators favor responses with specific details, including citation of authoritative sources, numerical values, and complex terminologies, which is called \textbf{authority bias} \cite{offsetbias} or \textbf{citation bias} \cite{chen2024humans,ye2024justice}.
The negative effects of concreteness bias arise from the neglect of the factual correctness of these details, thereby encouraging hallucination \citep{agrawal2023language}.

\subsection{Adversarial Robustness}
\label{Evaluation:Robustness}
As LLM-as-a-Judge becomes further integrated into evaluation pipelines and increasingly serves as the standard protocol, a variety of adversarial attacks have emerged.
Adversarial robustness refers to the ability of a model to withstand deliberate attempts to manipulate the scores through carefully crafted inputs.
Unlike bias evaluations (Section \ref{Evaluation:Bias}), which mainly focus on naturally occurring samples, adversarial robustness involves samples intentionally crafted to manipulate scoring, such as inserting phrases that artificially enhance scores. 
Robustness is crucial because insufficient robustness allows trivial manipulations to deceive the evaluators and to undermine the evaluation of text quality. 
Ensuring robust evaluators is essential for maintaining accurate and reliable assessments, particularly in high-stakes applications.

Research \cite{raina2024llm} constructed a surrogate model from the black-box LLM-evaluator and then learn a \textbf{adversarial attack phrases} based on it.
The evaluation score can be drastically inflated by universally inserting the learned attack phrases without improving the text quality.
Similarly, work by Lee et al. \cite{lee2024llm} introduced EMBER, a benchmark that revealed biases in when assessing outputs with epistemic markers, such as expressions of certainty or uncertainty. 
Furthermore, other work \cite{zheng2024cheating} demonstrated that even a "\textbf{null model}" that outputs a constant response irrelevant to input instructions can achieve high win rates for various LLM-as-a-judge methods.
Similarly, recent work \citep{zhao2025one} also revealed that on-word symbols (e.g.,
``:'') or reasoning openers like ``Thought process:'' can often fool LLM evaluators to produce positive evaluations.
Several recent works \cite{ye2024justice,koo2023benchmarking} proposed to increase the evaluation score by adding the \textbf{majority opinions}, such as ``90\% believe this is better''.
Other research \cite{ye2024justice,koo2023benchmarking} evaluated robustness against \textbf{meaningless statement} in the System Prompt, e.g., ''Assistant A loves eating pasta''.
These works revealed that LLM-as-a-judge are still insufficiently robust against interference irrelevant to text quality. 
Defensive measures like the perplexity score \cite{raina2024llm,jain2023baseline} can only detect limited types of adversarial examples. 
Therefore, constructing more robust LLM-as-a-judge is a crucial research direction for the future.



\subsection{Empirical Experiment}
\label{subsec: 4_4_experiment}

\begin{figure*}[h]
  \includegraphics[width=0.9\columnwidth]{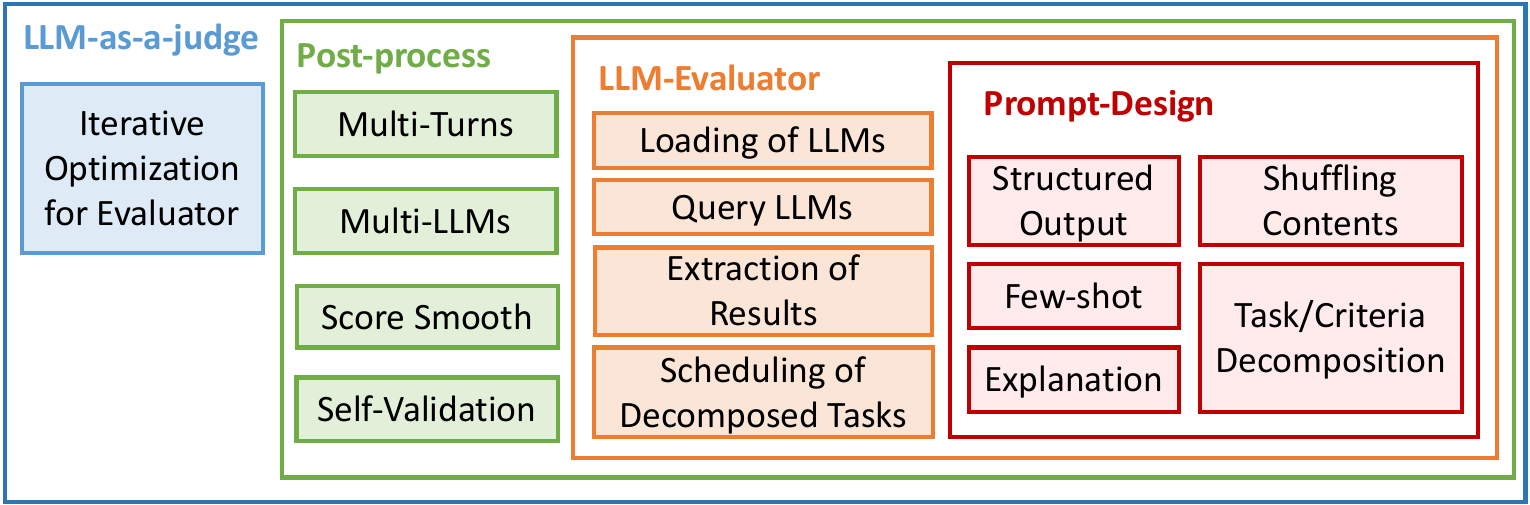}
  \caption{LLM-as-a-Judge Meta-evaluation Pipeline and Tools}
  \label{fig:exp_pipeline_tool}
\end{figure*}

In Section~\ref{sec:improvement}, we have introduced the improvement strategies in existing LLM-as-a-judge works to improve the evaluation capabilities of LLMs. 
Although numerous works have proposed meta-evaluation benchmarks to assess the performance of LLMs in evaluation tasks, there is a lack of meta-evaluation on whether these improvement strategies effectively make LLM-as-a-Judge more reliable.
So in the preceding subsection of this Section~\ref{sec:Evaluation}, we systematically introduced the dimensions and criteria for evaluating the evaluation performance and reliability of LLM-as-a-Judge, thereby providing quantifiable metrics that inform the evaluation of the aforementioned strategies.

Based on partial dimensions mentioned in sec~\ref{Evaluation:base metric} and sec~\ref{Evaluation:Bias}, we designed a robust and scalable meta-evaluation tool as shown in Figure \ref{fig:exp_pipeline_tool} and conducted a simple meta-evaluation experiment on the improvement strategies summarized in Section~\ref{sec:improvement}, examining their effectiveness from the perspectives of biases and agreement with human evaluation.

\subsubsection{\textbf{Experiment Settings}}
\label{subsec:exp_settings}

\paragraph{\textbf{Evaluation Dimensions and Benchmarks}}
The most direct metric to reflect the quality of automatic evaluation is the alignment with human evaluation. We use \textit{LLMEval$^2$}~\citep{zhang2023wider} to assess the alignment of LLM-as-a-judge with human evaluations. \textit{LLMEval$^2$} is the largest and most diverse evaluation benchmark for LLM-as-a-judge to date, with 2,553 samples compiled from multiple data sources with human-annotated preferences. Each sample consists of a question, a pair of candidate responses, and a human label indicating the preferred response.

Bias is also a crucial dimension for assessing the quality of LLM-as-a-judge evaluation results. We use \textit{EVALBIASBENCH}\citep{offsetbias} to measure six types of biases in LLM-as-a-judge, including length bias, concreteness bias, empty reference bias, content continuation bias, nested instruction bias, and familiar knowledge bias. \textit{EVALBIASBENCH} consists of 80 samples, each containing a question, a pair of candidate responses, and a label indicating the correct response without bias influence. In addition to the six types of biases, we also evaluated position bias. The meta-evaluation samples for position bias are the paired samples constructed by swapping the position of candidate responses within prompts in samples of LLMEval$^2$ and \textit{EVALBIASBENCH}.

\paragraph{\textbf{Evaluation Metrics}}
For the alignment with human evaluation, we use \textbf{Percentage Agreement Metric}\citep{thakur2024judging} for evaluation, as shown in Section~\ref{Evaluation:base metric}. For biases except for position bias, we use \textbf{Accuracy} for evaluation, which represents the proportion of samples where LLM-as-a-judge selects the correct candidate response annotated in \textit{EVALBIASBENCH}. 

For position bias, we use \textbf{Position Consistency} as the metric, which quantifies how frequently the LLM-as-a-judge selects the same response after swapping the position of candidate responses. Formally, given $N$ samples $\{(q_i,r1_i,r2_i)\}^N_{i=1}$, for each sample $(q_i,r1_i,r2_i)$, we queried the LLM-as-a-judge with two prompts $P(q_i,r1_i,r2_i)$ and $P(q_i,r2_i,r1_i)$, and obtained two evaluation results $S_i^{r12}$ and $S_i^{r21}$. Each $S_i$ is $r1_i$, $r2_i$ or "TIE". Then we calculate the Position Consistency as follows:
\begin{equation*}
    \text{Position Consistency}=\frac{\sum_{i=1}^N\mathbb{I}(S_i^{r12}=S_i^{r21})}{N}
\end{equation*}
where $\mathbb{I}(\cdot)$ is the indicator function.

\paragraph{\textbf{Target LLMs and Strategies}}
For LLMs, we selected six LLMs commonly used in the automatic evaluation, including closed-source LLMs GPT-4, GPT-3.5, and open-source LLMs Qwen2.5-7B, LLaMA3-8B, Mistral-7B, and Mixtral-8$\times$7B.

For improvement strategies, we selected \textit{Providing Evaluations with Explanations}, \textit{Self Validation}, \textit{Summarize by Multiple Rounds}, and \textit{Vote by Multiple LLMs}, since these strategies are all straightforward and relatively common in many works. We adopt GPT-3.5 as the base evaluator for the meta-evaluation of these improvement strategies.

\paragraph{\textbf{Model Configuration}}
For closed-source LLMs, we interact using OpenAI's official APIs. The model versions we selected are GPT-4-turbo and GPT-3.5-turbo, specifically referencing gpt-4-turbo-2024-04-09 and gpt-3.5-turbo-0125 respectively\footnote{\url{https://platform.openai.com/docs/models}}.

For open-source LLMs, we adopt Qwen2.5-7B-Instruct\footnote{\url{https://huggingface.co/Qwen/Qwen2.5-7B-Instruct}}, Meta-Llama-3-8B-Instruct\footnote{\url{https://huggingface.co/meta-llama/Meta-Llama-3-8B-Instruct}}, Mistral-7B-Instruct-v0.3\footnote{\url{https://huggingface.co/mistralai/Mistral-7B-Instruct-v0.3}}, Mixtral-8$\times$7B-Instruct-v0.1\footnote{\url{https://huggingface.co/mistralai/Mixtral-8x7B-Instruct-v0.1}}, deployed on an Ubuntu machine equipped with a 40GB NVIDIA A100 GPU. 

To stabilize the evaluation results of LLMs, we set the hyper-parameter \textit{temperature} to 0 to reduce the impact of randomness in LLMs' output. For \textit{Summarize by Multiple Rounds}, we conduct 5 rounds for each sample and verify the effects of three different processing methods for results of multiple rounds: \textit{majority voting}(- majority@5), \textit{taking the mean score}(- mean@5), and \textit{taking the best score}(- best@5). For \textit{Vote by Multiple LLMs}, we conduct experiments on two settings, each involving three LLMs. Setting 1 consists of GPT-4-turbo, GPT-3.5-turbo, and LLaMA3-8B-Instruct, while setting 2 consists of GPT-4-turbo, GPT-3.5-turbo, and Qwen2.5-7B-Instruct.

\subsubsection{\textbf{Results and Analysis}}
\label{subsec:exp_results}
\paragraph{\textbf{Comparison with Different LLMs}}
The experiment results on different LLMs are shown in Table~\ref{tb:exp_results_models}. 
Comparing the evaluation performance of different LLMs, we found GPT-4 outperformed other LLMs with a large margin across all meta-evaluation dimensions and showed fewer biases. 

\begin{table*}[t]
    \small
    \centering
    \begin{adjustbox}{max width=\textwidth}
    \begin{tabular}{lcccccccc}
        \toprule
        \multirow{4}{*}{LLMs} & Alignment & \multicolumn{7}{c}{Biases} \\
        & with & \multirow{2}{*}{Position} & \multirow{2}{*}{Length} & Concre- & Empty & Content & Nested & Familiar \\
        & Human & & & teness & Reference & Continuation & Instruction & Knowledge \\
        & (n=5106) & (n=2633) & (n=34) & (n=28) & (n=26) & (n=24) & (n=24) & (n=24) \\
        \midrule
        GPT-4-turbo & 61.54 & 80.31 & 91.18 & 89.29 & 65.38 & 95.83 & 70.83 & 100.0 \\
        GPT-3.5-turbo & 54.72 & 68.78 & 20.59 & 64.29 & 23.08 & 91.67 & 58.33 & 54.17 \\
        \midrule
        Qwen2.5-7B-Instruct & 56.54 & 63.50 & 64.71 & 71.43 & 69.23 & 91.67 & 45.83 & 83.33 \\
        LLaMA3-8B-Instruct & 50.72 & 38.85 & 20.59 & 57.14 & 65.38 & 75.00 & 45.83 & 54.17 \\
        Mistral-7B-Instruct-v0.3 & 55.42 & 59.78 & 26.47 & 67.86 & 53.85 & 66.67 & 37.50 & 41.67 \\
        Mixtral-8$\times$7B-Instruct-v0.1 & 56.29 & 59.06 & 50.00 & 78.57 & 42.31 & 83.33 & 29.17 & 83.33 \\
        \midrule
        gemini-2.0-thinking & 60.75 & 76.84 & 94.12 & 89.29 & 50.00 & 100.00 & 83.33 & 100.00 \\
        o1-mini & 60.16 & 76.73 & 91.18 & 89.29 & 53.85 & 95.83 & 75.00 & 95.83 \\
        o3-mini & 61.66 & 74.63 & 82.35 & 92.86 & 73.08 & 95.83 & 87.50 & 91.67 \\
        deepseek r1 & 56.48 & 69.17 & 94.12 & 100.00 & 50.00 & 100.00 & 75.00 & 87.50 \\
        \bottomrule
    \end{tabular}
    \end{adjustbox}
    \caption{The meta-evaluation results for different LLMs. All the values are percentages.}
    \label{tb:exp_results_models}
    \vspace{-20pt}
\end{table*}
\begin{table*}[t]
    \small
    \centering
    \begin{adjustbox}{max width=\textwidth}
    \begin{tabular}{lcccccccc}
        \toprule
         \multirow{2}{*}{LLMs} & \multicolumn{2}{c}{human=model1} & \multicolumn{2}{c}{human=model2} & \multicolumn{2}{c}{human=TIE} \\
         & aligned / total & accuracy (\%) & aligned / total & accuracy (\%) & aligned / total & accuracy (\%) \\
        \midrule
        GPT-4-turbo & 1418 / 2071 & 68.47 & 1438 / 2070 & 69.47 & 286 / 962 & 29.73 \\
        
        gemini-2.0-thinking & 1354 / 2070 & 65.41 & 1621 / 2070 & 78.31 & 127 / 962 & 13.20 \\
        
        o1-mini & 1444 / 2070 & 69.76 & 1401 / 2071 & 67.65 & 227 / 963 & 23.57 \\

        o3-mini & 1448 / 2004 & 72.26 & 1206 / 2004 & 60.18 & 399 / 943 & 42.31 \\

        deepseek r1 & 1369 / 2071 & 66.10 & 1342 / 2071 & 64.80 & 173 / 964 & 17.95 \\
        \bottomrule
    \end{tabular}
    \end{adjustbox}
    \caption{The results of each human label in LLMEval$^2$. Only valid responses are counted when calculating the accuracy, while samples that couldn't receive responses due to triggering Azure OpenAI's content management policy are excluded. So there are some differences in the total values of different models.}
    \label{tb:exp_results_llmeval2_detail}
    \vspace{-20pt}
\end{table*} 
\begin{table*}[t]
    \small
    \centering
    \begin{adjustbox}{max width=\textwidth}
    \begin{tabular}{lcccccccc}
        \toprule
        \multirow{4}{*}{Improvement Strategies} & Alignment & \multicolumn{7}{c}{Biases} \\
        & with & \multirow{2}{*}{Position} & \multirow{2}{*}{Length} & Concre- & Empty & Content & Nested & Familiar \\
        & Human & & & teness & Reference & Continuation & Instruction & Knowledge \\
        & (n=5106) & (n=2633) & (n=34) & (n=28) & (n=26) & (n=24) & (n=24) & (n=24) \\
        \midrule
        GPT-3.5-turbo & & & & & & & & \\ 
        \,- base & 54.72 & 68.78 & 20.59 & 64.29 & 23.08 & 91.67 & 58.33 & 54.17 \\
        \,- \textit{w/} explanation & 52.47 & 48.97 & 35.29 & 60.71 & 38.46 & 91.67 & 41.67 & 50.00 \\
        \,- \textit{w/} self-validation & 54.86 & 69.31 & 23.53 & 60.71 & 23.08 & 91.67 & 41.67 & 50.00 \\
        \,- \textit{w/} multi rounds & & & & & & & & \\ 
        \qquad- majority@5 & 54.68 & 70
        11 & 26.47 & 67.86 & 23.08 & 95.83 & 54.17 & 50.00 \\
        \qquad- mean@5 & 54.72 & 69.58 & 11.76 & 57.14 & 26.92 & 87.50 & 50.00 & 50.00 \\
        \qquad- best-of-5 & 51.95 & 58.72 & 5.88 & 42.86 & 19.23 & 87.50 & 37.50 & 45.83 \\
        multi LLMs (\textit{set 1}) & 57.66 & 32.28 & 26.47 & 64.28 & 46.15 & 87.50 & 66.67 & 62.50 \\ 
        multi LLMs (\textit{set 2}) & 58.19 & 70.98 & 64.71 & 71.43 & 69.23 & 91.67 & 45.83 & 83.33 \\
        \bottomrule
    \end{tabular}
    \end{adjustbox}
    \caption{The results for different strategies based on GPT-3.5-turbo. All the values are percentages.}
    \label{tb:exp_results_strategies}
    \vspace{-20pt}
\end{table*}


Therefore, when conditions allow, using GPT-4 as an evaluator may obtain more objective and less biased evaluations. 
For open-source LLMs, we found that Qwen2.5-7B-Instruct showed exceptional evaluation capabilities, outperforming other open-source LLMs in the experiments. 
Moreover, it surpassed GPT-3.5-turbo in most dimensions except for Position Bias and Nested Instruction Bias, indicating that it can be a promising choice as an open-source LLM-as-a-Judge, with the potential to serve as a robust base model for specialized evaluators in specific scenarios.

Additionally, we observed that, apart from Concreteness Bias and Content Continuation Bias, the performance of LLMs, except GPT-4-turbo, was generally poor, particularly in the Length Bias. 
Even GPT-4-turbo experienced substantial degradation in Empty Reference Bias and Nested Instruction Bias. 
While Position Bias can be mitigated by swapping the positions of the inputs, addressing other biases may require researchers to explore more effective evaluation strategies. 
Meanwhile, we also observed that different LLMs in the experiments show no significant differences in the alignment with humans, and there is clear space for improvement.

Therefore, to achieve reliable evaluations, suggestions for model selection include: First, prioritizing powerful or fine-tuned LLMs as foundational evaluators. This approach increases the likelihood of obtaining human-aligned judgments while leveraging stronger instruction-following and comprehension capabilities to handle evaluation prompts with complex strategies. Additionally, it is advisable to conduct small-scale meta-evaluations before model selection to detect potential biases. As experimental results demonstrate, different models exhibit varying susceptibility to distinct biases—even powerful LLMs like GPT-4-Turbo can be severely impacted by specific biases such as Empty Reference. Thus, understanding a model's bias profile before selection aids in formulating effective evaluation strategies and obtaining reliable results.

\paragraph{\textbf{Comparison with Different Strategies}}
Table~\ref{tb:exp_results_strategies} shows the effectiveness of different improvement strategies for enhancing the evaluation performance of GPT-3.5-turbo. 
The results reveal that not all strategies effectively improve LLM-as-a-judge's performance. 
\textit{Providing with Explanation} (\textit{w/} explanation) provides interpretability by offering reasons alongside evaluation scores or selections, which aids in logical backtracking during human review. However, in terms of evaluation performance and bias mitigation, it generally has a negative impact. This performance decline is speculated to be caused by deeper biases introduced by self-explanation. \textit{Self Validation} (\textit{w/} self-validation) shows minimal effectiveness, likely due to the LLMs' overconfidence, which may limit its re-evaluation efforts during self-validation. We will further discuss this limitation in Section~\ref{subsec: challenge_reliability}.

\textit{Summarize by Multiple Rounds} with majority voting (\textit{w/} majority@5) is a strategy with clear benefits, showing improvements across multiple dimensions. It suggests that taking the majority voting results from repeated evaluations helps reduce the impact of randomness in LLMs, thereby addressing bias issues. However, \textit{Summarize by Multiple Rounds} with taking mean score (\textit{w/} mean@5) or with taking best score (\textit{w/} best-of-5) did not improve the evaluation performance and even had some adverse effects. Compared to \textit{w/} majority@5, which selects the major result from multiple rounds, \textit{w/} mean@5 might include results with biases in the mean score calculation, and similarly \textit{w/} best-of-5 could potentially select overly high scores influenced by biases. Therefore, the latter two strategies do not effectively mitigate the impact of biases on automated evaluation. 

The evaluation results of \textit{Vote by Multiple LLMs} (multi LLMs \textit{set 1} and \textit{set 2}) are closely related to the LLM selection. Comparing \textit{set 1} and \textit{set 2}, where LLaMA3-8B-Instruct was replaced by Qwen2.5-7B-Instruct in \textit{set 2}, it revealed significant differences in performance across various dimensions. In \textit{set 1}, the poor performance of GPT-3.5-turbo and LLaMA3-8B-Instruct in the Length Bias negatively impacted the overall performance, whereas in \textit{set 2}, the performance in this dimension was better, which was aligned with Qwen2.5-7B-Instruct. Similar trends were observed in dimensions like Position Bias, Familiar Knowledge Bias, and so on. This suggests that when multiple LLMs are adopted for joint evaluation, the differences between their evaluation performances must be carefully considered.

Based on current experiments, several evaluation strategy design suggestions may be beneficial for achieving reliable evaluations. First, integrating multiple-source results serves as a straightforward and effective approach. After selecting reliable models as evaluators, integrating multi-source results effectively mitigates the negative impacts of inherent noise (e.g., stochasticity in LLMs), enhancing the stability and trustworthiness of evaluation results. Second, experiments comparing \textit{majority@5} and \textit{mean@5} strategies reveal that pairwise evaluations by LLMs yield more reliable results than pointwise—likely because the comparison process better captures nuanced distinctions. Finally, simultaneously generating evaluations and explanations may not be advisable. While explanations help humans understand LLM decision processes, their simultaneous execution may compromise the quality of evaluations.

\paragraph{\textbf{Evaluation of Reasoning LLM-as-a-Judge}}
\label{subsec:evaluation_of_reasoning_model}
As discussed in Sections~\ref{subsec:evaluation_pipeline} and~\ref{quick_practice}, judgment serves as the foundation for effective reasoning capabilities. In other words, models with stronger reasoning capabilities are generally better equipped to perform as reliable judges. To validate this assumption, we conducted evaluations on several reasoning LLMs, including o1-mini, o3-mini, Gemini-thinking, and Deepseek-R1. 
The results in Tables~\ref{tb:exp_results_models} and~\ref{tb:exp_results_llmeval2_detail} provide key insights into the performance of reasoning-focused LLMs. While these models—gemini-2.0-thinking, o1-mini, o3-mini, and deepseek r1—demonstrate \textbf{competitive alignment and accuracy} relative to the top-performing GPT-4-turbo, \textbf{their improvements in tasks requiring human alignment are not as pronounced as expected.}

GPT-4-turbo remains the benchmark for alignment, achieving the highest accuracy rates of 68.47\% and 69.44\% in the human=model1 and human=model2 scenarios, respectively. It also excels in resolving ambiguous cases, with an accuracy of 29.67\% in the human=TIE scenario, outperforming all other models. 
Among reasoning-enhanced models, gemini-2.0-thinking shows strong performance in the human=model2 scenario, achieving an accuracy of 78.27\%—surpassing GPT-4-turbo—highlighting its capability to identify models that align more closely with human judgment. However, this strength does not consistently extend to other tasks or scenarios. Similarly, o1-mini and deepseek-R1, while trailing slightly behind GPT-4-turbo and gemini-2.0-thinking, outperform non-GPT models like Mixtral-8$\times$7B-Instruct-v0.1, demonstrating the added value of reasoning-based enhancements in alignment tasks.
These results indicate that reasoning-enhanced LLMs \textbf{provide meaningful advancements over baseline models but fall short of delivering consistent advantages in alignment-related tasks}, suggesting room for further optimization in this area.

\paragraph{\textbf{Experiment Summary}}
Due to the inherent capabilities and potential risks of LLMs, common improvement strategies for LLM-as-a-judge are not fully effective in improving the performance or mitigating biases. The limitations and challenges will be further discussed in Section~\ref{sec:challenges}. 

Based on the current experimental analysis, an empirical strategy for pairwise comparison evaluation tasks is \textbf{to select more powerful LLMs and to adopt two evaluation strategies: one is swapping the positions of the evaluation contents, the other is taking the majority voting results from multiple rounds of evaluation}, which can effectively mitigate biases. As for improving the alignment with humans, further exploration is still needed.

\subsection{Rethinking Meta-evaluation}
\label{subsec: 4_5_rethinking_eval}


While prior work has introduced various evaluation dimensions, datasets, and metrics, such efforts remain insufficient for evaluating LLM-as-a-Judge. 
Based on a comprehensive review of existing works and the empirical analyses for experiments in sec~\ref{subsec: 4_4_experiment}, it can be observed that current meta-evaluation frameworks continue to face substantial limitations, underscoring the need for more systematic and robust approaches.

(1) \textit{Need for Unified and Comprehensive Benchmark}. 
Given the diverse evaluation dimensions, such as agreement, multiple types of bias, and adversarial robustness, there is a pressing need for a \textbf{unified} benchmark that systematically and comprehensively quantifies these biases within a single framework.
As shown in Table~\ref{table:benchmark}, 
\textit{EVALBIASBENCH} \citep{offsetbias} was proposed as a test set to measure six types of bias. 
Other work \cite{ye2024justice} is dedicated to proposing a unified bias testing process, including automated perturbation and a unified metric. 
They constructed a bias quantification framework \textit{CALM}, which covers 12 types of bias.
Despite these efforts, existing work only covers a subset of evaluation dimensions and lacks a comprehensive framework that includes all relevant aspects.
As a result, many current studies that adopt LLM-as-a-judge still need to design their own meta-evaluation protocols and conduct manual verification to justify the reliability.
Establishing a unified, systematic, and authoritative meta-evaluation benchmark would significantly advance the development and adoption of LLM-as-a-judge.

(2) \textit{Challenges of Controlled Study.} 
When evaluating a specific dimension, especially a particular type of bias, it is often challenging to isolate the bias of interest from other confounding factors such as additional biases or general quality-related characteristics.
For instance, in the case of position bias, lengthening the response could potentially alter the style, fluency, and coherence, or even introduce new biases such as self-enhancement bias. 
Additionally, the tendency for GPT-4 to favor its own responses over those of GPT-3.5 can be interpreted as either self-enhancement bias or a proper tendency towards higher quality text.
Therefore, it is essential for analytical work to carefully control for these variances.

\section{Applications}\label{sec:applications}

\begin{figure*}[h]
  \includegraphics[width=0.85\columnwidth]{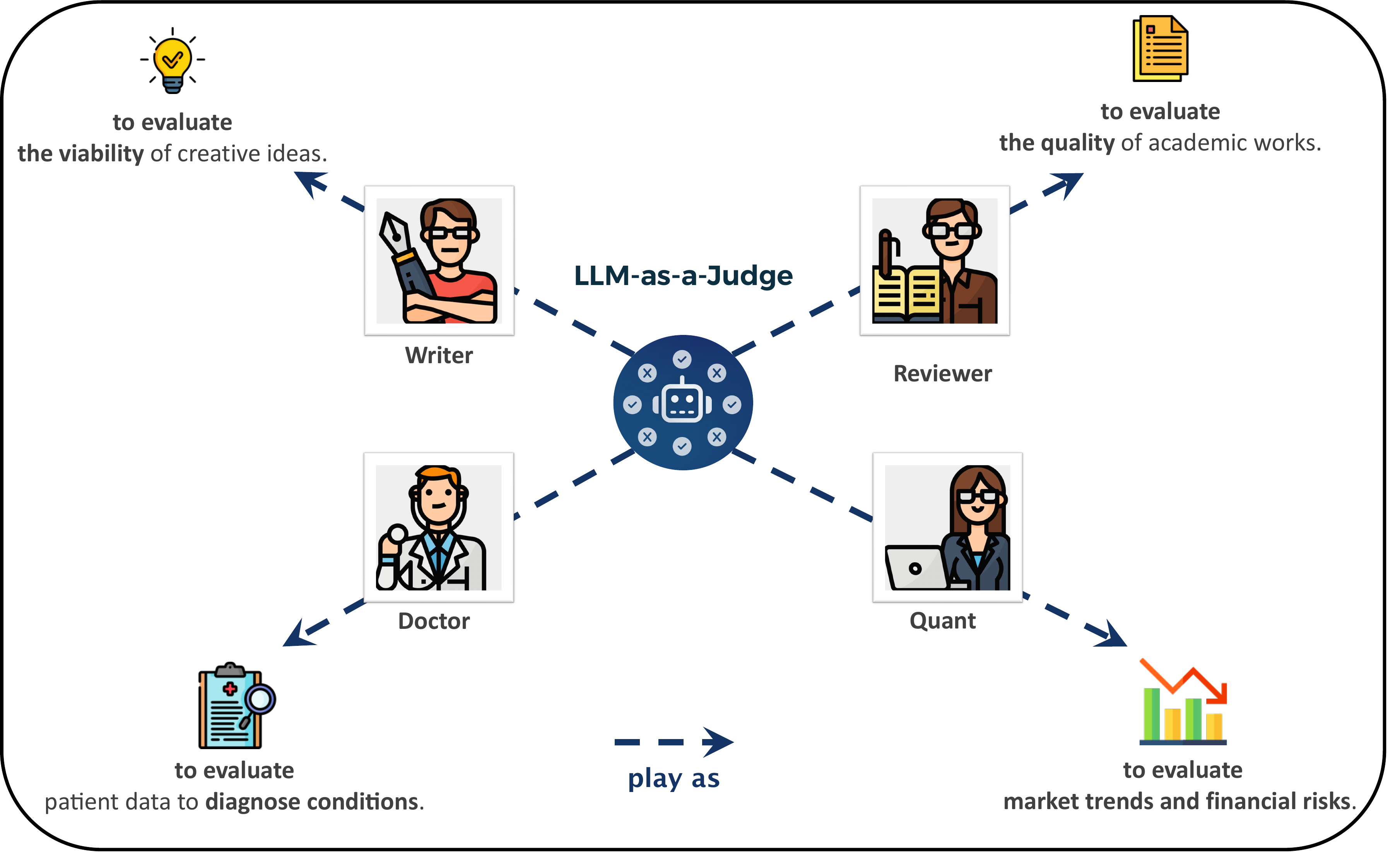}
  \caption{The development process and future prospects of LLM-as-a-Judge.}
  \label{fig:role_play}
\end{figure*}

The application of LLM-as-a-Judge spans a wide spectrum, reflecting both technical advancements and domain-specific demands. In machine learning, they are used to evaluate NLP tasks, assess social intelligence, and support multi-modal evaluation. Beyond the technical sphere, LLMs are increasingly applied in critical domains, as shown in Figure~\ref{fig:role_play}, such as finance, law, and scientific discovery (Ai4Sci)~\cite{inno_aigeo_zhao2024artificial}, where domain expertise and rigorous evaluation are indispensable. 

In the AI era~\cite{inno_new_world}, LLM-as-a-Judge systems are also demonstrating their potential to assist or even replace human judgment across a broad range of professional domains. Many roles inherently require the ability to evaluate, assess, or adjudicate complex scenarios, and LLMs, with their advanced data processing and pattern recognition capabilities, are particularly well-suited to support or enhance these tasks~\cite{inno_environ_wang2023filling}. For instance, writers can leverage LLMs to assess the viability and originality of creative ideas by analyzing narrative structures and market trends; doctors can employ LLMs to diagnose conditions and predict outcomes from medical records and imaging data~\cite{inno_ai_disease_zhang2023confinement,inno_medicine_tang2024llms,inno_cancer_zhong2023artificial}; quantitative analysts can utilize LLMs to forecast market movements and assess risks by identifying patterns in financial data; and judges can rely on LLMs to interpret laws and precedents, aiding in the adjudication of legal cases.
These emerging applications further demonstrate the adaptability of LLM evaluators, opening pathways toward broader integration in practice. This section systematically reviews these domains, outlining how LLM-as-a-Judge contributes to reliable and scalable evaluation in each.


\subsection{Machine Learning}


Large Language Models (LLMs) have emerged as powerful evaluative tools across a wide range of machine learning (ML) tasks, giving rise to the paradigm of ``LLM-as-a-Judge''. This approach leverages the language understanding and reasoning capabilities of LLMs not only to generate responses or predictions, but also to assess, critique, and compare model outputs in various scenarios. Applications span classic NLP tasks, text generation, reasoning and procedural tasks, information retrieval, as well as more recent advances in social intelligence and multi-modal understanding. In the following, we categorize the main areas where LLM-as-a-Judge has been actively applied within machine learning, and analyze the unique requirements and challenges encountered in each domain.

\subsubsection{\textbf{NLP}} LLMs have been successfully employed as evaluators in several NLP tasks, including sentiment analysis, machine translation, and text summarization. In sentiment analysis, numerous biases influencing LLM-based judgments have been identified, prompting the creation of automated frameworks to systematically quantify these biases.

\paragraph{\textbf{Text Generation}} 
Text generation tasks, such as dialog response generation, summarization, story creation, and creative writing, require content that is safe, accurate, and contextually relevant, though there isn't a single "correct" answer~\cite{badshah2024reference, bermejo2024enhancing}. Unlike traditional metrics-based evaluations, LLM-as-a-judge offers a nuanced, adaptable, and customized assessment. According to~\citet{nips_llm_as_a_judge}, LLMs like GPT-4 can evaluate text generation comparably to humans. This method has been used to evaluate outputs from single models and to compare multiple models in competitive settings. For instance,~\citet{human_like_summarization} employ ChatGPT for human-like summarization evaluation, while~\citet{wu2023large} propose a comparison-based framework where LLMs act as judges to evaluate summarization quality.

Modern LLMs excel at generating detailed, long-form responses, but longer outputs increase the risk of hallucinations. To address this,~\citet{cheng2023evaluating} and~\citet{zhang2024balancing} use GPT-4 to identify logically structured yet nonsensical statements. Additionally,~\citet{wang2024halu} propose a critique-based system to evaluate hallucinations by selecting relevant evidence and providing detailed critiques. Beyond hallucinations, generating harmful or unsafe responses is a significant concern. To tackle this,~\citet{salad_bench} introduce MD-Judge and MCQ-Judge for evaluating safety-related QA pairs, focusing on queries designed to provoke unsafe responses. However, an overly cautious approach can lead to excessive refusal responses, affecting user experience. To explore this,~\citet{xie2024sorry} conduct a meta-evaluation of various LLM-as-a-judge frameworks, assessing refusal tendencies in response to potentially unsafe queries. Additionally,~\citet{yu2024xfinder} introduce an LLM-based answer extractor to accurately identify critical parts of answers in text generation, and~\citet{an2023eval} propose L-Eval, a framework for standardized evaluation of long-context language models, followed by~\citet{bai2024longbench} who use LLM-as-a-judge to filter evaluation data for long-context LLMs.

Recent studies have also used LLM-as-a-judge to evaluate the general capabilities of generative models through debate-based frameworks. For example,~\citet{chanchateval} introduce a multi-agent debate framework to facilitate autonomous discussions and assess the quality of generated responses in tasks. Similarly,~\citet{moniri2024evaluating} propose an automated debate framework to evaluate LLMs on domain knowledge, problem definition, and inconsistency recognition.

\paragraph{\textbf{Reasoning}}  
Enhancing the reasoning capabilities of LLMs can overcome the limitations of scaling laws, unlocking their full potential. Effective reasoning is essential for tackling complex problems, making informed decisions, and delivering accurate, context-aware responses. 
\citet{cot} introduce Chain-of-Thought (CoT) prompting to facilitate step-by-step reasoning. More sophisticated cognitive structures~\cite{yao2023tree,hao2023reasoning} have been proposed to further enhance reasoning, yet selecting a reliable reasoning path remains a significant challenge. LLM-as-a-judge has been employed to address this issue.

Some studies focus on sample-level reasoning path selection. \citet{gao2023strategyllm} present a strategy evaluator for assessing candidate strategies. \citet{kawabata2024rationale} propose REPS (Rationale Enhancement through Pairwise Selection), which uses pairwise self-evaluation to select valid rationales. \citet{lahoti2023improving} demonstrate that LLMs can identify and enhance response diversity by aggregating multiple critiques. In multi-agent frameworks, \citet{liang2023encouraging} introduce multi-agent debating (MAD), where a judge LLM selects the most reasonable response. Similarly, \citet{li2024smoa} utilize a judge LLM in layer-based multi-agent collaboration to improve response quality and efficiency.

For step-level reasoning path selection, LLMs act as process reward models (PRMs) to evaluate state scores. \citet{creswell2022selection} break down reasoning into Selection and Inference, using LLMs to judge potential reasoning traces. \citet{xie2024improving} propose the Kwai-STaR framework, transforming LLMs into state-transition reasoners for mathematical reasoning. \citet{let_verify_step} train LLMs as PRMs for inference-time supervision and best-of-N sampling. \citet{setlur2024rewarding} introduce process advantage verifiers (PAVs) to generate rewards based on the likelihood of future correct responses. Advanced cognitive structures are also simulated; \citet{hao2023reasoning} use LLMs as a world model with Monte Carlo Tree Search (MCTS) for deliberate path selection. \citet{besta2024graph} model LLM outputs as graphs to evaluate coherence and logical reasoning.
Additionally, critique-based LLM judges~\cite{ankner2024critique,lancriticeval,yao2024mcqg,yuksel2024multi} provide detailed feedback to enhance the reasoning process.

\citet{yao2022react} pioneered the use of LLMs in an interleaved manner to generate reasoning traces and task-specific actions. Reasoning traces guide the model in updating action plans, while actions facilitate interaction with external sources. Building on this, \citet{yang2023auto} introduced Auto-GPT, which leverages LLM-as-a-judge to enhance tool usage accuracy. By integrating a variety of external tools, LLMs become more versatile, improving planning performance through judicious tool selection. \citet{sha2023languagempc} explored the potential of LLMs in decision-making for complex autonomous driving scenarios, requiring human-like commonsense reasoning. \citet{zhou2024self} employed a self-discovery process where LLMs judge queries and select the most suitable reasoning structure for subsequent inference.

\paragraph{\textbf{Retrieval}}
The role of LLM-as-a-judge in retrieval encompasses both traditional document ranking and dynamic Retrieval-Augmented Generation (RAG) approaches. In traditional retrieval, LLMs enhance ranking accuracy through advanced prompting techniques, enabling effective document ordering with minimal labeled data. RAG frameworks leverage LLMs' ability to generate content guided by retrieved information, supporting applications requiring complex or evolving knowledge integration.

Recent studies have explored LLMs as judges for document ranking, aiming to boost precision and reduce reliance on extensive training data. \citet{zhuang2024beyond} embed fine-grained relevance labels within LLM prompts, enabling models to distinguish subtle relevance variations for refined document ordering.
Innovations in listwise ranking include \citet{ma2023zero}'s Listwise Reranker with a Large Language Model (LRL), which reorders document identifiers without task-specific training data. \citet{zhuang2024setwise} introduce a Setwise prompting strategy for zero-shot ranking, enhancing efficiency without sacrificing performance. To address positional biases, \citet{tang2024found} propose permutation self-consistency, averaging multiple list orders to yield order-independent rankings. \citet{naacl_text_ranker} critique pointwise and listwise ranking prompts, proposing Pairwise Ranking Prompting (PRP) with medium-sized, open-source LLMs as a cost-efficient alternative to larger models.

Recent advancements in RAG have explored LLMs' capacity for self-evaluation and improvement without annotated datasets or parameter adjustments. 
\citet{tang2024self} propose Self-Retrieval, consolidating information retrieval within a single LLM using natural language indexing, transforming retrieval into a document generation and self-assessment process. 
In question answering, LLMs are increasingly used as evaluative agents. \citet{rackauckas2024evaluating} introduce an LLM-based evaluation framework generating synthetic queries from user interactions and domain-specific documents, with LLMs evaluating retrieved documents and ranking RAG agent variants via RAGElo. \citet{zhang2024large} study LLMs' ability to assess relevance versus utility in open-domain QA, demonstrating effective distinction and adaptability with counterfactual passages.

Domain-specific RAG systems reveal LLMs' potential to navigate complex queries by integrating specialized knowledge structures. \citet{wang2024biorag} present BIORAG, enhancing vector retrieval with hierarchical knowledge structures and a self-aware evaluated retriever. \citet{li2024dalk} introduce DALK, combining an LLM with a continuously evolving Alzheimer's Disease knowledge graph, using self-aware knowledge retrieval for noise filtering. \citet{jeong2024improving} propose Self-BioRAG, adapting RAG principles to biomedical applications~\citet{inno_bioinformatics_liu2023bioinformatics}, with LLMs selecting the best evidence for answer generation. 

Within NLP, especially for tasks such as text generation, reasoning and retrieval, LLM-as-a-Judge enables flexible, scalable, and human-aligned evaluation. However, the open-endedness and diversity of NLP tasks (such as dialog or story generation) mean that the requirements for judgment reliability often focus on robustness to context, avoidance of hallucinations, and sensitivity to subtle errors or biases. In domains such as safety or factuality evaluation, higher reliability is needed due to the risk of generating unsafe or misleading content. This specificity raises demands for both the reasoning ability and the transparency of the LLM judge, as highlighted by the increasing focus on critique-based and meta-evaluations.


\subsubsection{\textbf{Social Intelligence}}
As the capabilities of large language models (LLMs) continue to grow, they are increasingly being applied to tasks that require nuanced social understanding—abilities traditionally considered uniquely human. Social intelligence encompasses a range of competencies, including the interpretation of social contexts, adherence to ethical and cultural norms, understanding of emotional cues, and participation in multi-turn interactions that involve negotiation, persuasion, or empathy. Evaluating these capabilities requires moving beyond conventional academic benchmarks toward more interactive and context-sensitive frameworks.

Recent studies have begun to systematically explore the social intelligence of LLMs. For example, \citet{xu2024academically} conduct a comprehensive assessment comparing LLMs' social reasoning skills with their performance on academic tasks. Their findings indicate that although LLMs have made substantial progress in structured problem-solving, they still lag significantly in social intelligence relative to human standards. To facilitate more nuanced evaluation, \citet{zhou2023sotopia} developed SOTOPIA, a simulated environment where multiple LLM-based agents interact in rich social scenarios with assigned goals and social constraints. The accompanying evaluation framework, SOTOPIA-EVAL, employs GPT-4 as an automated judge to assess the agents’ performance on dimensions such as goal achievement, financial decision-making, and maintenance of social relationships. This line of research highlights both the potential and the limitations of current LLMs in replicating human-like social reasoning and interaction. \citet{zhang2025sentient} introduces “Agent-as-a-Judge,” a scalable framework in which large language models are instructed to act as sentient evaluators of multi-agent social interactions. By embedding LLMs in role-playing scenarios that demand theory-of-mind, empathy, and conflict resolution, the study shows that GPT-4-based judges correlate strongly with human ratings (r = 0.83) on higher-order social cognition tasks, while also revealing systematic biases that increase with agent anonymity.

Further efforts are being made to refine evaluation protocols and expand the scope of social intelligence benchmarks. Some researchers are incorporating human-in-the-loop evaluations to calibrate automated judgments, while others are designing more diverse cultural and linguistic scenarios to test cross-cultural adaptability. These developments are critical for deploying LLMs in applications such as virtual assistants, educational tools, and interactive entertainment, where social alignment is essential for user trust and engagement.

\subsubsection{\textbf{Multi-Modal Evaluation}}
The emergence of multi-modal large language models (MLLMs) that integrate text with visual, auditory, and other sensory inputs has created a need for robust evaluation frameworks capable of assessing cross-modal understanding and generation. Multi-modal evaluation introduces unique challenges such as modality alignment, semantic consistency across domains, and the integration of contextual information from heterogeneous sources.

Several recent benchmarks have been developed to address these challenges. \citet{mllm-as-a-judge} introduced a comprehensive benchmark for evaluating MLLMs on tasks including image captioning, visual question answering, and mathematical reasoning with visual inputs. Their study revealed that while LLM-based judges perform well in pairwise comparisons—often matching human preferences—they struggle with absolute scoring and batch ranking tasks, where consistency and calibration are more difficult to achieve. In the context of non-English modalities, \citet{wu2024alignmmbench} presented a benchmark focused on Chinese multi-modal alignment, identifying specific challenges related to coherence and reasoning. They proposed a calibrated evaluation model that significantly improves judgment consistency over existing systems.

To increase transparency in multi-modal evaluations, \citet{xiong2024llava} explored the use of LLM-as-a-judge to not only score model outputs but also generate natural language rationales explaining each assessment. This dual approach improves the interpretability of model judgments and helps developers identify failure modes. In a specialized application, \citet{chen2024automated} constructed the first benchmark for evaluating large vision-language models (LVLMs) on self-driving corner cases. Their results demonstrate that LLM-based judges correlate more closely with human evaluations than judgments provided by LVLMs themselves, underscoring the generalizability of text-based judges even in vision-dominated tasks.

Surveying broader trends, \citet{jiang2024multi} reviewed advancements in multi-modal and multi-agent systems, highlighting mechanisms designed to enhance inter-agent collaboration and reduce cognitive bias. Ongoing research is extending multi-modal evaluation to dynamic video content, audio-visual integration, and embodied AI environments, where temporal and interactive dimensions further complicate assessment. These efforts collectively contribute to more reliable, scalable, and human-aligned evaluation paradigms for multi-modal AI systems.

\subsection{Other Specific Domains}
\subsubsection{\textbf{Finance}}

LLMs have demonstrated significant potential in the finance domain, particularly in tasks such as forecasting, anomaly detection, and personalized text generation~\cite{zhao2024revolutionizing}. As financial applications frequently require rigorous evaluations with high stakes, there is an increasing demand for reliable and transparent LLM-based evaluators tailored to financial contexts.

Current research on LLM-as-a-Judge applications in finance primarily falls into three main areas. Firstly, considerable efforts have focused on designing evaluators that effectively incorporate expert domain knowledge. For instance, Brief et al.~\cite{brief2024mixing} investigated multi-task fine-tuning techniques specifically developed to improve LLM performance in finance-related assessments. Similarly, Yu et al.~\cite{yu2024fincon} introduced FinCon, a multi-agent system that utilises conceptual verbal reinforcement from an LLM-based evaluator to support financial decision-making processes.

Secondly, developing robust benchmarks and evaluation frameworks constitutes a major area of research. Representative efforts include UCFE, a financial evaluation benchmark based on user feedback~\cite{yang2024ucfe}; IndoCareer, a dataset consisting of professional financial examination questions~\cite{koto2024cracking}; and AI-generated, domain-specific evaluation sets designed to systematically assess LLMs’ understanding of financial knowledge and reasoning abilities~\cite{raju2024constructing}.

\begin{figure*}[t]
  \includegraphics[width=0.9\columnwidth]{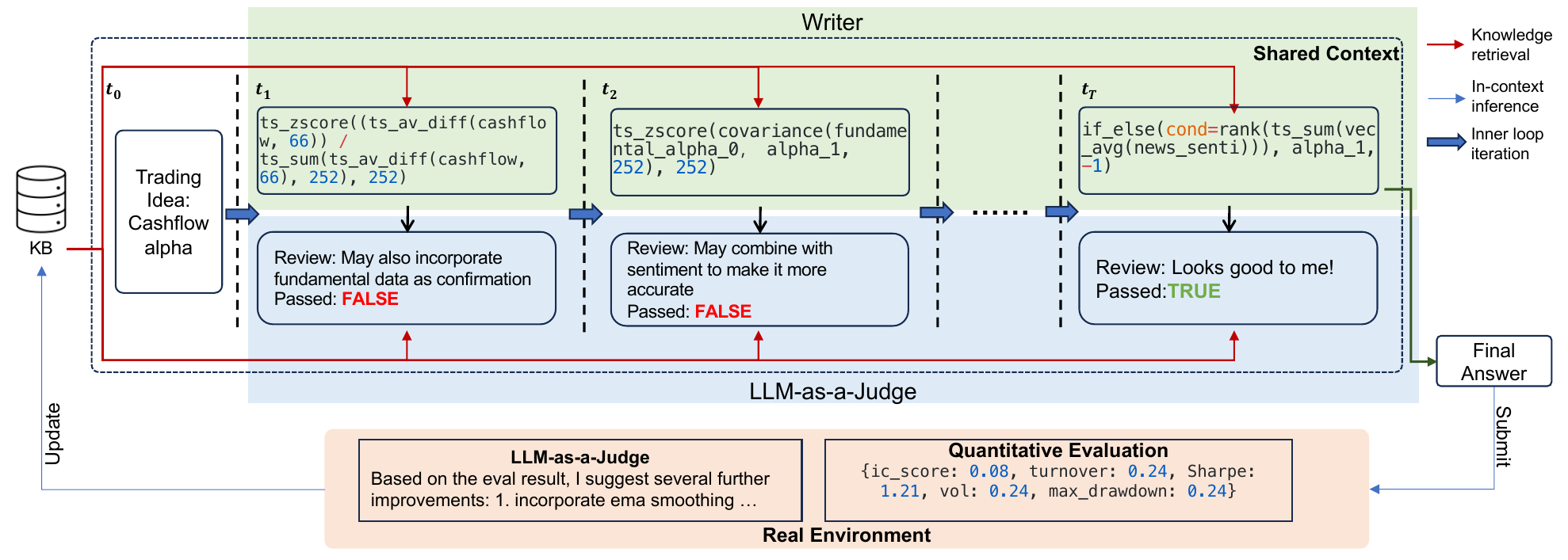}
  \caption{Illustration of using dual-LLM iterative feedback loop for alpha generation in finance. Figure adapted from \citet{wang_quantagent_2024}.}
  \label{fig:quantagent}
\end{figure*}

Thirdly, targeted research addresses specific application scenarios within finance, including quantitative investment strategies, credit scoring, and ESG (Environmental, Social, and Governance) scoring. Wang et al.~\cite{wang_quantagent_2024} proposed the QuantAgent framework, which employs a dual-LLM iterative loop to refine trading signals. In this architecture, one LLM generates initial trading ideas, while a second LLM rigorously evaluates and iteratively improves these ideas using quantitative metrics such as the information coefficient and Sharpe ratio (see Figure~\ref{fig:quantagent}). 
Additionally, recent studies have demonstrated promising LLM-based evaluators for credit scoring~\cite{yoon2023design,babaei2024gpt} and ESG scoring~\cite{zhao2024revolutionizing}, further highlighting the broad potential of LLMs in financial applications.

Despite these advancements, current applications remain in their early stages, constrained by several critical limitations. Ensuring factual accuracy and consistent judgments is challenging, particularly when navigating complex financial regulations or rapidly evolving market data. Additionally, LLMs primarily excel in qualitative assessments and textual analysis—such as reviewing financial news or reports—and currently cannot autonomously perform quantitative tasks such as portfolio optimization or high-frequency trading. As such, their role remains predominantly auxiliary.

In summary, LLM-as-a-Judge applications in finance are in an emergent phase, showing promise in qualitative analysis but limited by inconsistencies in handling dynamic market data and regulatory complexities. Users demand high reliability in the form of factual accuracy and consistent judgments to mitigate financial risks, echoing broader challenges in bias mitigation and robustness discussed earlier in this survey. As such, integrating domain-specific validation will be key to building trust and enabling scalable deployment.

\subsubsection{\textbf{Law}}
LLMs have shown growing capabilities in providing professional advice in specialized fields such as legal consultation, particularly excelling in tasks like text summarization and legal reasoning. Given the complexity, sensitivity, and societal importance of legal decisions, the legal domain exhibits heightened concerns regarding potential biases, factual inaccuracies, and transparency challenges inherent in LLM-based evaluators. Thus, rigorous evaluation methodologies are imperative for responsible deployment in this context.

Current research on LLM-as-a-Judge applications in the legal domain can be categorized into two main areas. Firstly, considerable efforts are devoted to developing specialized evaluators that directly incorporate expert legal knowledge and practices. For example, Ma et al.~\cite{ma2024leveraginglargelanguagemodels} employ general LLMs with expert-designed few-shot prompts to simulate the annotation process, effectively identifying legally relevant facts and precedents. This demonstrates the potential of automated judicial evaluators. Cheong et al.~\cite{cheong2024not} propose a comprehensive four-dimensional framework for developing responsible LLM-driven legal advice systems. This framework explicitly considers user characteristics, query specificity, AI capabilities, and broader social implications, emphasising the need for a nuanced and context-aware approach to LLM integration in legal practice. Similarly, Ryu et al.~\cite{ryu2023retrieval} introduce Eval-RAG, a retrieval-augmented evaluator designed to validate LLM-generated legal texts. In experiments involving Korean legal question-answering tasks, Eval-RAG achieved closer alignment with human expert judgements than conventional evaluation methods, reinforcing the value of integrating domain-specific retrieval mechanisms.

Secondly, the development of comprehensive benchmarks and evaluation datasets represents a crucial research area, aiming to systematically measure and improve LLM legal reasoning capabilities. Representative examples include multi-domain datasets such as IndoCareer, which contains professional legal examination questions~\cite{koto2024cracking}, and LegalBench, a collaborative benchmark specifically designed to assess legal reasoning skills across multiple jurisdictions and languages~\cite{guha2024legalbench}. Additionally, language-specific benchmarks have been developed, including LexEval for evaluating Chinese legal texts~\cite{li2024lexeval} and Eval-RAG for Korean legal contexts~\cite{ryu2023retrieval}. In addition to evaluating technical reasoning abilities, targeted benchmarks have also been introduced to assess ethically sensitive aspects of legal advice, such as ethical reasoning~\cite{zhang2024evaluation} and the potential harmfulness or biases present in generated outputs~\cite{andriushchenko2024agentharm}, highlighting the complexity and sensitivity of LLM evaluation in the legal domain.

Despite these advancements, significant gaps and limitations persist, restricting the widespread adoption of LLMs in the legal domain. Legal reasoning inherently requires precision, transparency, and strict adherence to established statutes and case precedents. However, current LLMs are still prone to factual hallucinations and may overlook subtle but legally significant distinctions. Bias is also a major concern. LLMs trained on large-scale datasets may unintentionally incorporate societal biases, leading to unfair or skewed legal judgments, especially in sensitive cases involving ethics or human rights. Additionally, due to the complexity and variability of jurisdiction-specific legal frameworks, general-purpose LLMs often struggle to capture nuanced differences without detailed contextualization and additional guidance.

Future research should prioritize addressing these limitations through targeted innovations. Promising directions include enhancing factual reliability by integrating robust legal databases and retrieval mechanisms, thereby enabling real-time cross-verification of LLM-generated outputs against authoritative legal sources. Another important avenue is the development of bias mitigation strategies, such as selective filtering of training data or calibration processes guided by expert legal annotations. Crucially, greater interpretability and transparency will be essential. Future LLM judges should ideally provide explicit rationales, relevant statutory references, and coherent justifications for each decision. Ultimately, close collaboration between AI developers and legal professionals will be vital to navigating regulatory frameworks, ensuring ethical compliance, and facilitating the responsible deployment of LLM-as-a-Judge systems within legal contexts.

\subsubsection{\textbf{AI for Science}}

LLMs have demonstrated notable potential in scientific fields, particularly in domains such as medicine and mathematics, where they increasingly act as evaluators to enhance accuracy and consistency~\cite{inno_infectious_diseases_zhou2024harnessing,inno_medicine_tang2024llms,inno_aigeo_zhao2024artificial}.  

Current research on LLM-as-a-Judge in science can be grouped into three major directions. First, specialized evaluators are being crafted for high-stakes clinical reasoning. Brake and Schaaf~\cite{brake2024comparing} showed that a LLaMA-2 evaluator attains human-level agreement (Cohen’s~$\kappa\!=\!0.79$) when checking the internal consistency of clinical notes, while Krolik et al.~\cite{krolik2024towards} reported similarly strong performance for judging medical Q\&A responses. Hosseini et al.~\cite{hosseini2024longform} released the first public benchmark targeting long-form medical QA; their study found that open-weight LLMs acting as judges correlate well with physician ratings on criteria such as correctness, helpfulness and harmfulness.

Second, mathematics research has embraced step-wise reward modeling to verify reasoning chains. WizardMath applies Reinforcement Learning from Evol-Instruct Feedback and surpasses GPT-3.5 on GSM8K and~MATH~\cite{wizardmath2023}. Building on this idea, Math-Shepherd introduces an automatic process-reward model that verifies each reasoning step without human annotations and then reinforces the solver, boosting Mistral-7B accuracy to 84.1 \% on GSM8K~\cite{wang2023mathshepherd}. Tong et al. proposed DART-Math, a difficulty-aware rejection-tuning pipeline that focuses training on hard problems and achieves state-of-the-art results across six math benchmarks~\cite{tong2024dart}. For multimodal scenarios, Lu et al. created MathVista to evaluate textual–visual reasoning, revealing persistent weaknesses when diagrams are involved~\cite{lu2023mathvista}.

Third, comprehensive benchmarking frameworks are emerging to assess scientific LLM judges at scale. Nature Medicine’s 2024 review highlighted gaps in automatic metrics and advocated clinician-in-the-loop pipelines for summarization tasks~\cite{naturemed2024summ}. Stanford HAI’s MedHELM introduced a holistic evaluation suite covering eleven clinical tasks and showed that even top commercial models still lag on medication-safety questions~\cite{medhelm2024}. In mathematics, Xia et al.~\cite{xia2024logicjudge} built a logic-coherence judge that scores entire proof trajectories rather than final answers, providing richer diagnostic signals.

Limitations and future directions.  Despite rapid progress, current medical evaluators often rely on exam-style or short-form outputs and struggle with real-world clinical complexity; hallucination detection remains an open problem~\cite{medhelm2024}. Math-centric judges are still brittle on open-ended or multimodal problems, and exhaustive step verification is computationally costly. Future work should fuse LLM judges with symbolic solvers or medical knowledge bases for fact-checking, incorporate uncertainty estimation (e.g.\ calibrated refusals), and design multimodal-aware judges capable of interpreting figures and tables. Finally, establishing public, diverse scientific benchmarks—paired with transparent reporting of failure cases—will be crucial for trustworthy deployment of LLM-as-a-Judge systems in scientific research and practice.

In summary, the LLM-as-a-Judge paradigm in scientific domains is progressing rapidly, supported by specialised evaluators and benchmarks. However, its uptake remains limited by multimodal constraints and the high cost of verification in complex settings. Reliability requirements therefore centre on hallucination detection and evidence-based rigour to maintain scientific validity, echoing the survey’s discussion of output consistency and adversarial robustness. Integrating large language models with structured knowledge bases is likely to meet these high-stakes needs for precise, verifiable judgements.

\subsubsection{\textbf{Others}}

LLMs have also been employed as evaluators to enhance efficiency and consistency across various fields. In software engineering, a method was proposed for using LLMs to evaluate bug report summarizations, demonstrating high accuracy in assessing correctness and completeness, even surpassing human evaluators who experienced fatigue \cite{kumar2024llms}. This approach offers a scalable solution for evaluation. In education, automated essay scoring and revising have been explored using open-source LLMs, achieving performance comparable to traditional deep-learning models. Techniques such as few-shot learning and prompt tuning improved scoring accuracy, while revisions effectively enhanced essay quality without compromising original meaning \cite{song2024automated}. In content moderation, an LLM-based approach was developed to identify rule violations on platforms like Reddit, achieving high true-negative rates but encountering challenges with complex rule interpretation, emphasizing the necessity of human oversight for nuanced cases \cite{kolla2024llm}. In behavioral sciences, the LLM-as-a-Judge framework was evaluated for assessing user preferences based on personas, revealing limitations in reliability and consistency due to oversimplified personas, but improved significantly through verbal uncertainty estimation, achieving high agreement with human evaluations for high-certainty cases \cite{dong2024can}. These applications of LLMs as evaluators highlight their growing potential in diverse sectors, emphasizing the need for integrating domain-specific knowledge and refining methodologies.

Moreover, LLMs as evaluators demonstrate significant advantages in qualitative assessments that are difficult to quantify, such as evaluating service quality, analyzing user experience feedback, and assessing creative content like art or literature reviews. LLMs' capability to understand and generate nuanced language makes them well-suited for subjective evaluation tasks traditionally requiring human judgment. Future research will focus more on these areas, exploring how LLMs as judges can enhance assessment accuracy and consistency where traditional quantitative methods fall short.

In summary, the LLM-as-a-Judge paradigm is gaining traction across domains such as software engineering, education and behavioural science, where it enables scalable qualitative assessment. Nevertheless, its deployment is constrained by context-specific inconsistencies and by the difficulty of replacing fatigued human reviewers. Reliability requirements differ by field. They often centre on the consistent interpretation of nuanced rules or subjective preferences so as to minimise bias, echoing the survey’s broader themes of prompt optimisation and self-consistency. Therefore, domain-specific adaptation will be essential to satisfy these varied demands and to support reliable, efficient applications.

\section{Challenges}\label{sec:challenges}

In this chapter, we explore the key challenges that arise when utilizing LLMs for evaluation tasks, particularly in the context of LLM-as-a-Judge. Despite their growing capabilities, LLMs still face significant issues related to reliability, robustness, and their backbone models' limitations. Understanding these challenges is crucial for advancing the use of LLMs in a fair, consistent, and reliable manner. We address these concerns under four main themes: reliability, robustness, powerful backbone models, and the ethical and social implications of their use.

\subsection{Reliability}
\label{subsec: challenge_reliability}
The reliability of LLM-as-a-Judge is a primary concern, as it directly impacts the consistency and fairness of evaluations. While human judges also exhibit inherent biases, LLMs introduce their own unique set of reliability issues. These issues stem from the probabilistic nature of the models and their sensitivity to input nuances. We can further break down reliability challenges into several key areas.

\paragraph{\textbf{In-Context Learning Sensitivity}}
LLMs' in-context learning ability, where a model learns from examples in a prompt, can introduce significant reliability issues. Minor changes in prompt wording or the order of examples can lead to unstable and inconsistent results. For instance, \textbf{position bias} is a well-documented issue where LLMs tend to favor the first or last response in a list, leading to unfair evaluations if the examples are poorly arranged. Similarly, the inherent randomness of an LLM's generation can cause inconsistent inter-rater reliability, where the model gives different scores for the same input.

\paragraph{\textbf{Overconfidence and Self-Enhancement}}
A major issue is \textbf{overconfidence}, where models, particularly those trained with Reinforcement Learning with Human Feedback (RLHF), may offer overly favorable scores for their own responses, leading to misleading evaluations. This is often tied to \textbf{self-enhancement bias}, where an LLM is more likely to give a higher score to an answer it generated itself compared to an equivalent answer generated by a different model.

\paragraph{\textbf{Model Selection and Generalization}}
The choice of LLM itself significantly impacts evaluation dependability. The black-box nature and version dependency of commercial models like GPT-4 hinder reproducibility. While fine-tuned evaluators may seem like a solution, they often exhibit \textbf{overfitting}, meaning their evaluation capabilities may not generalize well beyond their training data. These models can also inherit subtle biases from their training datasets, necessitating careful meta-evaluation to ensure fairness.

\subsection{Robustness}
The robustness of LLM-as-a-Judge refers to its ability to resist adversarial attacks and inconsistent inputs. While attacks on traditional Natural Language Generation (NLG) models are well-studied, attacks on LLM-as-a-Judge are relatively under-explored. These attacks aim to exploit a judge model's biases, inconsistencies, or loopholes to manipulate its decision-making process.

\paragraph{\textbf{Adversarial Attacks}}
Unlike traditional attacks that aim to make a model generate harmful content, attacks on LLM-as-a-Judge aim to subtly manipulate the input to change the evaluation outcome. For example, an attacker could introduce \textbf{imperceptible perturbations} to the text---such as paraphrasing a key sentence or adding a misleading but grammatically correct phrase---to trick the judge model into a different conclusion. These attacks are particularly insidious because the manipulated input appears harmless to a human but can cause a significant deviation in the LLM's judgment.

\paragraph{\textbf{Input Sensitivity and Jailbreaking}}
LLM judges are also susceptible to \textbf{jailbreaking} techniques. An attacker could craft a prompt that bypasses the model's safety and fairness filters, causing it to produce a biased or inconsistent evaluation. The model might be prompted to take on a "persona" with specific prejudices, leading to skewed judgments. This is a significant concern for open-source LLM judges, which may not have the same level of fine-tuning for safety as proprietary models.

\paragraph{\textbf{Brittleness of Scoring Mechanisms}}
The reliance on specific scoring formats can be a point of weakness. If a prompt requires an exact numerical score (e.g., "rate on a scale of 1-5"), a clever attacker could create input that confuses the model, causing it to output text instead of a number, thus breaking the automated scoring pipeline. This \textbf{brittleness} undermines the reliability and automation of the entire evaluation system.

\subsection{Limitations of Backbone Models}\label{subsec:challenges_backbone}

The effectiveness of any LLM-as-a-Judge system is directly tied to its underlying backbone model's capabilities. A primary bottleneck lies in the lack of robust models that can serve as reliable judges for complex, multimodal content. While powerful in text, current multimodal LLMs like GPT-4 Vision still struggle with sophisticated reasoning that integrates different modalities. For example, in a medical context, a judge model might need to evaluate a diagnosis based on both a textual description of symptoms and an image of an X-ray. A flawed model could miss a subtle but critical inconsistency between the two, leading to an inaccurate judgment. This limitation poses a significant challenge to achieving reliable evaluations in a wide range of real-world scenarios.

Even for purely text-based tasks, current LLMs have limitations in abstract and causal reasoning. When evaluating a response that requires a deep understanding of logical consistency, a model might produce a seemingly confident but fundamentally flawed evaluation. For instance, in judging a complex scientific paper or legal argument, the model may give a high score for stylistic fluency while failing to identify a subtle but fundamental flaw in the logical chain or causal argument. This gap between the model's superficial fluency and its true reasoning depth undermines trust in its ability to handle sophisticated analytical processes and highlights the need for more robust reasoning backbones.




\subsection{Interpretability and Transparency of Judgments}\label{subsec:challenges_interpretability}

While current LLM-as-a-Judge systems can provide seemingly reasonable scores or conclusions for many tasks, their evaluation process is often an opaque black box. This lack of transparency significantly limits user trust and constrains the application in high-stakes domains such as medicine, law, and education. For instance, in a legal context, an LLM judge might accurately summarize a case and suggest a verdict, but it cannot explicitly show which case precedents it referenced or which legal statutes it prioritized in its reasoning process. This is fundamentally different from human lawyers, who can always provide a traceable rationale for their judgment. A core challenge, therefore, lies in developing methods that can make the reasoning behind LLM-based judgments explicit and verifiable, allowing human experts to trace and validate the model's logic path.

\subsection{Meta-Evaluation and Temporal Consistency}\label{subsec:challenges_meta_eval}

Existing research predominantly focuses on assessing the LLM-as-a-Judge's evaluation results on a specific task, while rarely subjecting the "evaluator itself" to systematic scrutiny. This creates a critical gap, as we lack rigorous benchmarks to measure a judge's accuracy, stability, and bias. This raises a new research imperative: how do we evaluate the evaluator, thereby ensuring the reliability of LLM-as-a-Judge systems? Furthermore, extensive empirical evidence suggests that the performance of LLM-as-a-Judge is not static, with its judgments potentially drifting over time due to model updates or contextual changes. For example, a response to a controversial topic might be rated as acceptable by a model version from March, but subsequently penalized by a June version due to new safety fine-tuning. This "evaluation drift," a type of temporal unreliability, is a particularly pronounced issue in practical scenarios where users demand consistent and stable standards over the long term, ultimately undermining trust and raising concerns about fairness.

\subsection{Ethical and Social Implications}
Beyond technical challenges, the use of LLM-as-a-Judge raises critical ethical and social questions that must be addressed.

\paragraph{\textbf{Bias Amplification}}
LLMs are trained on vast amounts of internet data, which often contains societal biases related to gender, race, and other demographics. When used as judges, these models can amplify and perpetuate these biases, leading to unfair evaluations. For instance, a model might unfairly penalize a response written in a non-standard English dialect or a writing style associated with a marginalized group. This has profound social implications, especially in high-stakes areas like hiring or content moderation.

\paragraph{\textbf{Lack of Accountability and Transparency}}
The "black box" nature of proprietary LLMs makes it difficult to understand how they arrive at a specific evaluation. When an evaluation is deemed unfair or incorrect, there is no clear way to trace the error or hold the model accountable. This lack of \textbf{transparency} undermines trust in the evaluation process and can be particularly problematic in fields where human oversight and accountability are paramount.

\paragraph{\textbf{Impact on Creative and Diverse Outputs}}
By acting as a judge, an LLM can inadvertently shape the types of content being produced. If a model consistently favors a certain style, format, or tone, it could stifle creative and diverse outputs, leading to a homogenization of content. This phenomenon, known as \textbf{evaluation-driven convergence}, could harm innovation and reduce the richness of the information ecosystem.

\section{Future Work}\label{sec:future_work}
The field of LLM-as-a-Judge is rapidly evolving, moving from an emerging concept to a core component of modern AI development. While we've seen significant progress in using LLMs for evaluation, there are still critical challenges and vast opportunities to explore. This section outlines key future research directions that are essential for building more reliable, versatile, and impactful LLM-as-a-Judge systems. We'll delve into topics ranging from advancing the foundational reasoning and judgment capabilities of these models to expanding their application into new domains like data annotation and embodied intelligence. Our discussion also focuses on the necessity of developing more robust theoretical frameworks and benchmarks to ensure these systems are not only effective but also trustworthy and aligned with human values. This forward-looking agenda aims to guide the next generation of research, paving the way for LLM-as-a-Judge to become an indispensable tool in the pursuit of more intelligent and socially beneficial AI.

\subsection{Reasoning-Centric Judgement}\label{subsec:reasoning-centric}
Moving forward, Moving forward, the field is transitioning from traditional evaluation methods to a reasoning-centric approach made possible by LLM-as-a-Judge. We will first examine the symbiotic relationship between reasoning and judgment, highlighting how their synergy is essential for creating sophisticated and capable AI systems. Next, we will discuss how the integration of LLM-as-a-Judge into dynamic feedback loops enables continuous self-improvement, which is key to advancing model capabilities. We conclude with the long-term vision of Self-Evolving Judges, which can adapt and refine their own evaluative abilities over time.

\begin{figure*}[h]
  \includegraphics[width=0.7\columnwidth]{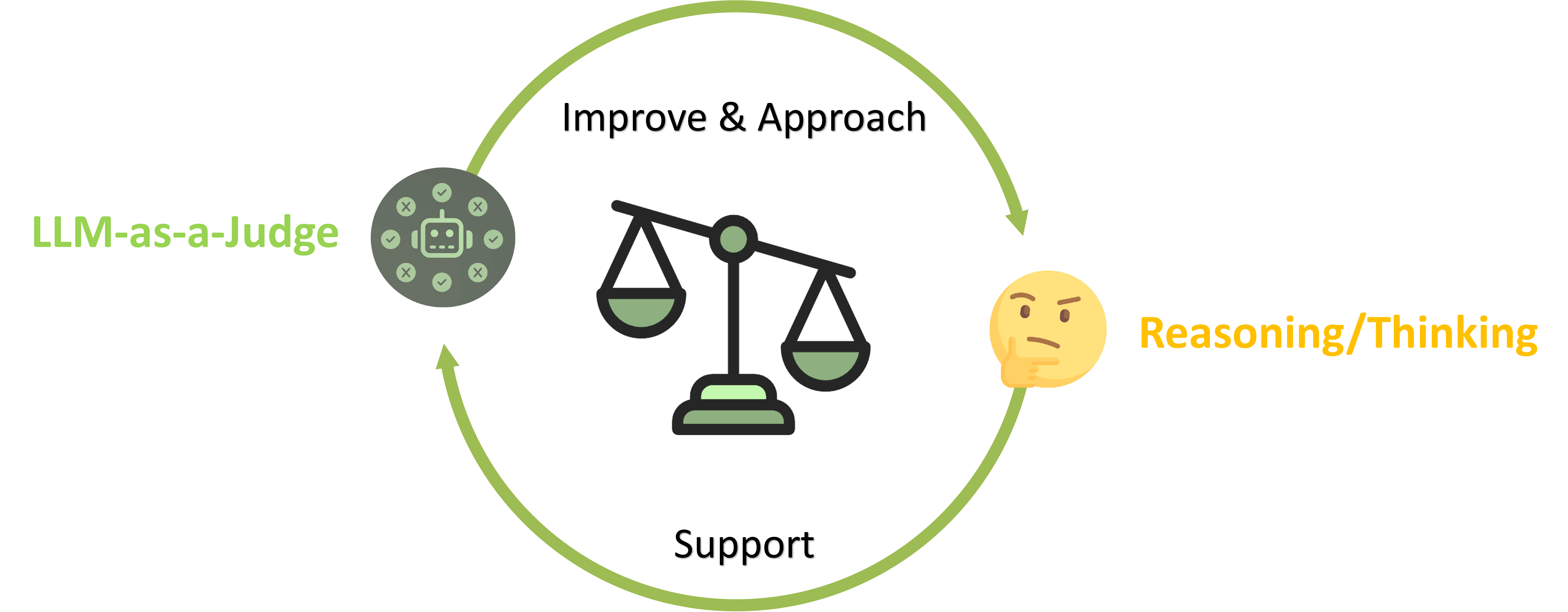}
  \caption{The relationship of LLM-as-a-Judge and Reasoning/Thinking.}
  \label{fig:reasoning_and_judge}
\end{figure*}

\subsubsection{\textbf{The Synergy Between Judgment and Reasoning}}

As shown in Figure~\ref{fig:reasoning_and_judge}, reasoning and judgment are two deeply connected yet distinct cognitive abilities. \textbf{Reasoning} is the logical process of drawing conclusions from evidence or premises. It's the engine that powers problem-solving, decision-making, and critical analysis. In contrast, \textbf{judgment} is the act of evaluating something—an idea, an output, or a situation—against a set of standards or principles to determine its quality or validity. This distinction is crucial, but their relationship is symbiotic.

The philosopher Immanuel Kant famously described judgment as “the faculty of thinking the particular as contained under the universal.” In the context of LLMs, this means applying a general rule or set of principles (the "universal") to a specific output (the "particular") to determine its quality. For example, an LLM-as-a-Judge might evaluate a generated summary against the universal principles of conciseness, accuracy, and coherence. The relationship between these two functions is not one-way; they are mutually reinforcing. Reasoning depends on judgment to validate its intermediate steps. As a model works through a complex problem, it needs to evaluate whether each step in its chain of thought is logical and contributes to a sound final conclusion. This internal evaluation is a form of judgment. Conversely, effective judgment requires strong reasoning to evaluate options against a logical framework. You can't make a good judgment without understanding the underlying logic.

This synergy is at the core of advanced LLM capabilities. When judgment is performed continuously and at a high frequency—for instance, an LLM evaluating every step of its own thought process—it starts to approximate the process of reasoning itself. The more a model systematically evaluates and refines its own thought processes, the more it becomes an effective reasoner. This is why \textbf{LLM-as-a-Judge} is more than just an evaluation tool; it's a mechanism for enhancing a model's reasoning capabilities.

\begin{figure*}[t]
  \includegraphics[width=0.9\columnwidth]{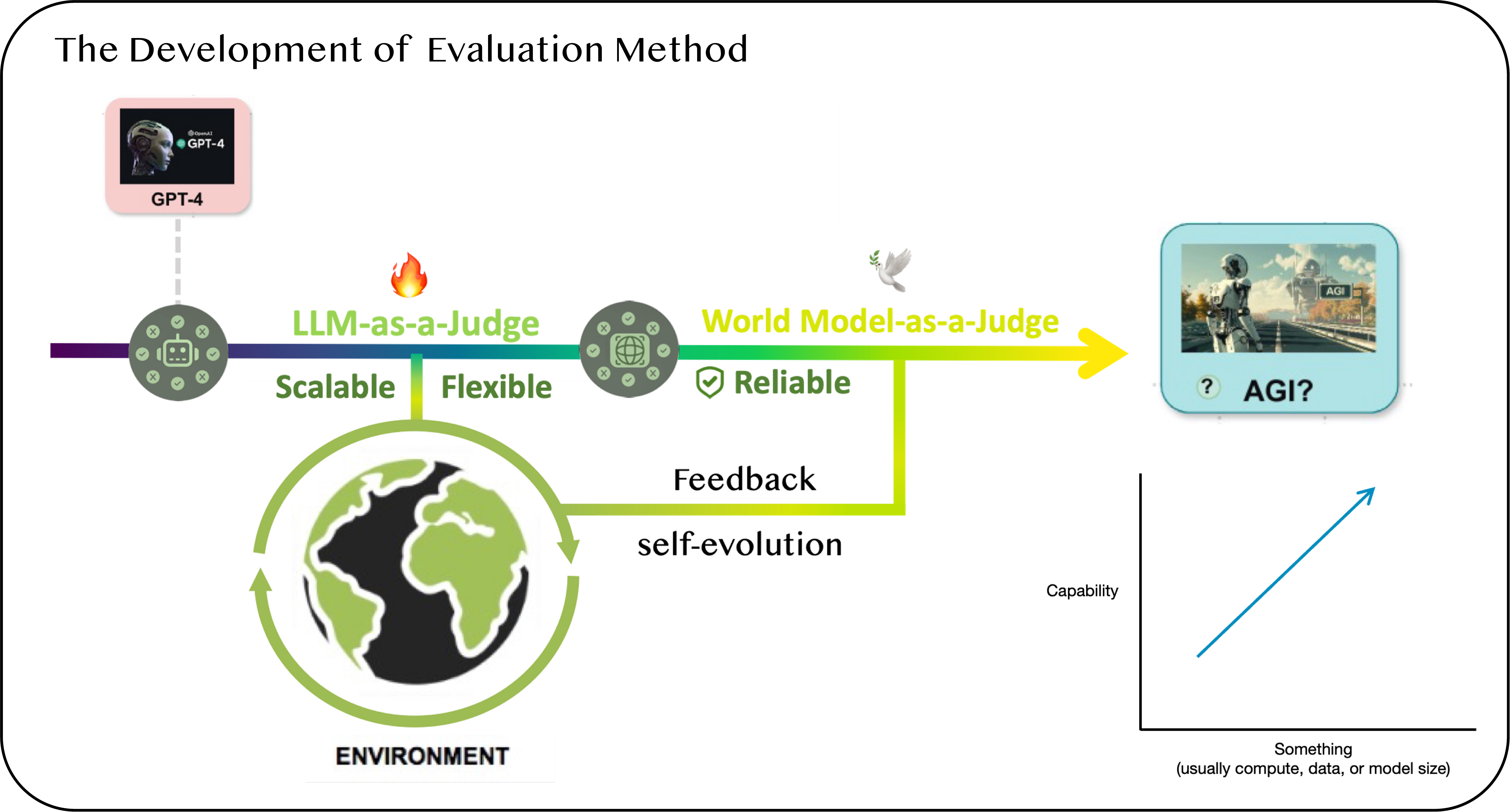}
  \caption{The development process and future prospects of LLM-as-a-Judge.}
  \label{fig:development_llm_as_a_judge}
\end{figure*}

\subsubsection{\textbf{Feedback Loops and Self-Improvement}}

The true power of the LLM-as-a-Judge paradigm is unleashed when it is integrated into a \textbf{feedback loop} that enables continuous self-improvement. A prime example is OpenAI's \textbf{o1} model, which demonstrates enhanced problem-solving ability through structured, iterative reasoning. A central component of o1's advancement is the use of LLM-as-a-Judge modules that evaluate reasoning paths at each stage, offering feedback that improves future steps. This allows the model to correct inconsistencies, identify simpler decompositions of complex problems, and progressively refine its outputs.

This dynamic feedback mechanism operates in two key modes:
\begin{enumerate}
\item \textbf{Training-Time Evaluation}: During the training phase, an LLM-as-a-Judge provides feedback on a model's reasoning process. This feedback is integrated into learning objectives, for example, via RLHF or a similar mechanism. By learning from its mistakes and successes, the model internalizes better reasoning strategies that generalize across tasks.
\item \textbf{Inference-Time Evaluation}: During the testing or deployment phase, the LLM-as-a-Judge dynamically evaluates the model's reasoning as it's happening. This real-time feedback allows the model to make on-the-fly corrections, refining its output and leading to better results without the need for additional training.
\end{enumerate}
This continuous loop of reasoning $\rightarrow$ judgment $\rightarrow$ refinement closely mirrors the principles of \textbf{Constitutional AI}, a framework where models are trained to self-critique and self-correct based on a set of predefined principles. For example, in models like DeepSeek-R1, the LLM acts as its own judge, refining its decisions through internal assessments. This self-generated feedback loop helps the model improve without needing external verification. By integrating LLM-as-a-Judge within such feedback-rich frameworks, models gain the ability to not only evaluate external content but also to introspect and evolve their own reasoning processes. This represents a significant step toward building AI systems that are capable of LLM optimization and self-improvement over time.

\subsubsection{\textbf{Self-Evolving Judges}}\label{subsec:future_self_evolving}

A long-term vision is to enable LLM-as-a-Judge systems to possess self-calibration and self-correction abilities. This would involve evaluators that can continuously refine their biases based on external feedback, thus evolving into more reliable "evaluation agents" over time. This direction aligns with the emerging idea of "World Model-as-a-Judge," as shown in Figure~\ref{fig:development_llm_as_a_judge} where an AI system can use its internal models of the world to make and justify judgments. For example, a judge could not only evaluate a proposed solution to a problem but also simulate the consequences of that solution in a hypothetical environment, identifying potential flaws before they occur. This vision implies that future evaluators will no longer be static tools but rather dynamic, evolving intelligent systems that can adapt and improve their own judgment capabilities, marking a significant step towards more autonomous and trustworthy AI.

\subsection{Theoretically Grounded Evaluation}\label{subsec:future_theoretically}

Current research on the reliability of LLM-as-a-Judge primarily relies on empirical benchmarks, but it lacks a solid theoretical foundation. As a next step, it is imperative to move beyond these empirical approaches and establish a more formal theoretical framework for evaluation. Future work should borrow ideas from fields like statistics and measurement theory to introduce formal definitions of concepts like consistency and robustness. For instance, researchers could adapt established metrics such as Cohen's Kappa or Krippendorff's Alpha to quantify the inter-rater reliability of different LLM judges. Such a framework would allow us to systematically characterize and improve the reliability of these evaluators, moving the field from a trial-and-error approach to one that is scientifically grounded and allows for verifiable, reproducible results.

\subsection{More Reliable LLM-as-a-Judge}
As highlighted in our Formulation~(\S~\ref{sec:formulation}) and Strategy~(\S~\ref{sec:improvement}), LLMs are probabilistic models that require extensive research and optimization to enhance their reliability as judges.
Although current methods have improved the reliability of LLM-as-a-Judge, many challenges, including adaptability and robustness, remain unresolved.
To enable probabilistic models to deliver evaluations closely aligned with real-world scenarios, future research should prioritize refining and implementing LLM-as-a-Judge across the evaluation pipeline.
There is considerable potential for improving reliability in various aspects, including in-context learning, model selection, post-processing techniques, and the overall evaluation framework for LLM-as-a-Judge. These efforts should prioritize not only enhancing the reliability of assessments but also developing methodologies to systematically evaluate and validate the robustness of these assessments. Furthermore, the establishment of comprehensive evaluation benchmarks and interpretable analytical tools will be crucial for assessing and improving the reliability of LLM evaluators.
Finally, the uncertain and evolving nature of robustness risks underscores the necessity of proactive mitigation strategies. These strategies should include the development of adversarial training techniques tailored to judgment tasks, the integration of robust uncertainty quantification methods, and the implementation of human-in-the-loop systems to oversee critical decisions. By addressing these challenges, we can build more resilient and dependable systems capable of maintaining high levels of reliability even under adversarial conditions.

\subsection{MLLM-as-a-Judge}

AI systems are evolving into highly versatile and multifunctional entities~\cite{inno_remote_sensing_data2024multimodal}. Traditionally, specialized models were required for distinct language processing tasks, such as sentiment analysis, syntactic parsing, and dialogue modeling. However, large language models (LLMs) have demonstrated competence across these tasks using a single set of weights~\cite{srivastava2022beyond}. Similarly, advancements are being made toward unified systems capable of processing multiple data modalities. Instead of employing distinct architectures for processing text, audio, and images, recent models like GPT-4o~\cite{openai2023gpt4}, Gemini~\cite{team2023gemini}, and LLaVA~\cite{liu2023llava} integrate these capabilities within a single framework. These developments highlight a growing trend toward unification in the structure and functionality of AI systems, which extends to the emerging paradigm of LLM-as-a-Judge. 

Currently, MLLM-as-a-Judge frameworks~\cite{mllm-as-a-judge} are emerging for evaluating models. However, research exploring how MLLM-as-a-Judge could be applied to the evaluation of data or agents remains limited. Beyond model evaluation, MLLM-as-a-Judge, much like LLM-as-a-Judge, is envisioned to have the capability to assess or annotate data, function as a Reward Model, or serve as a Verifier within intermediate reasoning processes. These expanded roles would allow MLLM-as-a-Judge to contribute more broadly to the AI pipeline.

The future of evaluation lies in developing robust multi-modal evaluators capable of reasoning and assessing complex content spanning text, audio, images, and video. While current multi-modal LLMs exhibit promising capabilities, they often lack the reasoning depth and reliability of their text-based counterparts. Future research must address these limitations, with a focus on enhancing reasoning capabilities, improving reliability, and enabling seamless integration across modalities. A practical multi-modal evaluator has the potential to not only advance AI research but also enable new applications in areas such as multi-modal content moderation and automated knowledge extraction.

\subsection{Advancing Evaluation Benchmarks}\label{subsec:future_benchmarks}

The development of more comprehensive and rigorous benchmarks is critical for advancing the reliability and applicability of LLM-as-a-Judge systems. This effort must proceed on two distinct but interconnected fronts: establishing a framework to evaluate the judge itself and expanding the scope of what the judge can evaluate.

\paragraph{\textbf{Evaluation of the Evaluator (Meta-Evaluation)}}
A critical future direction is the development of systematic meta-evaluation frameworks designed specifically to test the reliability, fairness, and consistency of LLM judges. Unlike existing benchmarks that evaluate a model's performance on a task, these new frameworks would focus on the evaluator itself. For example, a meta-evaluation benchmark could include a carefully constructed set of prompts with known adversarial qualities, such as subtle word substitutions or paraphrasing, to test a judge's robustness to input perturbations. Furthermore, these frameworks must be capable of assessing the judge's stability over time and across different model versions, thereby tracking and mitigating issues like "evaluation drift." This type of framework would be model-agnostic, capable of not only quantifying performance but also providing a basis for explaining a judge's behavior, which in turn fosters greater transparency and trust in the system.

\paragraph{\textbf{Expanding Benchmark Scope}}
While meta-evaluation ensures the integrity of the judge, we must also continue to develop more comprehensive and diverse benchmarks to push the boundaries of what LLM-as-a-Judge can accomplish. Future efforts could focus on creating high-quality, large-scale datasets that encompass a wide range of scenarios, including domain-specific applications, multimodal content, and real-world complexities. For instance, a new benchmark could feature legal documents and require the judge to evaluate the logical soundness of a legal argument, or include a combination of images and text, requiring the judge to identify inconsistencies between them. These benchmarks should also integrate more detailed and fine-grained evaluation metrics that go beyond simple scores. By establishing rigorous standards and datasets akin to ImageNet in scale and impact, the LLM-as-a-Judge field can achieve deeper insights into model performance and accelerate the development of more capable and reliable evaluation methodologies.

\subsection{LLM-as-a-Judge for Data Annotation}
{In contrast, LLM-as-a-judge is a general technique where you use LLM to approximate human labeling. When you ask an LLM to assess qualities like "faithfulness to source," "correctness," or "helpfulness," you define what these terms mean in the evaluation prompt and rely on the semantic relationships the LLM learned from training data. }
Despite its wide applications, data annotation poses significant challenges for current machine-learning models due to the complexity, subjectivity, and diversity of data. This process requires domain expertise and is resource-intensive, particularly when manually labeling large datasets. Advanced LLMs such as GPT-4~\cite{openai2023gpt4}, Gemini~\cite{team2023gemini}, and LLaMA-2~\cite{touvron2023llama2} offer a promising opportunity to revolutionize data annotation. LLMs serve as more than just tools but play a crucial role in improving the effectiveness and precision of data annotation. Their ability to automate annotation tasks~\citep{zhang2022automatic}, ensure consistency across large volumes of data, and adapt through fine-tuning or prompting for specific domains~\citep{song2023preference, ma2024leveraginglargelanguagemodels}, significantly mitigates the challenges encountered with traditional annotation methods, setting a new standard for what is achievable in the realm of NLP.

Whether in the field of scientific research or industry, we are all still suffering from insufficient target data and domain-specific data, or situations where the data quality is not high enough. Assuming that LLM-as-a-judge can achieve stable performance and be fair and reliable, we can use LLM to annotate data in scenarios where data is insufficient to expand the data. In scenarios with low data quality, we can assess the data quality through LLM, and label the quality tags to achieve the goal of selecting high-quality data. Currently, we have not been able to experimentally rely solely on LLM for a reliable evaluation of various scenarios of data; most of the time, we still rely on human annotation to ensure professionalism and reliability. LLM-as-a-judge often needs to learn from human annotations in order to perform certain labeling tasks.

\subsection{LLM-as-a-Judge for Scaling}\label{subsec:llm_judge_scaling}

The paradigm of LLM-as-a-Judge is poised to become a core mechanism for scaling AI development, particularly in an era where models are increasingly trained through interactive and iterative feedback loops. At its most fundamental level, this involves scaling data annotation. Traditional human labeling is a major bottleneck, as it is both costly and slow. LLM-as-a-Judge offers a path to rapidly generate massive, high-quality datasets that would be otherwise impossible to obtain. For instance, it can be used to generate preference labels for RLHF at a scale that is prohibitive for human annotators, thus fueling the development of more sophisticated models. The judge can also be deployed in semi-automated workflows where it provides initial annotations that are then quickly verified by humans, significantly boosting throughput and efficiency. Beyond static data creation, the judge can scale the entire model optimization process by acting as an automated critic or reward model. In multi-agent systems, an LLM judge can evaluate the quality of inter-agent communication and collaboration in real time, and in optimization pipelines, it can provide nuanced, dynamic feedback to guide model fine-tuning and enhance reasoning chains, thus serving as a scalable alternative to traditional, fixed metrics.

\subsection{LLM-as-a-Judge for Embodied Intelligence}\label{subsec:llm_judge_embodied}

While LLM-as-a-Judge has primarily focused on evaluating digital outputs like text and images, its application can be extended to the domain of embodied intelligence. This presents a novel and complex challenge: judging the actions and behaviors of agents in physical or simulated environments. Unlike evaluating a text response, judging an embodied agent requires assessing a sequence of actions, their spatiotemporal relationships, and their alignment with a high-level goal, which necessitates a deep understanding of physics and cause and effect. For instance, an LLM judge could evaluate a robot's performance in a complex task like preparing a meal. The judge would need to assess not only if the final product is correct, but also if the robot's movements were efficient, safe, and followed a logical sequence, such as correctly aligning a cup with a dispenser to avoid spillage. Similarly, in virtual environments or gaming, an LLM judge could assess an agent's strategic ability, navigation skills, or its capacity to solve puzzles in a human-like manner. This judge would act as a crucial feedback mechanism, providing high-dimensional, natural language feedback to guide the embodied agent's learning process. This rich feedback, far more informative than a simple scalar reward signal, could accelerate learning and lead to more generalizable and human-aligned intelligent systems.

\subsection{LLM-as-a-Judge for LLM Optimization}
LLM-as-a-Judge shows substantial promise for advancing LLM optimization. Recent studies~\cite{aaaj} have begun incorporating LLM-as-a-Judge into multi-agent frameworks to guide inter-agent interactions, thereby improving overall decision-making efficiency and quality. In addition, LLM-as-a-Judge has been employed in Reinforced Fine-Tuning (ReFT) pipelines~\cite{trung2024reft}, functioning as a crucial scoring module for evaluating the reasoning processes of models. By flexibly adapting to diverse content formats and domains, LLM-as-a-Judge offers a robust and efficient evaluation mechanism for a wide range of optimization tasks.

Despite these encouraging developments, current research efforts are still in their infancy. Future work should focus on broadening the application domains and strategies for implementing LLM-as-a-Judge, especially in complex, multi-modal scenarios. Furthermore, a systematic assessment of its reliability and generalization capabilities will be critical for fully realizing the potential of LLM-as-a-Judge in enhancing model performance and robustness.

\subsection{Domain-Specific Reliable Applications}\label{subsec:future_domain}

The reliability requirements for evaluation differ significantly across domains. Future work should focus on developing customized LLM-as-a-Judge systems tailored for specific application scenarios, such as medical diagnosis, legal adjudication, educational assessment, and scientific peer review. This requires more than just adjusting prompts or fine-tuning strategies. For instance, a judge designed for legal cases must be trained to prioritize strict adherence to legal precedents and statutes, ensuring its judgments are defensible and auditable. Similarly, a judge for medical diagnosis must be rigorously tested on its ability to understand clinical guidelines and interpret medical jargon correctly. This demands a specialized design concerning evaluation standards, bias control, and social responsibility to meet the unique needs of each field and earn the trust of domain experts.



\section{Conclusion}\label{sec:conclusion}

LLM-as-a-Judge has emerged as a promising paradigm for automated evaluation, offering scalability and adaptability that surpass traditional expert-driven or metric-based methods. By leveraging the reasoning capabilities of large language models, this framework excels in tasks such as text quality assessment, model evaluation, and automated data annotation. It is particularly valuable for large-scale, efficient, and adaptable evaluation. Its ability to process diverse content formats and integrate domain-specific knowledge makes it particularly well-suited for applications in education, peer review, and decision-making systems.

Despite these strengths, several challenges must be addressed to fully realize its potential. Ensuring reliability remains a key issue, because probabilistic outputs can introduce inconsistencies, overconfidence, and biases inherited from training data. Although techniques RLHF have improved alignment with human judgment, they do not eliminate all sources of subjectivity. Moreover, ensuring robustness is another critical concern. LLM-as-a-Judge can be susceptible to adversarial prompt manipulation and contextual framing biases, potentially causing unintended or unreliable evaluations. Finally, generalization across domains and modalities remains a significant hurdle, as current models struggle with evaluating multi-modal inputs, reasoning over structured data, and adapting to domain-specific evaluation standards.

To address these challenges, this paper has offered a comprehensive and principled roadmap. \textbf{First,} at the definitional level, we have provided both formal and informal definitions of LLM-as-a-Judge, thereby establishing the conceptual boundaries of this paradigm. Critically, we introduced a contextualized definition of reliability that accounts for input variability, model characteristics, and contextual dependencies, providing a foundational framework for designing trustworthy systems. \textbf{Second,} at the framework level, we have brought structure to the fragmented literature by organizing existing work around four foundational questions: What is LLM-as-a-Judge? How to use it? How to improve it? and How to evaluate it? This synthesis not only unifies a scattered body of research but also identifies critical gaps and opportunities for future exploration. \textbf{Third, }at the empirical level, we have performed comparative analyses of existing approaches and, more importantly, proposed a novel meta-evaluation benchmark tailored to assess the judge itself. This empirical contribution facilitates systematic performance assessments, revealing crucial trade-offs—such as robustness versus sensitivity—and providing actionable insights for constructing methodologically rigorous and practically deployable evaluation frameworks. \textbf{Finally, }at the perspective level, we have presented a comprehensive analysis that integrates the applications, challenges, and future directions of this paradigm. We have shown that LLM-as-a-Judge can serve as an integral component in high-stakes domains, from finance to law, by identifying unique, domain-specific reliability requirements. Our forward-looking agenda, which emphasizes theoretically grounded methodologies, systematic benchmarks for meta-evaluation, and hybrid human–AI frameworks, aims to guide the community toward LLM-as-a-Judge systems that are not only technically robust but also epistemically sound, socially trustworthy, and broadly applicable in critical sectors.

Ultimately, LLM-as-a-Judge is poised to become an integral component of next-generation evaluation systems, augmenting human expertise rather than replacing it. By addressing the challenges of reliability, robustness, and generalization, we can create more trustworthy, adaptive, and comprehensive evaluators, paving the way for their adoption across scientific research, education, industry, and beyond.







\bibliographystyle{ACM-Reference-Format}

\bibliography{reference}


\end{document}